\let\latexdocument\document
\let\latexenddocument\enddocument
\documentclass{clv3}[shortpaper,manuscript]
\let\document\latexdocument
\let\enddocument\latexenddocument

\makeatletter
\AtBeginDocument{%
  \if@filesw
    \immediate\openout\@mainqry=\jobname.qry
  \fi
}
\AtEndDocument{%
   \ifx\@biography\@empty\else{\par\ifbrief\vskip10pt\fi\biofont\noindent\@biography\par}\fi
   \immediate\closeout\@mainqry
      \ifodd\c@page\clearpage\thispagestyle{empty}\null\clearpage\else\clearpage\fi
}
\makeatother

\NewCommandCopy{\cnumdef}{\numdef}
\NewCommandCopy{\endcnumdef}{\endnumdef}
\let\numdef\relax \let\endnumdef\relax

\usepackage[normalem]{ulem}

\usepackage{xcolor}
\usepackage{latexsym}
\usepackage{amsmath,amssymb}
\usepackage{txfonts}
\usepackage[T1]{fontenc}
\usepackage{yfonts}
\usepackage{inconsolata}
\usepackage{multirow}
\usepackage{makecell}
\usepackage{graphicx}
\usepackage{pgf}
\usepackage{booktabs}
\usepackage{colortbl}
\usepackage{microtype}
\usepackage{xfp}
\usepackage{twoopt}
\usepackage{xspace}
\usepackage[size=tiny]{todonotes}
\usepackage{subcaption}
\usepackage{makecell}
\usepackage{tabularx, booktabs}
\usepackage{floatrow}

\usepackage{enumerate}
\usepackage{makecell}
\usepackage{expex}

\usepackage{pdfpages}

\usepackage{hyperref}
\definecolor{darkblue}{rgb}{0, 0, 0.5}
\hypersetup{colorlinks=true,citecolor=darkblue, linkcolor=darkblue, urlcolor=darkblue}

\usepackage{tipa}
\usepackage{arabtex}
\usepackage[utf8x]{inputenc}
\setcode{utf8}

\newcommand{\ndd}[0]{\textit{n{\small DD}}}
\newcommand{\mdd}[0]{\textit{m{\small DD}}}
\newcommand{\height}[0]{\textit{Height}}
\newcommand{\degree}[0]{\textit{tree{\small Degree}}}
\newcommand{\rootdis}[0]{${\textit{d}}_{\textit{root}}$}
\newcommand{\depvar}[0]{\textit{depth{\small Var}}}
\newcommand{\degvar}[0]{\textit{degree{\small Var}}}
\newcommand{\depmean}[0]{\textit{depth{\small Mean}}}
\newcommand{\degmean}[0]{\textit{degree{\small Mean}}}
\newcommand{\headratio}[0]{${\textit{Ratio}}_{\textit{head-final}}$}
\newcommand{\headdis}[0]{${\textit{d}}_{\textit{head-final}}$}
\newcommand{\leaves}[0]{\textit{\#Leaves}}
\newcommand{\crossing}[0]{\textit{\#Crossings}}
\newcommand{\lpath}[0]{${\textit{Height}}_{\textit{dependency}}$}
\newcommand{\random}[0]{${\textit{d}}_{\textit{randomTree}}$}

\newcommand{\fix}[1]{\textcolor{black}{#1}}
\newcommand{\wz}[1]{\textcolor{black}{#1}}
\newcommand{\se}[1]{\textcolor{black}{#1}}
\newcommand{\sen}[1]{\textcolor{black}{#1}} 
\newcommand{\ab}[1]{\textcolor{black}{#1}}
\newcommand{\yc}[1]{\textcolor{black}{#1}}
\newcommand{\ycj}[1]{\textcolor{black}{#1}}
\newcommand{\ycf}[1]{\textcolor{black}{#1}} 

\newcommand{\ddm}[0]{DDM}

\begin{document}

\dochead{
\se{Syntactic Language Change} 
in English and German:
Metrics, Parsers, and Convergences
}

\runningtitle{\se{Syntactic Language Change} 
in English and German:
Metrics, Parsers, and Convergences}
\runningauthor{}

\author{Yanran Chen} 
\affil{Natural Language Learning Group (NLLG), University of Mannheim, Germany \\ \texttt{yanran.chen@uni-mannheim.de}}

\author{Wei Zhao} 
\affil{University of Aberdeen, UK \\ \texttt{wei.zhao@abdn.ac.uk}}

\author{Anne Breitbarth} 
\affil{Ghent University, Belgium \\ \texttt{anne.breitbarth@ugent.be}}

\author{Manuel Stoeckel} 
\affil{Text Technology Lab (TTLab), \\Goethe University Frankfurt, Germany \\ \texttt{manuel.stoeckel@em.uni-frankfurt.de}}

\author{Alexander Mehler} 
\affil{Text Technology Lab (TTLab), \\Goethe University Frankfurt, Germany\\ \texttt{mehler@em.uni-frankfurt.de}}

\author{Steffen Eger} 
\affil{Natural Language Learning Group (NLLG), University of Mannheim, Germany\\ \texttt{steffen.eger@uni-mannheim.de}}

\maketitle

\begin{abstract}

Many studies have shown that human languages tend to optimize for lower complexity and increased communication efficiency. Syntactic dependency distance, which measures the linear distance between dependent words, is often considered a key indicator of language processing difficulty and working memory load. 
\ab{The current paper looks at diachronic trends in}
\se{syntactic language change}
in both English and German, \ab{using corpora of parliamentary debates from the last c.\ \ycj{160} years}. We base our observations on 
five dependency parsers, including the widely used Stanford CoreNLP \ab{as well as} \ycj{4} newer alternatives. 
\se{Our} analysis \yc{of syntactic language change} goes beyond linear dependency distance and explores 15 
\se{metrics} 
\yc{relevant to dependency distance minimization (\ddm) and/or based on tree graph properties, such as} 
the tree height and degree variance. 
\se{Even though we have evidence that recent parsers trained on modern treebanks are not heavily affected by data `noise' such as spelling changes and OCR errors in our historic data, we find that results of syntactic language change are sensitive to the parsers involved, which is a caution against using a single parser for evaluating syntactic language change as done in previous work. We also show that syntactic language change over the time period investigated is largely similar between English and German for the different metrics explored: only 4\% of cases we examine yield opposite conclusions regarding upwards and downtrends of syntactic metrics across German and English. We also show that changes in syntactic measures seem to be more frequent at the tails of sentence length distributions. To our best knowledge, ours is the most comprehensive analysis of syntactic language \sen{change} using modern NLP technology in recent corpora of English and German.}
\end{abstract}


\section{Introduction}

\begin{figure}[h]
    \centering
    \includegraphics[width=.67\textwidth]{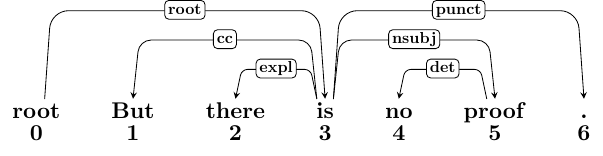}
    \includegraphics[width=.31\textwidth]{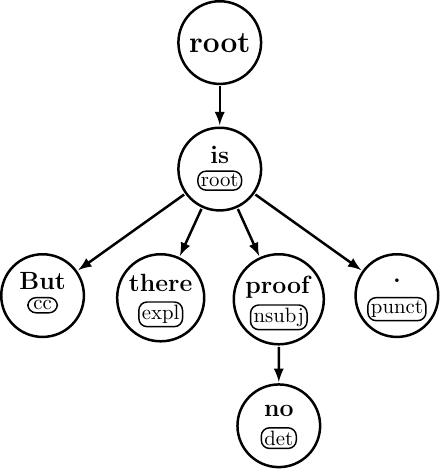}
    \caption{Dependency relations of sentence ``%
    \fix{But there is no proof.}'' in linear order (left) and tree graph (right).}
    \label{fig:intro}
\end{figure}

Many studies have shown 
that human languages are being optimized 
\ab{in the direction of} lower complexity and higher efficiency in terms of communication \citep{gibson2019efficiency,ferrer2022optimality,degaetano2022toward}.
Among others, syntactic dependency distance, i.e., the linear distance between syntactically dependent words, is often perceived as \ab{an} indicator of language processing difficulty and working memory load \ab{in} humans \citep{temperley2007minimization,liu2008dependency,zhu2022investigating,juzek2020exploring}. 
The corresponding optimization principle is well-established,  i.e., dependency distance minimization (DDM), which refers to the tendency/preference of syntactically related words being placed closer to each other.

A large array of works have confirmed the existence of 
\ddm{}
in natural languages (e.g., \citet{temperley2007minimization,futrell2015large,juzek2020exploring,zhang-etal-2020-efficient}).
On the one hand, studies suggest that the mean dependency distance (or its variant\se{s}) of a real sentence is shorter than that of 
random baselines (e.g., 
sentence\se{s} with random word order) and this may be a universal property of most natural languages \citep{gildea2010grammars,futrell2015large,ferrer2022optimality}.
On the other hand, there is also evidence of DDM 
in 
diachronic investigations of language change 
i.e., the mean dependency distance is decreasing over time \citep{lei2020dependency,liu2022dependency,juzek2020exploring,zhang2023investigation,zhu2022investigating,krielke2023optimizing}.

\yc{The recent diachronic investigations involve using dependency parsers to automatically parse 
sentences extracted from a corpus covering a large time span, and observing 
\se{how} dependency distance \se{i}n the 
\se{parsed outputs varies over time} 
(e.g., \citet{lei2020dependency,juzek2020exploring,zhang2023investigation}). The main advantage of such studies over 
approaches 
which leverage human annotated treebanks \citep{gulordava-merlo-2015-diachronic} 
is that they do not require expensive manual dependency annotations and thus allow for exploring arbitrary large diachronic corpora. 
However, to our best knowledge, all of them 
rely on 
\se{a single} dependency parser, mostly the
the Stanford CoreNLP parser \citep{manning-etal-2014-stanford}, which was published 
9 years ago. 
In this work, we first inspect \se{the} parsers' legitimacy in our \se{historical} use case, as they are mostly only trained and evaluated on modern treebanks, which cover a small \se{recent time} period and may 
\se{exhibit}
\se{substantially} less data noise \se{(e.g., OCR errors)} compared to 
\sen{historic}\ab{al} 
corpora (\S\ref{sec:parser}). Subsequently,
we investigate
\emph{whether different parsers yield the same trends regarding language change (\S\ref{sec:anaysis_agree}).}}

\yc{Dependency relations can also be structured as a rooted directed tree, as shown in Figure \ref{fig:intro} (right). Previous works only focus on liner dependency distance; instead, we also explore metrics beyond linear dependency distance, which are relevant to \ddm{} and/or based on tree graph properties, such as the tree height and degree variance (\S\ref{sec:measure}).
}

\yc{Further, the majority of relevant studies are for English, with the exception that \citet{krielke2023optimizing} investigates \ddm{} \ycj{(a.o.)} in both English and German \se{(from 1650 to 1900)}. She compares the trends in scientific and general language, where the texts 
for general language 
\se{span} 
multiple domains, such as news and fiction. 
In our work, we 
focus 
\se{on a homogeneous and directly comparable genre of} 
political debates \se{in both English and German}, \se{stretching from the 1800s to beyond the 2000s}.}

\se{Our main research questions are:
\begin{itemize}
\item RQ1 (parsers): Are parsers trained on modern treebanks reliable to parse (our) historical data, which is affected by OCR and spelling error changes --- especially the German data?
\item RQ2 (parsers): When predicting trends of syntactic language change, can one rely on the predictions of a single parser? 
\item RQ3 (languages): Do English and German mostly change similarly regarding syntax (convergence) or do they have divergent patterns of syntactic language change?
\item RQ4 (metrics): How do English and German change when looking at syntactic dependency graph properties beyond mean dependency distance?
\end{itemize}
}
\ycf{We will make our code+data public upon acceptance.\footnote{\url{https://github.com/cyr19/syntaxchange}}}

\section{Related Work}

\fix{\ab{The current paper} connects to language change with a focus on syntactic  
change 
based on \se{\ddm}. 
}

\paragraph{\textbf{Language Change}}
\fix{Language change has been 
\se{researched for a long time along multiple dimensions (and in multiple communities),}
including 
semantic, syntactic, morphological change etc. 
\textbf{Semantic change} investigates the changes in word meaning over time. For example, 
\citet{giulianelli-etal-2020-analysing} document semantic change 
in the words ``boy'' and ``girl'': 
Girl used to be a term for young people of either sex, while boy described male servants; the present meaning of ``girl'' and ``boy'' were first 
\se{observed in} 
the 15th and 16th centuries, respectively.
Earlier works \se{on studying semantic change} relied on methods like 
latent semantic analysis \citep{sagi2011tracing} or 
treated semantic change detection as a diachronic word sense analysis problem \citep{mitra-etal-2014-thats,frermann-lapata-2016-bayesian}. 
\se{More} recent studies \se{have} leveraged 
\se{static or} contextualized embeddings \citep{pmlr-v70-bamler17a,bamman-etal-2014-distributed,chen-etal-2023-fastkassim,tahmasebi2021survey, rother-etal-2020-cmce,ma2024graph}
, jointly trained 
\se{across} varying periods \citep{pmlr-v70-bamler17a,bamman-etal-2014-distributed,dicarlo2019,dubossarsky-etal-2019-time,haider-eger-2019-semantic}%
, or independently trained for different periods \citep{hamilton-etal-2016-diachronic}
. The latter scenario then involves remapping embeddings for different periods into one shared vector space 
\citep{hamilton-etal-2016-diachronic} or inducing the second-order embeddings, which represents the words' distance to a reference vocabulary \citep{eger-mehler-2016-linearity,rodda2017panta}. 
\textbf{Morphological change} focuses on how word formation and inflection \ab{change} over time \citep{anderson2015morphological}. 
For example, it has been shown that in most Germanic languages, including English, case morphology has gradually declined over time \citep{WeermanDeWit+1999+1155+1192}. 
\citet{moscoso-del-prado-martin-brendel-2016-case} 
explore case changes in Icelandic, using historical corpora annotated with part-of-speech and other morphological information. The rules of verbal inflection have been explored for many languages. 
For example, more frequently used English verbs have undergone faster regularization \citep{lieberman2007quantifying};\footnote{\ab{Note that on the contrary, highly frequent tokens in many cases resist regularization. This is known as the conservation effect \cite{bybee1985}. In German, for instance, lower-frequency strong verbs are historically more likely to adopt the weak inflection, which has a higher type frequency. Opposed to that, highly frequent verbs tend to preserve, and in some cases even extend, irregular inflection, as discussed e.g.\ in \cite{nuebling1998}}.} 
verb inflection tends to be\se{come} weaker in German \citep{carroll2012quantifying} and Dutch \citep{knooihuizen2014relative} over time.
\yc{\citet{ZHU201810} further confirm 
inflection decay, i.e., languages become less inflectional, in Modern English, using entropy-based algorithms.} 
\textbf{Syntactic change} has been discussed 
for changes in specific synta\se{ctic} 
patterns and/or metrics reflecting language complexity. For instance, 
\citet{krielke2021relativizers,krielke-etal-2022-tracing,krielke2023optimizing,degaetano2022toward} investigate 
syntactic shifts in scientific English and/or German, 
observing several phenomena, 
such as 
the tendency towards heavy noun phrases
 and a simple\se{r} sentence structure with decreasing usage of 
clauses 
over time. Similar phenomena were also observed in 
\citet{halliday2003language,banks2008development,biber2011historical,biber2016grammatical}.
One of the widely adopted approaches to model language complexity revolves around syntactic dependency distance. Studies have indicated that humans tend to position syntactically related words closer together \citep{tily2010role,gulordava-merlo-2015-diachronic,liu2017dependency,zhang2023investigation}; this hypothesis/phenomenon is called 
\se{\ddm{}}. 
}

\paragraph{\textbf{Dependency Distance Minimization}}
\fix{The most relevant approaches to ours are those based on sentential dependency structures, where (a.o.) the dependency distance and the corresponding principle (i.e., DDM) 
have been researched extensively. 
From a \textbf{synchronic} perspective, many studies have shown that the mean dependency distance (\mdd) (or 
variant\se{s}) of a real sentence is shorter than that of some random baselines:  
for example, \citet{FerreriCancho2004EuclideanDB} 
and \citet{liu2007probability} observed that the mean dependency distance of Romanian/Czech and Chinese sentences is significantly shorter than that of random sentences, respectively.
\citet{liu2008dependency,futrell2015large} 
provide 
large-scale evidence of DDM
, which suggests that DDM may be a universal property of human languages. They use 
treebanks 
of 
different languages from various linguistic families to perform the tests 
and show 
that the dependency distance of real sentences is shorter than chance. 
More recently, \citet{ferrer2022optimality} analyze 
the optimality (based on dependency distance) of 93 languages from 19 linguistic families. 
Their results imply that 50\% of human languages have been optimized to a level of at least 70\%. Compared to English, 
German has been less optimized\ab{.} 
This finding is in line with \citet{gildea2010grammars}, 
\ab{who observe} that DDM has a much weaker impact on German than on English. 
}

\fix{
On the other hand, \textbf{diachronic} investigation of dependency distance has received an increasing interest in the past few years (e.g., \citet{zhang2023investigation,zhu2022investigating,liu2022dependency}). \citet{tily2010role} may be the first relevant work. 
He 
\se{finds} 
that the total dependency distance of English sentences decreased 
from 900 to 1500 (from Old English to Early Modern English). As the corpora only contain constituency annotations, 
he 
automatically convert\se{s} 
the\se{m} 
into dependency annotations for analysis. 
\citet{gulordava-merlo-2015-diachronic} compare 
Latin / ancient Greek from different time periods using manually annotated dependency treebanks, \yc{which contain around 1k-10k sentences for each period,} finding that languages from the earlier periods exhibit lower DDM levels against the optimum
than those from the later periods.} 

\fix{
More recent studies leverage 
dependency parsers to automatically parse 
sentences extracted from diachronic corpora, and then base 
their observations on the parsing results \citep{lei2020dependency,juzek2020exploring,liu2022dependency,zhu2022investigating,zhang2023investigation,krielke2023optimizing}. An important benefit of such 
\se{approaches} 
is that they do not require expensive dependency annotations
; therefore, arbitrary corpora beyond treebanks can serve as the research resource. \citet{lei2020dependency,liu2022dependency} investigate 
the diachronic trend of dependency distance based on the State of the Union Addresses corpus,\footnote{\url{https://www.presidency.ucsb.edu/}} which contains the political speeches of 43 U.S. presidents from 1790 to 2017. They show 
that the \mdd{} and its normalized 
\se{variant} 
(\ndd{}) \citep{lei2020normalized} of American English sentences are decreasing over time. 
\ycf{While} 
\citet{lei2020dependency} 
observe 
a consistent downward trend across different 
sentence length groups, 
\citet{liu2022dependency} additionally divide 
the sentences of $\leq$ 10 words (the shortest sentence group in \citet{lei2020dependency}) into two subgroups: sentences of 0-4 and 5-10 words, finding the existence of \se{an} anti-\ddm{} phenomenon 
for the 0-4 group, 
\yc{i.e., \ddm{} does not apply to short sentences because of the constraints by other language optimization principles \citep{ferrer2021anti}}.  
\citet{zhu2022investigating,zhang2023investigation} compare 
the diachronic trend of \mdd{} and/or \ndd{} \yc{for the past 100-200 years} across varying text genres, based on the COHA corpus,\footnote{\url{https://www.english-corpora.org/coha/}} which contains 
texts in American English  
\se{across} 4 domains: News, Magazine, Fiction, Non-fiction. 
Their findings are often inconsistent. E.g., \citet{zhu2022investigating} 
\se{find} a downtrend of \mdd{} only for the Fiction genre 
\se{but} 
an upward trend
for the News and Magazine genres \ycj{during the period 1900-2009}. In contrast, \citet{zhang2023investigation} 
observe 
a 
\ycj{continuous} decrease in \mdd{} and \ndd{} for the Magazine genre \ycj{spanning from 1815 to 2009.}
Nevertheless, both of their results imply that the diachronic trends of \mdd{} vary for 
different 
genres. 
\citet{juzek2020exploring} explore 
diachronic shifts in syntactic patterns for scientific English, leveraging the Royal Society Corpus \citep{fischer-etal-2020-royal}, 
which \se{contains} 
scientific publications from 1665 to 1996. They also observe 
a decrease in \mdd{} over time.
As 
follow-up work, \citet{krielke2023optimizing} 
comprehensively investigate\se{s} 
language optimization for scientific English and German from 1650 to 1900, in comparison to that for general language. 
She 
\se{finds} 
that while the \mdd{} for scientific language 
decreased over time, \mdd{} for general-\se{domain} language 
does not show a 
decrease \se{over time} during the same period. 
}

\fix{However, \se{all} those approaches 
have three main limitations: 
(1) all of them rel\se{y} 
on one single parser, mostly the Stanford CoreNLP parser \citep{manning-etal-2014-stanford} with an exception that \citet{krielke2023optimizing} leverage\se{s} 
the UDPipe parser \citep{straka-2018-udpipe}. 
In 
\se{our} work, we 
explore whether different parsers \yc{yield the same}
\se{outcomes} \se{regarding our diachronic investigations}. 
(2) Some of 
\se{the works}
may 
\se{report} biased \se{results} due to 
improper sentence grouping. 
\yc{We show a clear trend of shorter sentences appearing more frequently in
later periods,
consequently, the average sentence length per decade is decreasing over time.}  
\citet{zhang2023investigation} \ycj{do not differentiate sentences by lengths and report the diachronic trends on all sentences.}
 \citet{liu2022dependency,lei2020dependency} divide
 the sentences into groups having 1-10 (0-4, 5-10), 11-20, 21-30, and 31+ words, and analyze 
 trend\se{s} for each group separately. 
 \ycj{However, we demonstrate that even when restricting the maximum difference in sentence lengths to 3 tokens within a group, a slight downtrend in sentence length over time persists (refer to \S\ref{sec:effect_len}).}
 \ycj{Given the substantial length disparity within a group in their works (e.g., 10 for group 11-20), and lacking information on the length distribution in their data, }
 it is hard to tell whether
 the observed downtrend of \mdd{} is the effect of decreasing sentence length or really due to \ddm{}. 
 Those facts \sen{may} cast 
 doubts on the legitimacy of their findings.
 (3) \se{Almost}  
 \se{all} of \se{the works} 
 focus on English change, except for  
 \citet{krielke2023optimizing}, who compare\se{s} 
 the changes in scientific and general English and German. 
 The dataset for general languages used  
 \se{in \citet{krielke2023optimizing}}
 contain texts in multiple genres like fiction and news.
 Nevertheless, as \citet{zhu2022investigating,zhang2023investigation} suggest, trends of \mdd{} may differ for various text genres.
 Instead, in this work, we \yc{focus on one specific genre, namely \emph{political debates}, \se{across both languages}.}
}

\section{Data Preparation}
\label{sec:data}

\paragraph{\textbf{Corpora}}
\begin{figure}[!ht]
    \centering
    \includegraphics[width=\textwidth]{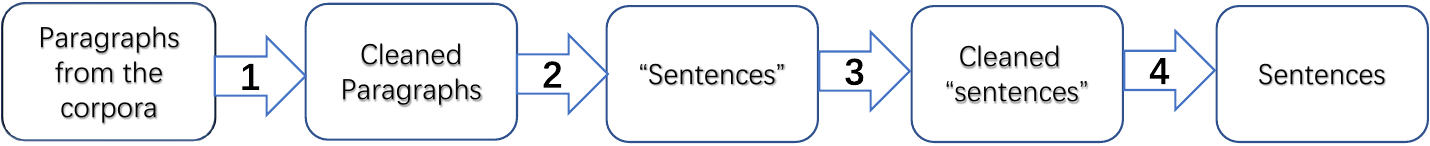}
    \caption{4-step pipeline for sentence extraction from the corpora: \ycj{1.\ \emph{paragraph-level preprocessing}, 2.\ \emph{sentence segmentation with Spacy}, 3.\ \emph{postprocessing}, 4.\ \emph{filtering}.}}
    \label{fig:preprocessing}
\end{figure}
In this work, we use corpora consisting of political debates and speeches in 
\se{German and English}
to ensure a fair comparison. For \textbf{English}, we select the \textbf{\texttt{Hansard}} corpus, which comprises the official reports of UK parliament debates \se{since 1803}.
To compile the data for the time period 1803—2004, we extract XML files from the \texttt{Hansard} archive,\footnote{\url{https://www.hansard-archive.parliament.uk/}} retaining every section labeled as ``debate''. 
Data from 2005 to April 2021 is obtained from a Zenodo repository.\footnote{\url{https://evanodell.com/projects/datasets/hansard-data/} (v3.1.0)} 
For \textbf{German}, we utilize the \textbf{\texttt{DeuParl}} corpus published by \citet{FairGer};\footnote{\sen{The original corpus was published in \citet{deuparl:walter}}. The version we use contains further clean-up.}
it contains plenary protocols from both the Reichstag (the former German parliament\ab{, until 1945}) and the Bundestag (the current German parliament), covering the period from 1867 to 2022.

\begin{table}[!ht]
\small
\centering
\begin{tabularx}{.9\textwidth} { >{\hsize=.31\hsize}X | >{\hsize=.12\hsize}X | >{\hsize=.21\hsize}X | >{\hsize=.22\hsize}X}
\toprule
\multicolumn{2} {l|}{Hansard} & \multicolumn{2} {l}{DeuParl} \\ \midrule
\makecell[l]{``The Earl of Liverpool [...]\\
\colorbox{red}{[238}\\
stance, when he [...]} & 
\makecell[l]{
\colorbox{red}{Blankets}\\
\colorbox{red}{18,800}\\
\colorbox{red}{13,500}\\
\colorbox{red}{6,360}}
&
\makecell[l]{
Ich will Ihnen \colorbox{red}{z.}\\
\colorbox{red}{B.} sagen, ist [...] \\
}
& 
\makecell[l]{
\colorbox{red}{(Bravo!}\\
\colorbox{red}{Rechts.)}\\
Reichstag des [...]\\
\colorbox{red}{20.}\\
\colorbox{red}{Sitzung am 27.}\\
\colorbox{red}{März 1867.}
}
\\ \bottomrule
\end{tabularx}
\caption{Examples of problematic data. Line breaks in the table signify actual breaks in the data, \ycf{with problematic texts highlighted in red.}}
\label{tab:realdata}
\end{table}

\paragraph{\textbf{Preprocessing}}
To extract sentences from the corpora, we employ a 4-step preprocessing approach, outlined in Figure \ref{fig:preprocessing}. Initially, we conduct preprocessing at the paragraph level, with the aim of providing cleaner inputs for the sentence tokenizer. For example, as 
Table \ref{tab:realdata} (first column) shows,
in the original data of \texttt{Hansard}, the page numbers are often not excluded from the sentence due to OCR errors, which occupy an individual line (``[238''). In \texttt{DeuParl}, the text is often split into lines by periods, no matter whether they serve as the ending punctuation. As shown in 
Table \ref{tab:realdata}, ``z.B.'' \fix{\se{(}English: ``e.g.''\se{)}} is 
accidentally split into two lines (third column); a similar case exists for ``27.\ März'' \fix{(English: ``27.\ March'')} in the last column. 
With paragraph-level preprocessing, we delete the line break and the page numbers (for example) before applying the sentence segmentation.
In the case of \texttt{DeuParl}, which is in plain text format, we first divide each file into paragraphs of 50 lines, in order to ensure not to exceed the maximum input length of the sentence tokenizer. For \texttt{Hansard}, the paragraphs are naturally organized within the XML files. Subsequently, we utilize Spacy \citep{spacy2} 
to segment the paragraphs into sentences. The next step involves 
postprocessing on the segmented paragraphs, primarily focused on correcting errors stemming from the sentence tokenizer. 
Furthermore, we find inconsistent usage of semicolons as sentence-ending punctuation. For example, in the UD treebanks, semicolons are occasionally permitted as the concluding punctuation of sentences; at other times, multiple individual sentences connected by semicolons are treated as one sentence. However, the sentence tokenizer tends to split the sentences by semicolons.
With simple postprocessing, we manage to correct such segmentation errors \ycj{by concatenating sentences separated by semicolons}. 
Finally, we apply a filtering process to exclude texts that may not qualify as complete and standalone sentences;
this is achieved through the following simple and intuitive rules: 
\begin{itemize}
    \item Sentences must start with a capitalized character.
    \item Sentences must end with a period, or a question mark, or an exclamation mark.
    \item Sentences must contain a verb based on the part-of-speech 
    tags.
    \item The number of (double) quotation marks must be even.
    \item The number of 
    left brackets must be equal to that of 
    right brackets.
\end{itemize} 

It is possible that such preprocessing/filtering biases the results later discussed to some extent, as some sentences 
\se{are} ignored through it; however, 
directly applying sentence segmentation to the original data leads to incomplete sentence segments or data from other parts, e.g., a budget table, as 
Table \ref{tab:realdata} (second column) shows, 
which can 
result in larger biases for the mismatch with the 
definition 
of the dependency parsing task, i.e., parsing the syntactic structure of a sentence, and consequently affect the metrics calculated on a sentence level.



\subsection{Validation \& Correction}
To validate the feasibility of our preprocessing pipeline
and study the influence of data noise on metrics such as dependency distance (discussed in \S\ref{sec:analysis_noise}), for each corpus, 
we
apply 
the pipeline 
to five randomly sampled paragraphs from each decade, resulting in approximately 150-800 sentences per decade. We then manually review ten randomly selected outputs for each decade and make corrections if any issues are identified. Additionally, the data from the 2020s is merged into the 2010s, given that only partial data is available for the former.

\begin{table}[!h]
\small
    \centering
    \newcolumntype{Y}{>{\centering\arraybackslash}X}
    \begin{tabularx}{\textwidth} { >{\hsize=.32\hsize}X | >{\hsize=.06\hsize}Y | >{\hsize=.07\hsize}Y | >{\hsize=.07\hsize}Y | >{\hsize=.09\hsize}Y | >{\hsize=.07\hsize}Y | >{\hsize=.32\hsize}X }
    \toprule
    \textbf{text} & \textbf{date} & \textbf{is sent} & \textbf{has issues} & \textbf{issues} & \textbf{origin of issues} & \textbf{correction}  \\ \midrule
    Ich bitte, daß diejenigen Herren, welche für den Fall der Annahme des Z 33b in demselben die Worte „oder an Druck m anderen öffentlichen Orten" aufrecht erhalten wollen, sich von ihren Plätzen erheben. & 1883-4-6 & TRUE & TRUE & extra material; symbol; spelling
    & ocr; ocr; historic & Ich bitte, dass diejenigen Herren, welche für den Fall der Annahme des § 33b in demselben die Worte „oder an anderen öffentlichen Orten" aufrecht erhalten wollen, sich von ihren Plätzen erheben. \\ \bottomrule
    \end{tabularx}
    \caption{Illustration of the formatted annotation.}
    \label{tab:annotation}
\end{table}

\paragraph{\textbf{Annotation}}
We first check the \textbf{German} sentences from \texttt{DeuParl}. Two annotators with an NLP background, one male faculty member who is a German native, and one female PhD student who speaks German fluently, check the outputs. In this phase, we do not have annotation guidelines; the annotators are asked to freely make comments on each text 
\se{following} 
their intuition and make corrections if they identify any issues in the text.\footnote{The annotation is done with Google spreadsheet.} 
Subsequently, by inspecting the comments and comparing them to the original PDF files,\footnote{\url{https://www.reichstagsprotokolle.de/}} 
we standardize and format the annotation scheme. As 
\ycj{Table \ref{tab:annotation}} shows, 
the formatted annotation has 5 columns: 

\begin{itemize}
    \item \textbf{(1) is\_sent}: A \textbf{binary} annotation (`TRUE'/`FALSE') for sentence identification, i.e., whether the text is a sentence (minor issues in the sentence, such as space or spelling errors, are allowed).
    \item \textbf{(2) has\_issues}: A \textbf{binary} annotation (`TRUE'/`FALSE') about the existence of any issues (only if (1) is `TRUE').
    \item \textbf{(3) issues}: A sequence of labels for the \textbf{category of the issues} identified, e.g., spelling errors  
    (only if (1) and (2) are `TRUE'). 
    \item \textbf{(4) origin\_of\_issues}: A sequence of labels for the \textbf{origin of the issues}. For example, if 
    \se{a}
    spelling error 
    \se{is} a consequence of OCR errors, then the ``origin'' of 
    \se{it} should be ``OCR''.
    The labels here are aligned in
    a one-to-one manner with those in (3) (only if (1) and (2) are `TRUE').
    \item \textbf{(5) correction}: The \textbf{corrected sentence} (only if (1) and (2) are `TRUE').
\end{itemize}

\begin{table}[!ht]
\small
\begin{tabularx}{\textwidth} { >{\hsize=.2\hsize}X | >{\hsize=.5\hsize}X | >{\hsize=.3\hsize}X }
\toprule
\textbf{Category}              & \textbf{Example}                                                                                                                                                                                                                                                & \textbf{Correction}                           \\ \midrule
Spelling              & Das ist die Mehrheit; die Diskussion ist {\color[HTML]{FE0000} geschloffen}.                                                                                                                                                                                                  & {[}...{]} geschlossen.               \\\midrule
Space                 & Der EVG-Vertrag spreche von der „westlichen Verteidigung", und im Protokoll der NATO-Staaten sei vom Zusammenschluß der {\color[HTML]{FE0000}w e s t europäischen} Länder die Rede.                                                                                          & {[}...{]} westeuropäischen {[}...{]} \\\midrule
Missing Material      & Ich bin der Meinung — und viele an uns herangekommene Klagen lassen auch darauf schließen {\color[HTML]{FE0000}—} , dass die Auffassung des Reichssparkommissars, dass das Personal der Deutschen Reichspost voll ausgelastet, aber nicht überlastet sei, nicht richtig ist. & add ``—''                            \\\midrule
Extra Material        & 
[...] daß durch den Bund zweierlei Recht für die Norddeutschen geschaffen werden soll, {\color[HTML]{FE0000}(Sehr richtig!)} daß gewissermaßen zweierlei Klassen von Norddeutschen geschaffen werden sollen, {\color[HTML]{FE0000}(Sehr gut!)} eine Selekta, die vermöge ihrer Gesittung [...]

& delete ``(Sehr gut!)'' and  ``(Sehr richtig!)''                       \\\midrule
\makecell{Punctuation \\\& \\Symbol}
 & Ich bitte, daß diejenigen Herren, welche für den Fall der Annahme des {\color[HTML]{FE0000}Z} 33b in demselben die Worte „oder an Druck m anderen öffentlichen Orten" aufrecht erhalten wollen, sich von ihren Plätzen erheben.                                              & 
 [...] § 33b [...]\\ \bottomrule
\end{tabularx}
\caption{\textbf{Categories of issues}; each is shown with an example and the corresponding correction. The parts having a certain issue or missing in the sentences are in \textcolor{red}{red}.}
\label{tab:issue_example}
\end{table}

We classify the \textbf{issues} into 5 categories: \emph{Spelling}, \emph{Space}, \emph{Missing Material}, \emph{Extra Material} and \emph{Punctuation \& Symbol}, as shown in Table \ref{tab:issue_example}. Additionally, we split the \textbf{origins of issues} into 4 categories: (1) \emph{Historic}, which indicates the issues are due to the differences in historical and modern language. For example, `daß' \ycf{(modern spelling: `dass')} is a spelling issue with the \emph{historic} origin. (2) \emph{OCR}, which indicates the issue is an OCR error. For example, the spelling issue in Table \ref{tab:issue_example} results from 
\se{an} OCR error \ycf{(`geschloffen' vs.\ `geschlossen').}
%
(3) \emph{Genre}, which represents 
issues that exist due to the characteristics of the text genre. In the text of 
political debate\se{s}, 
the language changes between written and spoken language frequently, and also, there exist interjections within a sentence, as shown in the fourth row of Table \ref{tab:issue_example} (Extra Material). (4) \emph{Preprocessing}, which denotes that the issues stem from our preprocessing 
pipeline, \se{e.g., extra space}. For details regarding those categories, 
we refer to Figure \ref{fig:guideline} in the appendix, which demonstrates 
our annotation guidelines 
following the refined annotation scheme.



Next, two annotators annotate the \textbf{English} text from \texttt{Hansard}, following the annotation scheme defined above. 
As the original PDF files of \texttt{Hansard} are inaccessible, we cannot identify the origins of the issues; thus, this annotation is omitted for \texttt{Hansard}.
Among the annotators, one is a male faculty member, while the other is a female Master's student, \yc{both of whom speak English fluently and work in the NLP field}. 

To check the \textbf{agreement} among annotators,
40 German sentences are jointly annotated, 20 for the sentence identification task and 20 for the issue identification/correction task. For English, 30 sentences are commonly annotated for all annotation subtasks.
For each binary annotation task, we calculate Cohen's Kappa \citep{cohen1960coefficient} for inter-agreement, whereas for the correction task,  
we compute the edit distance \citep{Levenshtein1965BinaryCC} between sentences. Let $s$ be the original sentence, and $s'_{a}$ and $s'_{b}$ be the sentences corrected by annotators A and B respectively. We calculate the edit distance (i) between $s$ and $s'_{a}$ ($ED(s,s'_{a})$), (ii) between $s$ and $s'_{b}$ ($ED(s,s'_{b})$), and (iii)  between $s'_{a}$ and $s'_{b}$ ($ED(s'_{a},s'_{b})$). 
If the corrections are similar to each other, (i) and (ii) should be close while (iii) should be small. 
Overall, the annotators obtained a decent level of agreement, with $\sim$ 0.7 on binary tasks and the edit distances of (i) = 3.88, (ii) = 3.96, and (iii) = 0.54 for the correction.

\begin{figure}[!ht]
    \centering
    \begin{subfigure}[t]{0.5\textwidth}
        \includegraphics[width=\linewidth]{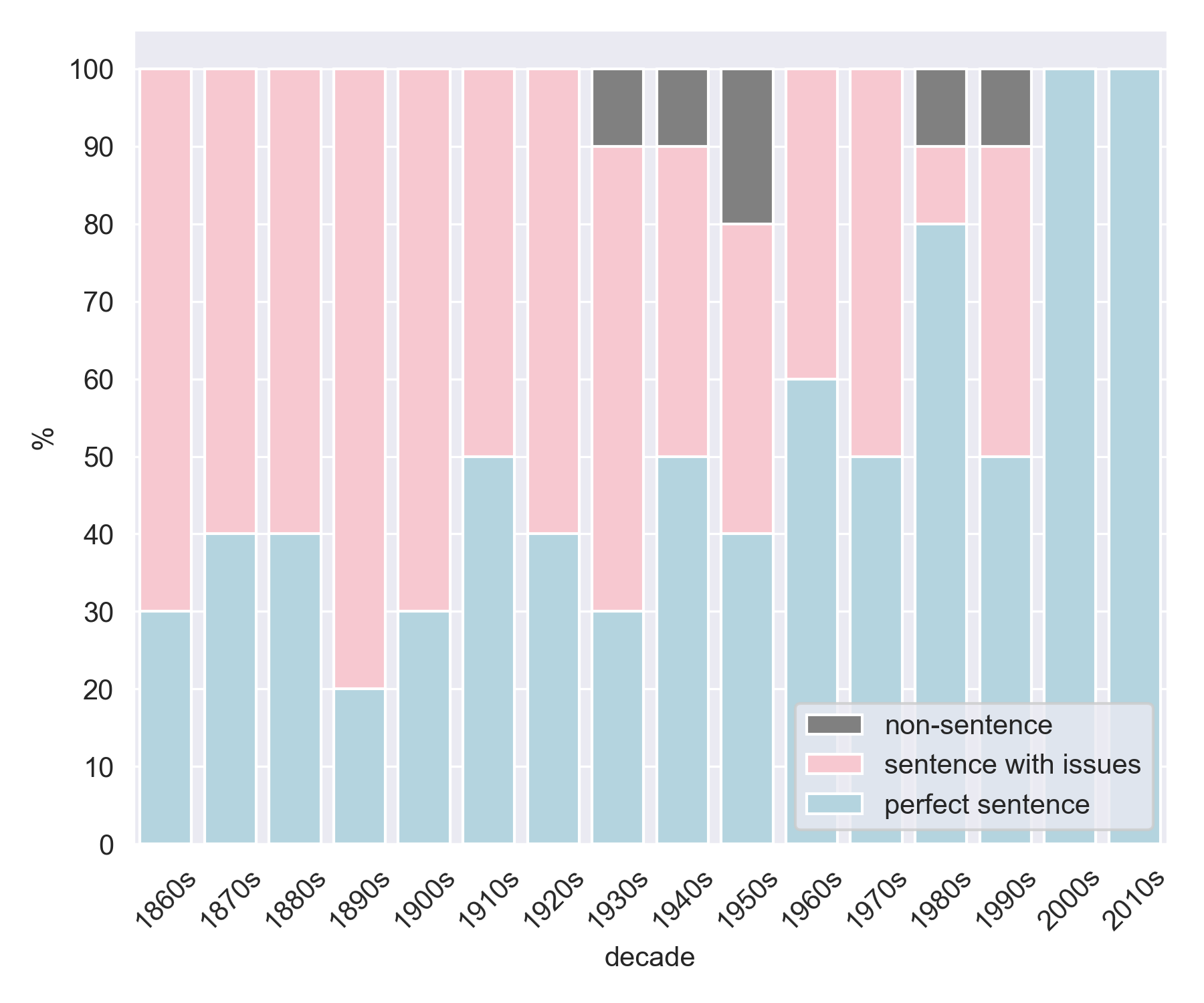}
        \caption{Deuparl}
    \end{subfigure}%
    \begin{subfigure}[t]{0.5\textwidth}
        \includegraphics[width=\linewidth]{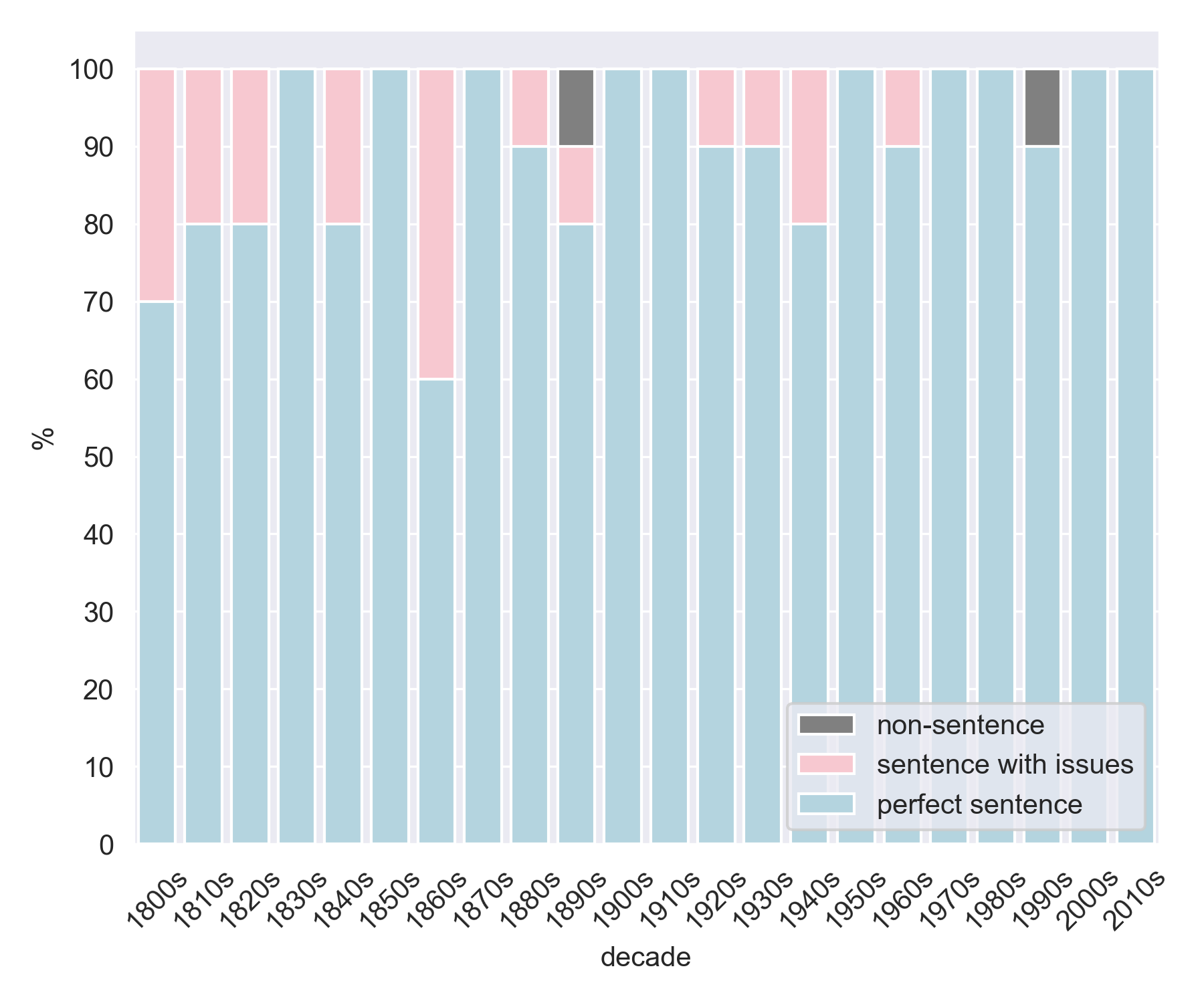}
        \caption{Hansard}
    \end{subfigure}%
    \caption{Distribution of the texts identified as perfect sentences (without any issues), sentences with issues, and non-sentences over time.}
    \label{fig:is_sent}
\end{figure}

\begin{figure}[!ht]
    \flushleft
    \begin{subfigure}[t]{0.37\textwidth}
        \includegraphics[width=\linewidth]{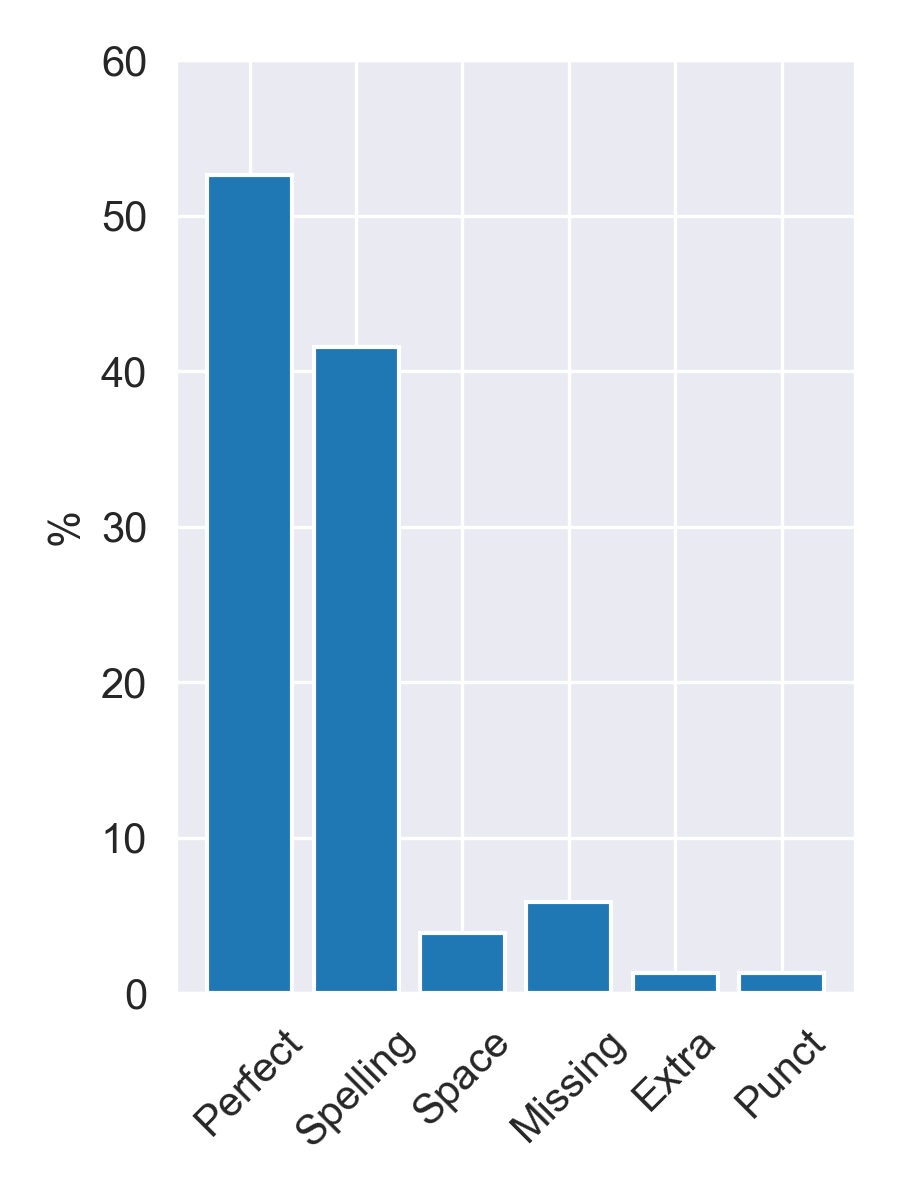}
        \caption{\textbf{Deuparl} - issues}
        \label{fig:de_issue}
    \end{subfigure}%
    \hspace{-.3cm}
    \begin{subfigure}[t]{0.25\textwidth}
        \includegraphics[width=\linewidth]{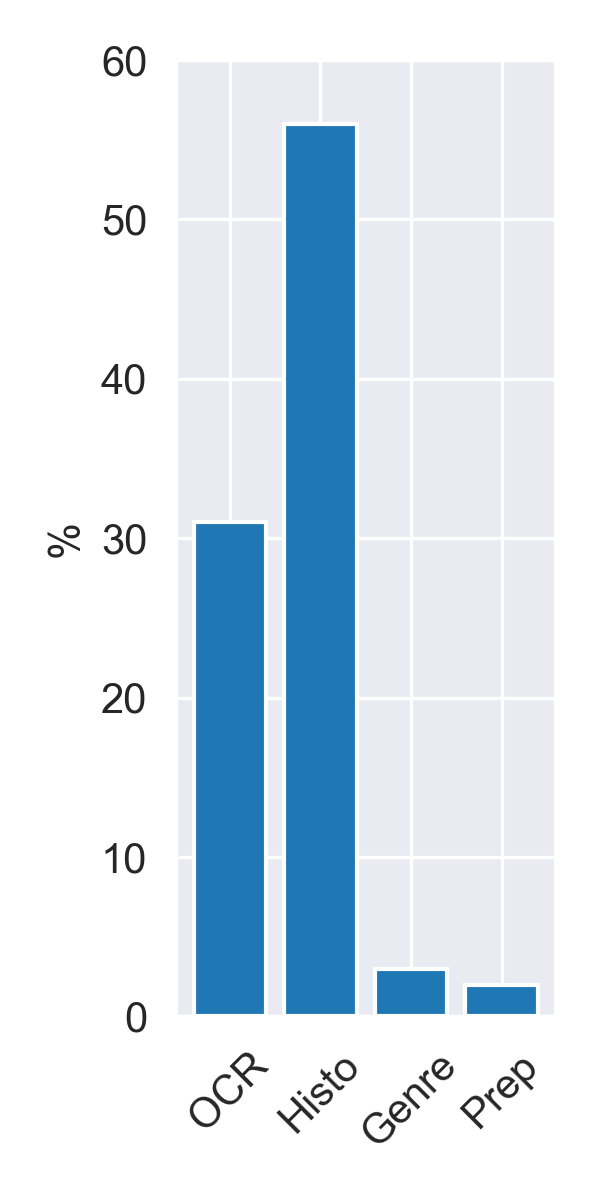}
        \caption{\textbf{Deuparl} - origins}
        \label{fig:de_origin}
    \end{subfigure}%
    \hspace{.2cm}
    \begin{subfigure}[t]{0.37\textwidth}
        \includegraphics[width=\linewidth]{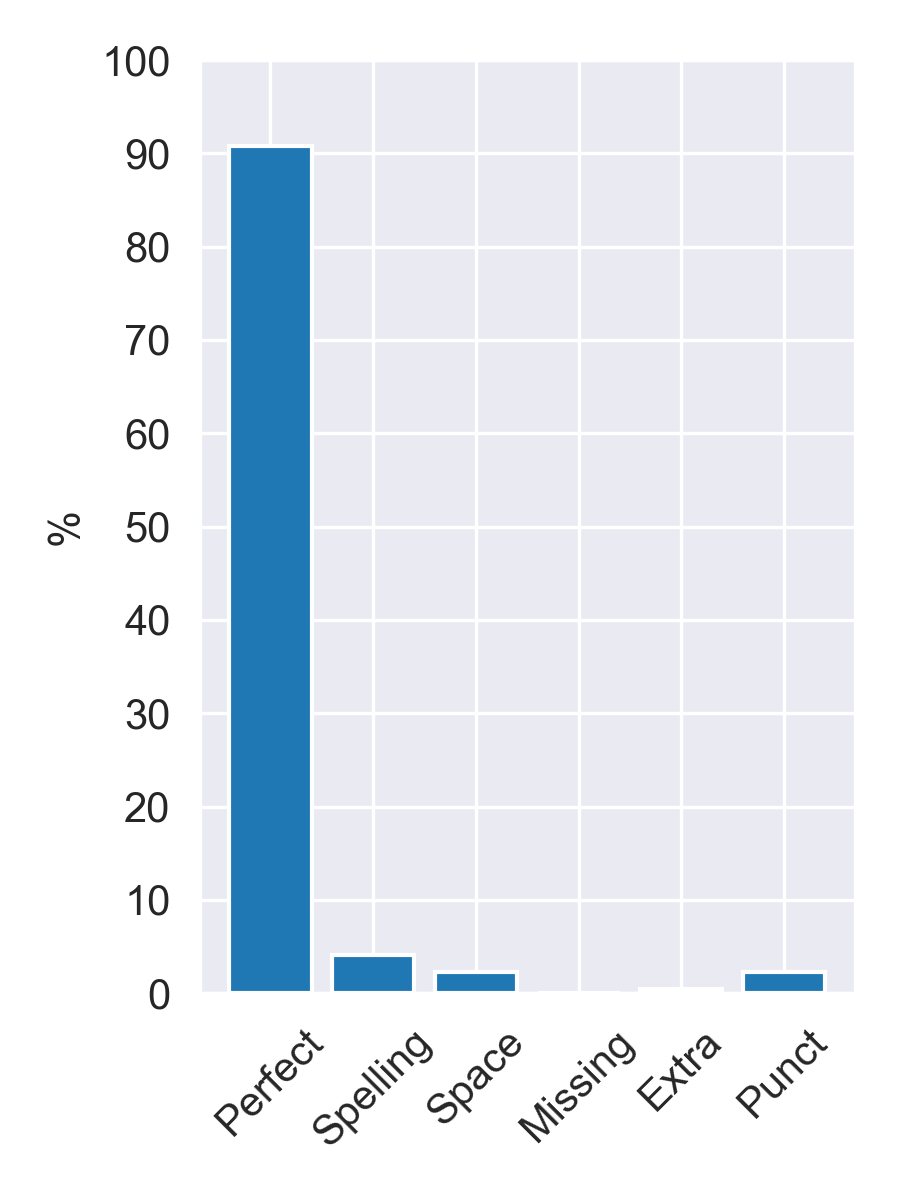}
        \caption{\textbf{Hansard} - issues}
        \label{fig:en_issue}
    \end{subfigure}%
    \caption{(a)/(c): Percentage of perfect sentences and sentences with a specific issue to all texts identified as sentences. (b): Percentage of sentences with issues from a specific origin to all sentences containing issues.}
    \label{fig:sent_quality}
    \vspace{-.3cm}
\end{figure}

\paragraph{\textbf{Results}}

We show the distribution of the texts identified as perfect sentences (without any issues), sentences with issues, and non-sentences in Figure \ref{fig:is_sent}. 
We observe that: (1) German data has more issues compared to English. For the older time periods, e.g., the first 3 decades in each corpus, 60-70\% German sentences are found to contain issues, while there are only 20-30\% English sentences 
found to be problematic. (2) Only a small portion of sentences were identified as non-sentence\ab{s}. Specifically, 6 out of 160 texts ($\sim$3.75\%) from \texttt{DeuParl} and 2 out of 220 texts ($\sim$0.1\%) from \texttt{Hansard} are found not to be sentences. 
(3) The proportion of problematic sentences is decreasing over time; for both languages, all sentences after 2000 are flawless.

Next, we present the percentages of the sentences having a specific issue in Figure\se{s} \ref{fig:de_issue} and 
\ref{fig:en_issue}. 
Note that one sentence can have multiple issues, so the sum of the percentages does not need to equal 100\%. In English data, 90\% of the sentences are perfect --- only 1\%-2\% of the sentences have \emph{Spelling}, \emph{Space}, or \emph{Punct} issues. In contrast, only half of the German sentences are issue-free, with over 40\% of the sentences having \emph{Spelling} issues and 1\%-5\% of the sentences containing other issues. Furthermore, we find that \emph{Historic} origin issues dominate in the problematic German sentences---over 55\% of those contain at least one \emph{Historic} origin issue, followed by \emph{OCR} origin---$\sim$30\% of the flawed sentences have issues raised from OCR errors. 

In summary, our annotations suggest that English data from \texttt{Hansard} is substantially less problematic than the German data from \texttt{Deuparl}. Notably, the most prevalent issues 
in the German data, 
namely historical spelling and OCR spelling error\ab{s}, 
rarely appear in the English data. 
This leads us to question whether the \texttt{Hansard} Corpus has already undergone preprocessing like spelling normalization. 
Through contact with the UK Parliament enquiry services,\footnote{\url{https://www.parliament.uk/site-information/contact-us/}} 
we received feedback indicating that the data has not been preprocessed regarding spelling normalization. As for OCR errors, a statement on their website\footnote{\url{https://hansard.parliament.uk/about}} explicitly mentions that OCR errors are curated once identified. As a consequence, 
we conclude that English has undergone smaller changes in terms of spelling in the past 200 years compared to German, \se{making German the more interesting (and much less research\ab{ed}) language from a historical perspective}.

\subsection{Dataset Construction}\label{sec:datasets}

\begin{figure}[!ht]
    \centering
    \includegraphics[width=.6\textwidth]{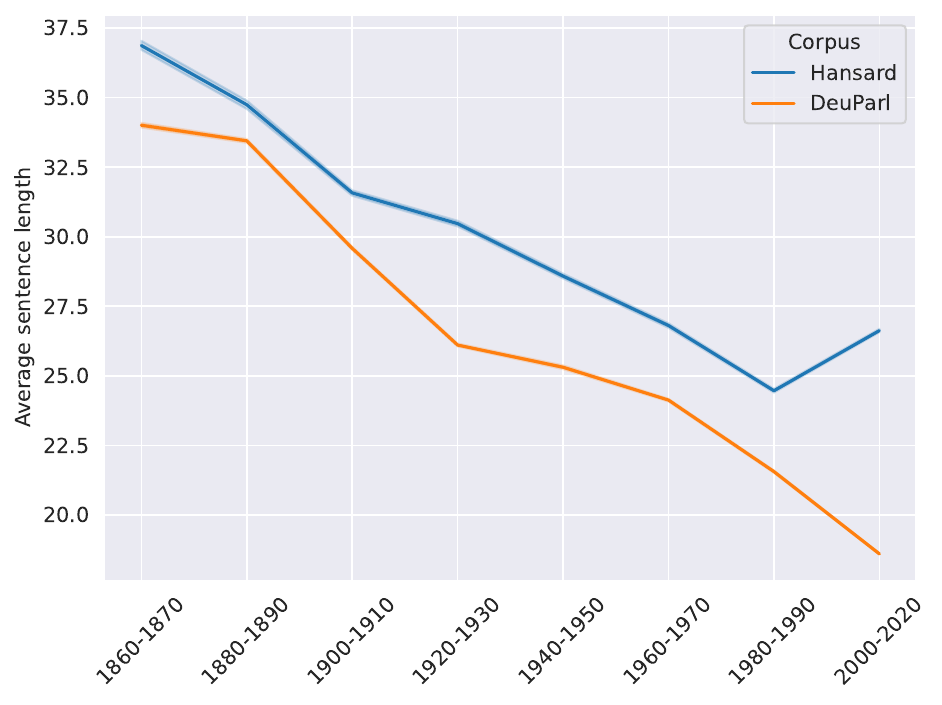}
    \caption{Average sentence length per decade group over time.}
    \label{fig:len_avg}
\end{figure}

We apply the preprocessing pipeline introduced 
above to
get up to 200k sentences for each decade and language. 
\ycj{
To make the English and German data comparable, we limit the time period for English to the decade 1860-2020.
Next, we tokenize the sentences and do part-of-speech tagging using Stanza \citep{qi-etal-2020-stanza}. The length of sentences is then the number of tokens after tokenization. 
Figure\sen{s} \ref{fig:len_dis} \ycf{(in the appendix)} and \ref{fig:len_avg} 
show the distribution of sentence lengths and the average sentence length over time.
We see 
that shorter sentences appear more often in the later periods and thus the average sentence length is overall decreasing over time for both languages. For instance, the average sentence length in DeuParl decreases from $\sim$35 to $\sim$20 
\sen{in} 
the time window 1860-1870 to 2000-2020. The only exception is the last subtrend for Hansard, where the average length increases by $\sim$2 tokens.
Metrics like \mdd{} are highly sensitive to sentence length \citep{lei2020normalized}, thus, it is important to 
\se{control for} 
sentence length when constructing the dataset for observation.
}


\subsubsection{Final Dataset}
\ycj{We select sentences with varying number of tokens for observation: 5, 10, 15, 20, 30, 40, 50, 60, and 70, covering short, middle and long sentences.
We aim to observe syntactic changes in sentences of the same length (i.e., same amount of tokens). Nevertheless, to obtain sufficient sentences for each length and time period, we set an offset of 2 tokens for sentence grouping, e.g., sentences of a length from 10 to 12 are grouped together; in addition, we consider decade groups with each spanning 
\ycf{2 decades (except for the last one, which spans 3 decades)}. 
For both English and German, the final dataset contains 450 sentences per length and decade group. These sentences are randomly sampled from a total of 32,400 preprocessed sentences for each language. 
}

 \section{Dependency Parsers}\label{sec:parser}

Unlike previous approaches, which 
solely relied on a single parser, typically the Stanford CoreNLP Parser \citep{manning-etal-2014-stanford}, we will base our observations on 
various parsers with the two most popular design choices: transition-based and graph-based.
Transition-based parsers sequentially predict the arc step by step based on the current state,
while graph-based parsers globally optimize the dependency tree, aiming to find the highest-scored one. The parsers used in this work include: (1) \textbf{transition-based}: CoreNLP \citep{manning-etal-2014-stanford}, StackPointer \citep{ma-etal-2018-stack}; (2) \textbf{graph-based}: Biaffine \citep{dozat2016deep}, Stanza \citep{qi-etal-2020-stanza}, TowerParse \citep{glavas-vulic-2021-climbing} and CRF2O \citep{zhang-etal-2020-efficient}. \ycf{Details of the parser configurations are provided in \S\ref{app:parser_setting} in the appendix.}

\subsection{Evaluation}
\par To explore the reliability of the parsers used, we evaluate them on (1) existing \textbf{UD treebanks}, (2) the treebanks built upon the two target corpora of this work (denoted as ``\textbf{Target Treebanks}''), and (3) adversarially attacked treebanks concerning 
the most severe issues identified in \S\ref{sec:data}. 

\paragraph{\textbf{Metrics}} 
We calculate the \textbf{Unlabeled Attachment Score (UAS)} and \textbf{Labeled Attachment Score (LAS)} for the parser performance. UAS and LAS are originally defined as the percentage of tokens with correctly predicted heads and head+relations respectively; in this scenario, the input of the parser is sentences pre-tokenized by humans. On the other hand, \citet{zeman-EtAl:2018:K18-2} test 
\se{parser} 
performance in a real-world scenario, i.e., parsing the raw text without any gold standard (e.g., sentence segmentation, tokenization, part-of-speech tags, etc.). 
The metrics are then re-defined as the harmonic mean (F1) of precision
P and recall R, where P is the percentage of the correct \fix{head (+relation)} to the number \fix{of} predicted tokens, while \fix{R} is the percentage of the correct \fix{head (+relation)} to the number of tokens in the gold standard. 
As we will use the parsers on 
sentences without tokenization, we argue that \emph{the re-defined metrics on raw texts are more indicative in our case}.  
For all evaluations, we adopt the evaluation script
from the CoNLL18 dependency shared task \citep{zeman-EtAl:2018:K18-2}.\footnote{\ycf{\url{https://universaldependencies.org/conll18/evaluation.html}; this script ignores the subtypes of the relations (e.g., for relation `acl:relcl' (relative clause modifier), only `acl' is considered.), but here, we consider the whole relations.}}

\subsubsection{Evaluation on UD Treebanks}
To ensure the correctness of the training, implementation, and usage of the 
parsers, we first evaluate the parsers on the test sets of the existing UD treebanks, for which we choose Universal Dependencies (UD) v2.12,\footnote{We exclude the treebank ``GUMReddit'' for English, as there are many nonsensical annotations.} 
and compare the results to 
\citet{zeman-EtAl:2018:K18-2}. Although our evaluation setup, parsers, and the version of the UD treebanks are 
different from theirs,
comparable (or better) results are expected since (1) the discrepancy in UD versions is small,\footnote{Most changes are about fixing a tiny portion of incorrect annotations, e.g., see change logs on \url{https://github.com/UniversalDependencies/UD_English-PUD/blob/master/README.md} and \url{https://github.com/UniversalDependencies/UD_German-GSD/blob/master/README.md}.}
(2) most of the tested parsers are published later than the shared task and (3) the parsers do not need to split the text into sentences by themselves here.

\paragraph{\textbf{Results}}

\begin{table}[!ht]
\centering
\begin{subtable}[t]{0.495\linewidth}
\resizebox{\linewidth}{!}{%
\begin{tabular}{@{}lcc|cc@{}}
\toprule
\multicolumn{1}{l}{} & \multicolumn{2}{c|}{UAS}                 & \multicolumn{2}{c}{LAS}                  \\
                     & UD                & TARGET               & UD                & TARGET               \\ \midrule
CoreNLP*              & 78.6±5.5          & 82.9 (+4.3)          & 73.4±6.5          & 78.7 (+5.3)          \\
Stanza*               & \underline{88.2±6.2}    & 88.0 (-0.2)          & \underline{84.9±8.0}    & 85.0 (+0.1)    \\
TowerParse*           & 87.1±4.5          & \textbf{92.8 (+5.7)} & 82.8±5.9          & \textbf{90.3 (+7.5)} \\
StackPointer         & 85.7±5.2          & 84.2 (-1.5)          & 81.3±6.2          & 80.2 (-1.1)          \\
CRF2O                & 87.2±4.6          & 86.6 (-0.6)          & 83.5±5.4          & 82.8 (-0.7)          \\
Biaffine             & \textbf{91.6±2.6} & \underline{90.1 (-1.5)}    & \textbf{88.5±3.3} & \underline{87.2 (-1.3)}          \\ \bottomrule
\end{tabular}%
}%
\caption{English}
\label{tab:parser_de}
\end{subtable}%
\hspace{.05cm}
\begin{subtable}[t]{0.495\linewidth}
\resizebox{\linewidth}{!}{%
\begin{tabular}{@{}lcc|cc@{}}
\toprule
             & \multicolumn{2}{c|}{UAS}                 & \multicolumn{2}{c}{LAS}                  \\
             & UD                & TARGET               & UD                & TARGET               \\ \midrule
CoreNLP*      & 74.0±2.9          & 74.4 (+0.4)          & 67.3±2.9          & 69.6 (+2.3)          \\
Stanza*       & 84.3±4.1          & 87.6 (+3.3)          & 78.8±4.4          & 82.0 (+3.2)          \\
TowerParse*   & \underline{86.2±2.0}    & \underline{89.7 (+3.5)}    & \underline{80.3±1.3}    & \underline{84.5 (+4.2)}    \\
StackPointer & 83.6±1.3          & 86.9 (+3.3)          & 78.2±1.6          & 80.8 (+2.6)          \\
CRF2O        & 78.5±1.5          & 81.9 (+3.4)          & 70.2±2.4          & 72.9 (+2.7)          \\
Biaffine     & \textbf{87.0±2.5} & \textbf{90.8 (+3.8)} & \textbf{81.7±1.8} & \textbf{84.5 (+2.8)} \\ \bottomrule
\end{tabular}%
}%
\caption{German}
\label{tab:parser_de}
\end{subtable}%
\vspace{-.3cm}
\caption{UAS and LAS on \texttt{\textbf{UD}} and \texttt{\textbf{Target}} treebanks. We show the macro average and the standard deviation of the metrics over the UD treebanks. The best performance in each criterion is bold and the second best one is underlined. We mark the parsers trained on mismatched data (with respect to UD versions or treebanks) with $^{*}$. \ycf{Values in brackets indicate the absolute differences.}}
\label{tab:parser_eval}
\end{table}

\ycf{In \S\ref{app:comparison_parser} in the appendix, we confirm the legitimacy of the used parsers by showing that the performance of our used parsers is mostly superior to those in \citet{zeman-EtAl:2018:K18-2}.}
Next, we show the macro average and the standard deviation of the UAS
and LAS 
over the UD treebanks in Table \ref{tab:parser_eval}
(column ``\textbf{UD}''). 
Biaffine consistently outperforms the others across different languages in terms of both UAS (en: 92\% vs. 79\%-88\%;
de: 87\% vs. 79\%-86\%) 
and LAS (en: 89\% vs. 73\%-85\%; 
de: 82\% vs. 70\%-80\%). 
Furthermore, it also exhibits the strongest robustness across various treebanks, as demonstrated by the smallest average standard deviation ($\sim$2.55 percentage points (pp)) over different metrics and languages. It is noteworthy that CoreNLP, the parser leveraged by all other related approaches, is the worst in our evaluation, according to all criteria.

Stanza 
trained on slightly mismatched data outperforms the other fine-tuned ones on average for English. Nevertheless, it is the least robust one according to its high standard deviations (6.2 pp vs. <5.5 pp in UAS; 8.0 pp vs. <6.5 pp in LAS). Interestingly, Towerparse, \yc{which was trained on mismatched data in terms of treebanks and versions,}
performs similarly to Biaffine on German treebanks,  only $\sim$1 pp lower UAS/LAS. 
However, we observe that it tends to predict multiple roots for one sentence and produce cycles in the dependency tree graphs, \yc{violating the UD dependency definition.} 
In our evaluation, it generates 440 multi-root predictions and 1,498 cycles, 
while the others rarely make such errors (only Stackpointer predicts multiple roots 28 times \fix{among the other parsers}). Besides, it also has to skip 28 sentences due to the limitation of its implementation.\footnote{See \ab{the} authors' explanation about skipping sentences in footnote 7 in \citet{glavas-vulic-2021-climbing}.}
CRF2O performs on par with the second-tier parsers on English data but ranks as the second-worst on German data. The reason may be that it is not capable of parsing non-projective relations, which 
occur more frequently in German compared to English, \yc{as shown in \citet{PhysRevE.96.062304}}. 
With its built-in evaluation function, which skips the non-projective sentences in the gold standard ($\sim$3k out of $\sim$23k sentences), the UAS and LAS raise from <80\% to >90\% on German treebanks.

\subsubsection{Evaluation on Target Treebanks}\label{sec:parser_target}
We leverage the dependency annotations 
\yc{from a 
\se{private} unpublished work \se{accessible to us}, which collects 
\se{human} annotations by manually correcting the automatic parsing results;} 
it includes \yc{annotations in UD format} for 111 sentences from Hansard and 163 from DeuParl \yc{from different time periods}. \ycj{Sentences were cleaned up before parsing.} 
\se{There were two annotators involved, who annotated independently and in case of disagreement resolved them. However, there is no agreement reported and it is unclear how reliable these human annotations are, as they correct the output from the parser, potentially leading to bias in correction (``priming effect'').}

\paragraph{\textbf{Results}}
We show the evaluation results on the Target treebank in Table \ref{tab:parser_eval} (column ``\textbf{Target}''). The values in the brackets indicate the absolute difference in metrics between the Target treebank and the UD treebanks. We observe that the Target treebank mostly does not negatively 
affect the parsers' performance. All parsers exhibit improved performance on the \textbf{German} Target treebank compared to the UD treebank, with an increase ranging from 0.4 to 4.2 pp. For \textbf{English}, the maximal performance drop is observed with StackPointer and Biaffine (1.1-1.5 pp), while CoreNLP and TowerParse achieve better results, with an increase ranging from 4.3 to 7.5 pp. Stanza and CRF2O display only minor fluctuations (<1 pp). 
Hence, our results 
\ycj{weakly support the parsers' proficiency in the target domain, considering potential bias in human annotations.}

\subsubsection{Evaluation on Adversarial Treebanks}\label{sec:parser_adv}
\yc{To inspect the effect of data noise on the parsers, we generate two adversarial datasets concerning the most two prevalent issues in German data identified in \S\ref{sec:data}, }
i.e., historical spelling and OCR-raised spelling errors. We perturb the merged German test set to create the adversarial treebanks. 
To perform the historical spelling attack, we chose 23 candidate words from our previous annotations and replaced them with their historical spellings in the test set. This resulted in 1,971 sentences that were affected by the attack. 
For (2), we randomly replace 10\%, 30\%, or 50\% of the characters in 1 or 2 tokens within a sentence with random characters, resulting in 6 attacking levels; we produce 2000 sentences per attack level. 

\paragraph{\textbf{Results}}

We illustrate the absolute difference in metrics between the sentences before and after attacking in Figure \ref{fig:attack} \ycf{in the appendix}. Both types of attacks consistently have a greater influence on LAS compared to UAS. This can be explained by \se{the fact} that LAS is more sensitive to perturbations because it considers both attachment and labeling accuracy, whereas UAS only considers attachment accuracy. Even if the attachment remains correct, changes to the words themselves (such as spelling or character replacement) can result in incorrect labels. As depicted in Figure \ref{fig:attack_his}, \textbf{historical spelling} has a slight negative impact on the parsers' performance, with a difference of less than 0.07 \ycf{pp} in LAS and that of up to around 0.05 \ycf{pp} in UAS. Among the parsers, Biaffine is the most stable one, followed by TowerParse; CoreNLP exhibits the lowest robustness. 

We present the results for \textbf{OCR spelling attack} in Figure \ref{fig:attack_ocr}. OCR spelling errors have a slightly larger impact, degrading 
\se{parser} 
performance up to around 0.12 \ycf{pp} in LAS and 0.09 \ycf{pp} in UAS, when 50\% characters of 2 tokens within a sentence are perturbed. Again, Biaffine is the most robust 
against OCR attack, followed by CRF2O and TowerParse, which demonstrate similar sensitivity to Biaffine when the attack level is high. At the highest attack level, CoreNLP, StackPointer, and Stanza show similar sensitivity. However, at low attack levels, StackPointer and Stanza demonstrate better robustness compared to CoreNLP. In real-world cases, OCR spelling errors often involve only one character within a sentence. For example, in the sentence ``...
wollen \emph{Tie} nun das letzte nehmen.'' (from our annotation in \S\ref{sec:data}), the only OCR error is in ``\emph{Tie}'', which should be corrected to ``\emph{Sie}''. Hence, Stanza and StackPointer are still superior to CoreNLP in real-world scenarios. 




\subsection{Summary}
\ycf{All parsers exhibit robustness and show minor (or no) degradation both under adversarial attacks and the target data.}
Among 
\ycf{them}, Biaffine shows the best performance in standard evaluations as well as the highest robustness under adversarial conditions. CoreNLP and CRF2O are often the worst ones in our evaluation;
besides,
they are unable to predict crossing edges, 
which we deem as an important indicator of language change. 
Non-projective sentences seem to substantially degrade CRF2O's performance; thus, we exclude it from further experiments. CoreNLP is retained, however, for comparison purposes. Therefore, \emph{we will use five parsers to perform the large-scale analysis in \S\ref{sec:analysis}: 
CoreNLP, StackPointer, Biaffine, TowerParse, and Stanza.}
\section{Metrics}\label{sec:measure}

Unlike most of the relevant approaches that only focus on the linear dependency distance (e.g., \citet{lei2020dependency,liu2022dependency,zhu2022investigating,zhang2023investigation}), we examine a large array of metrics \wz{that capture statistical patterns driven by graph theory or linguistic phenomena. In the following, we outline the definition and usage of these metrics over a German sentence---``Das hat alles sehr gut geklappt!'' (extracted from the UD GSD treebank).}

\begin{figure}[!ht]
    \centering
    \includegraphics[width=\textwidth]{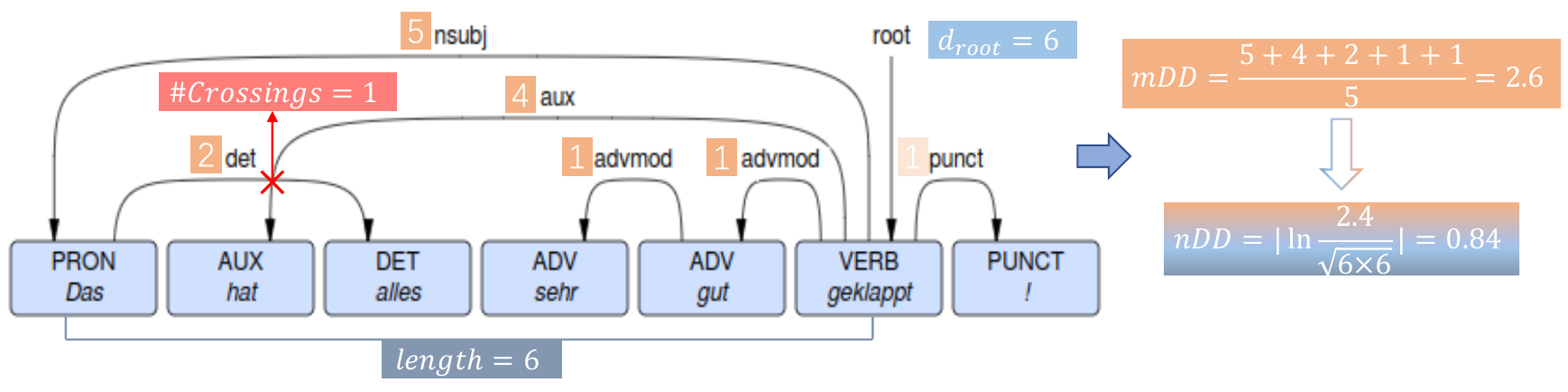}
    \caption{Illustration of observing \textbf{\protect\rootdis{}} and  \textbf{\protect\crossing{}}, along with calculating \textbf{\protect\ndd{}} and \textbf{\protect\mdd{}} using equation \ref{eq:mdd} and \ref{eq:ndd}, with the example sentence ``Das hat alles sehr gut geklappt!''. Arrows point from head to dependent tokens.
    Values with an \textcolor{orange}{orange} background indicate the dependency distance between a dependent-head pair.}
    \label{fig:example_linear}
\end{figure}

    \paragraph{\textbf{Root Distance (\protect\rootdis{})}} 
    \wz{\rootdis{} is 
    \ycf{the index of the word connecting to the pseudo root in a sentence.}
    An example is shown in Figure \ref{fig:example_linear}, 
    where \rootdis{} is 6 (from root to `geklappt'). \citet{lei2020normalized} consider \rootdis{} as a normalizing factor when computing the mean of dependency distances for all dependency pairs in a sentence. \citet{liu2022dependency} showed that \rootdis{} correlates positively with dependency distance over time in English; 
    specifically, \rootdis{} increases over time for long sentences while decreasing over time for short sentences.}
    \paragraph{\textbf{Mean Dependency Distance (\protect\mdd{})} \& \textbf{Normalized Mean Dependency Distance (\protect\ndd{})}}
    \wz{Following \citet{lei2020dependency,zhang2023investigation, lei2020normalized}, we consider two statistical features of dependency distances, namely \mdd{} and \ndd{} given by:}
    \begin{align}
        \mdd{} &= \frac{1}{n} \sum_{(i, j) \in D} |i - j|\label{eq:mdd} \\
        \ndd{} &= \left| \log\left(\frac{mDD}{\sqrt{d_{root} \times 
        \text{{length}}
        }}\right) \right|
        \label{eq:ndd}
    \end{align}
\wz{where $|i-j|$ represents the absolute difference between the position indices of a dependency pair $(i,j)$, $D$ is a set of dependency pairs in a sentence (punctuation \ycf{and root dependencies} are excluded), 
length 
is the number of words in a sentence, and $n$ is the size $D$. In Figure \ref{fig:example_linear}, \mdd{} and \ndd{} are \ycf{2.6} and 0.84 for example.}

    \paragraph{\textbf{Number of Crossings (\protect\crossing{})}}
    \wz{A crossing is said to be observed when two dependency relations overlap. \citet{ferrer2016crossings} show 
    that \crossing{}  correlates positively with dependency distance in 21 out of 30 languages, while \citet{liu2017dependency} 
    \se{find} 
    that machine-generated artificial texts without crossings produce shorter dependency distances compared to those with crossings. 
    \se{We count the number of crossings in a sentence, ignoring punctuation\ycf{s}.}
    An example is shown in Figure 8: the dependency relation between ``Das'' and ``alles'' crosses over the relation between ``geklappt'' and ``hat'', i.e., $\crossing{}=1$.}

\begin{figure}[!ht]
    \centering
    \includegraphics[width=\textwidth]{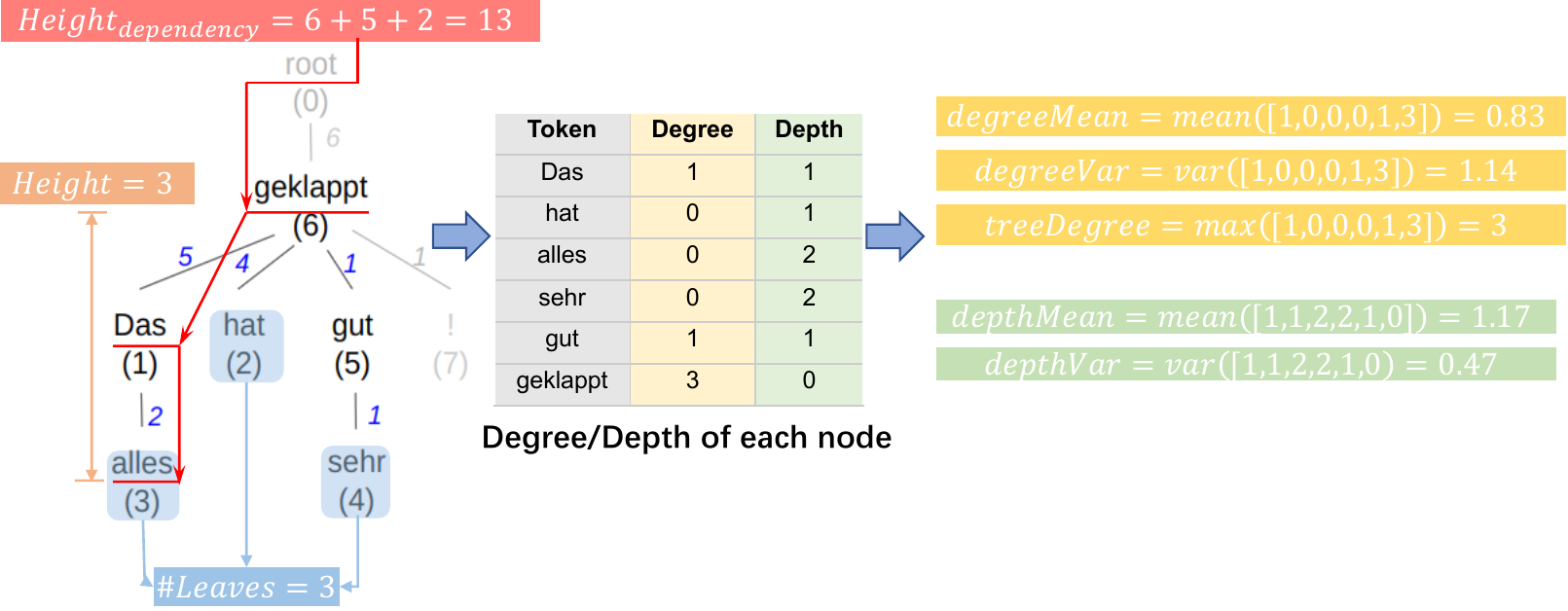}
    \caption{Illustration of calculating/observing of  \textbf{\protect\height{}}, \textbf{\protect\lpath{}}, \textbf{\protect\leaves{}}, \textbf{\protect\degree{}}, \textbf{\protect\degmean{}}, \textbf{\protect\degvar{}}, \textbf{\protect\depmean{}} and \textbf{\protect\depvar{}} with the example
sentence “Das hat alles sehr gut geklappt!”. Values in brackets are the position indices of the tokens. Values beside edges indicate the dependency distance of the corresponding dependency pairs.}
    \label{fig:example_tree}
\end{figure}

\paragraph{\textbf{Number of leaves (\protect\leaves{})}} 
\wz{We count the number of leaves (\leaves{}) of a dependency tree, excluding punctuations---see an example in Figure \ref{fig:example_tree} (left).}
    
    \paragraph{\textbf{Tree Height (\protect\height{})} \& \textbf{Longest Path Distance (\protect\lpath{})}} 
    \wz{\height{} is 
    the longest-path distance from the root node to a leaf, while \lpath{} 
    is the weighted longest-path distance where an edge over two nodes is weighted by the dependence distance between the nodes. 
    An example can be observed in Figure \ref{fig:example_tree} (left): \height{} $=3$, \lpath{} $=13$.}

    \paragraph{\textbf{\textbf{Depth Variance (\protect\depvar{}) \& Depth Mean (\protect\depmean{})}}}
    \wz{We denote the depth of a node as the path length from the node to the root. \depmean{} and \depvar{} represent the mean and variance of the depths of all nodes in a tree, as illustrated in Figure \ref{fig:example_tree} (middle).}

    \paragraph{\textbf{Tree Degree (\protect\degree) \& Degree Variance (\protect\degvar{}) \& Degree Mean (\protect\degmean{})}} 
    \wz{We 
    \se{measure} 
    the degree of a node as the number of outgoing nodes attached to that node, i.e., by counting the dependents of a head word in a dependency tree. The higher the degree of a node, the more dependents a head word has. 
    This metric is related to dependency distance—as both long dependency distances and head words with high degrees can complicate the syntactic structure.
    We also consider statistical features: the mean (\degmean{}) and variance (\degvar{}) of the degrees of all nodes, as well as the maximum of all nodes' degrees (\degree{}).} 
    \begin{figure}[!ht]
        \centering
        \begin{subfigure}[t]{0.54\linewidth}
            \includegraphics[width=\textwidth]{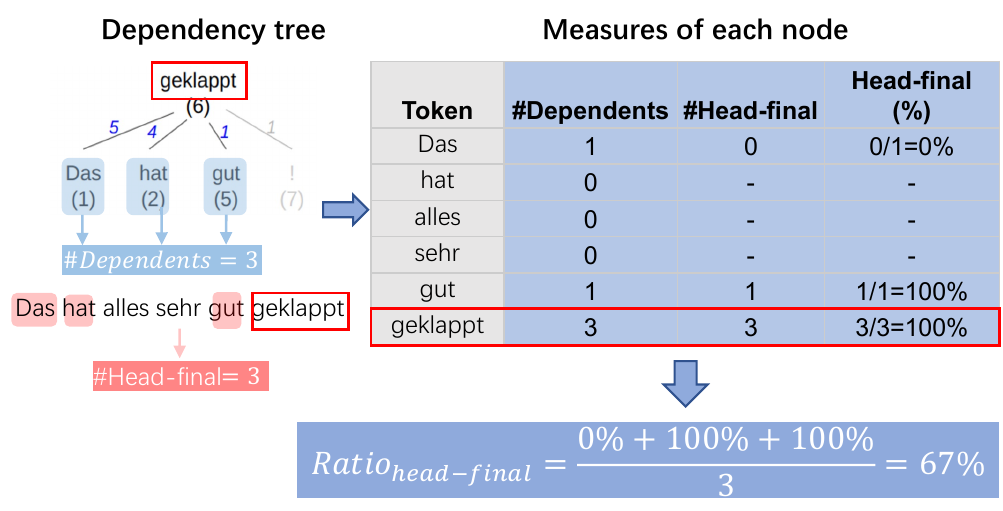}
            \caption{\protect\headratio{}}\label{fig:headratio}
        \end{subfigure}
        \begin{subfigure}[t]{0.45\linewidth}
            \includegraphics[width=\textwidth]{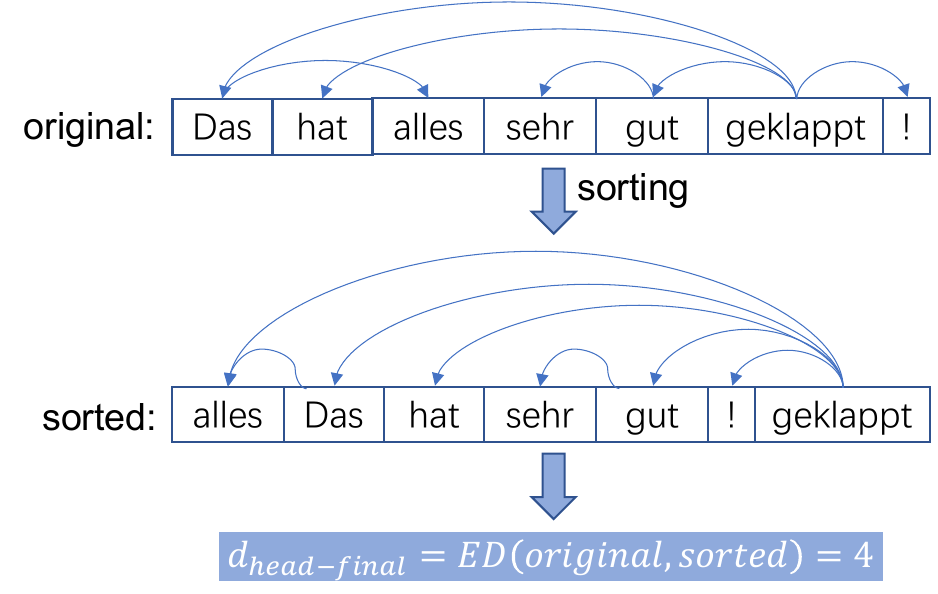}
            \caption{\protect\headdis{}}\label{fig:headdis}
        \end{subfigure}
        \caption{Illustration of calculating/observing \textbf{\protect\headratio{}} and \textbf{\protect\headdis{}} with the example
sentence “Das hat alles sehr gut geklappt!”. Arrows point from head to dependent tokens.}
    \end{figure}
    
    \paragraph{\textbf{Head-final Ratio (\protect\headratio{}) \& Head-final Distance (\protect\headdis{})}}
    \wz{For a dependency pair, if a head follows its dependent, then the pair is termed as a head-final pair. Otherwise, if a head precedes its dependents, then the pair is considered head-initial. In linguistics, this head directionality is profiled as a crucial parameter for classifying languages \citep{LIU20101567}. 
    In Figure \ref{fig:headratio}, 
    \ycf{for each head in a sentence, we calculate the percentage of its head-final pairs to all pairs with that head (head-final(\%)), and then average the head-final(\%) over all heads in that sentence,}
    termed as \headratio{}. We note that the sentence ``Das hat alles sehr gut geklappt!'' not only includes head-final dependency pairs but also head-initial pairs, e.g., the head `Das' precedes its dependent `alles.' Here, we consider how distant this sentence is, with a mix of head-initial and head-final pairs, compared to the same sentence but permuted to have only head-final pairs. This results in the metric \protect\headdis{} that is 
    the Levenshtein edit distance \citep{Levenshtein1965BinaryCC}
    between the original and permuted sentences.} 

    \paragraph{\textbf{Random Tree Distance (\protect\random{})}}
    \wz{We randomly permute the dependency tree structure in a sentence and then compute the tree edit distance between the original and permuted trees using the algorithm of \citet{zhang1989}.}

\subsection{\textbf{Sensitivity of metrics to data noise}}\label{sec:analysis_noise}
\ycf{While our results in \S\ref{sec:parser} indicate that the parsers perform decently on the target corpora, it remains uncertain whether the metrics derived from their parsing results are affected by data noise and to what extent. Since we will inspect the 
trends of the metrics over time, it is important to ensure that the data noise does not disrupt the rankings of those metrics.
In \S\ref{app:analysis_noise} in the appendix, we demonstrate the insensitivity of our metrics to data noise by showing that the metrics for the clean data are highly correlated to those for the noisy data (overall 0.926 Spearman).}

\section{Analysis}\label{sec:analysis}

Following \citet{zhu2022investigating,liu2022dependency}, we leverage the Mann Kendall (MK) trend test \citep{mann1945nonparametric} to indicate the diachronic change of the concerned metrics. 
\se{The} \se{MK}
trend test is widely used to detect trends in time series. It has three outputs: ``increasing'', ``decreasing'', and ``no trend''; an increasing or decreasing trend is detected only if it is significant with a p-value $<0.05$.
We perform it for the 9 sentence lengths and 15 metrics, totalling $15\times9=135$ test cases. 

We first explore the dependence of the trends on parsers in \S\ref{sec:anaysis_agree}, as the relevant approaches rely on a single parser (e.g., \citet{zhu2022investigating,liu2022dependency}). 
\se{Then,} 
we compare the syntactic changes between English and German in the remaining sections. Specifically, we inspect the overall similarity of syntactic changes between English and German in \S\ref{sec:analysis_like}, with a focus on sentence length in \S\ref{sec:analysis_len} and a particular metric in \S\ref{sec:analysis_measure}. Finally, we analyze
different and similar trends 
between the two languages 
in \S\ref{sec:analysis_diff} and \S\ref{sec:analysis_similarity} respectively.

\subsection{What is the dependence of language change trends on parsers?}\label{sec:anaysis_agree}

\begin{figure}[!ht]
\vspace{-.5cm}
    \centering
    \begin{subfigure}[t]{0.48\linewidth}
        \includegraphics[width=\linewidth]{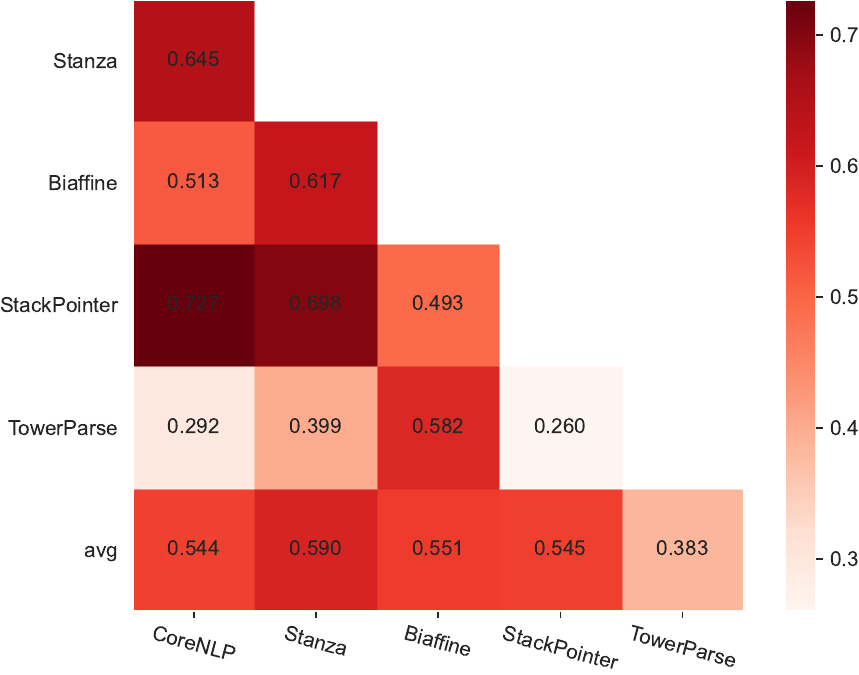}
        \caption{English}
        \label{fig:parser_cor_en}
    \end{subfigure}\hspace{.1cm}
    \begin{subfigure}[t]{0.48\linewidth}
        \includegraphics[width=\linewidth]{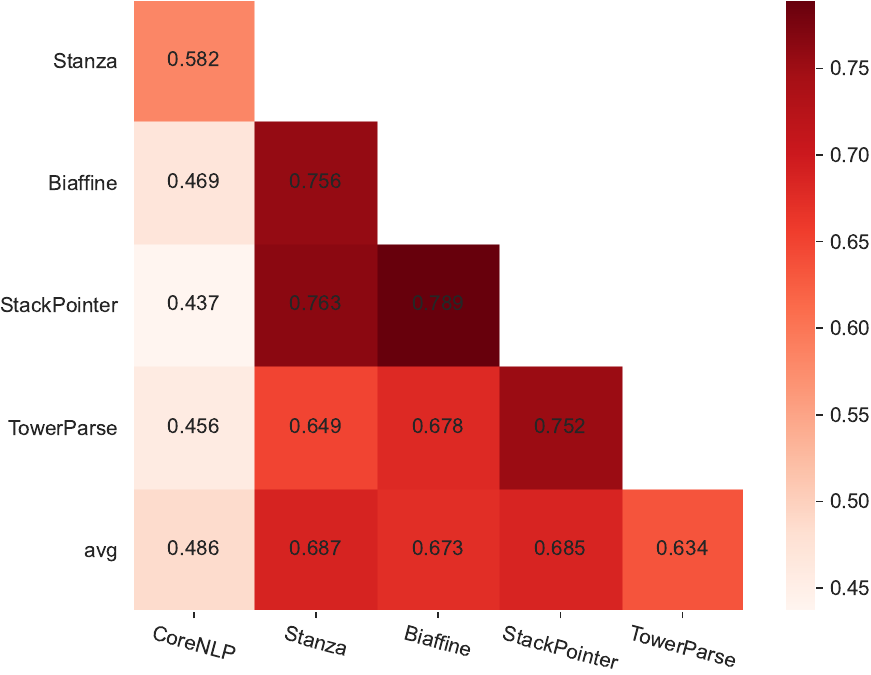}
        \caption{German}
        \label{fig:parser_cor_de}
    \end{subfigure}%
    \caption{Cohen's Kappa of MK results based on the dependency relations predicted by different parsers. Deeper colors denote higher agreement and vice versa. We show the average agreement for each parser in the last rows.}
    \label{fig:parser_cor}
    \vspace{-.3cm}
\end{figure}

The outputs of MK can be seen as the labels of a three-way classification task.  
Therefore, we calculate Cohen's Kappa \citep{cohen1960coefficient} to indicate the agreement between each parser pair. The values in Figure \ref{fig:parser_cor}
 are derived by calculating Cohen’s Kappa across the 135 MK outputs based on different parsers for each language. We then average the agreements over all parser pairs containing a certain parser to achieve the average agreement for each parser, shown in the last rows in Figure \ref{fig:parser_cor}.

Most parsers demonstrate a moderate agreement with others (0.4-0.6). On \textbf{English} data, as illustrated in Figure \ref{fig:parser_cor_en}, the highest agreement of \fix{0.73} is observed between CoreNLP and StackPointer, succeeded by Stanza and StackPointer (0.70).
Among the parser pairs, TowerParse and StackPointer obtain the lowest agreement of 0.26. Stanza obtains the highest average agreement with others (0.59), whereas TowerParse has the lowest one (0.38).  
For \textbf{German} (Figure \ref{fig:parser_cor_de}), 
the highest agreement is \fix{achieved between Biaffine and StackPointer (0.79), followed by Stanza and StackPointer (0.76); 
Stanza again
obtains the highest average agreement (0.69), while CoreNLP least agrees with others on German data (0.44-0.58).} 
In summary, our observations indicate that for both languages, \textbf{\emph{the use of different parsers may 
lead to various conclusions about the
diachronic trends of those metrics}}, despite their decent performance observed in \S\ref{sec:parser}.
This raises
concerns about previous approaches 
which solely rely on one parser. 

\begin{figure}[!ht]
    \centering
    \includegraphics[width=\textwidth]{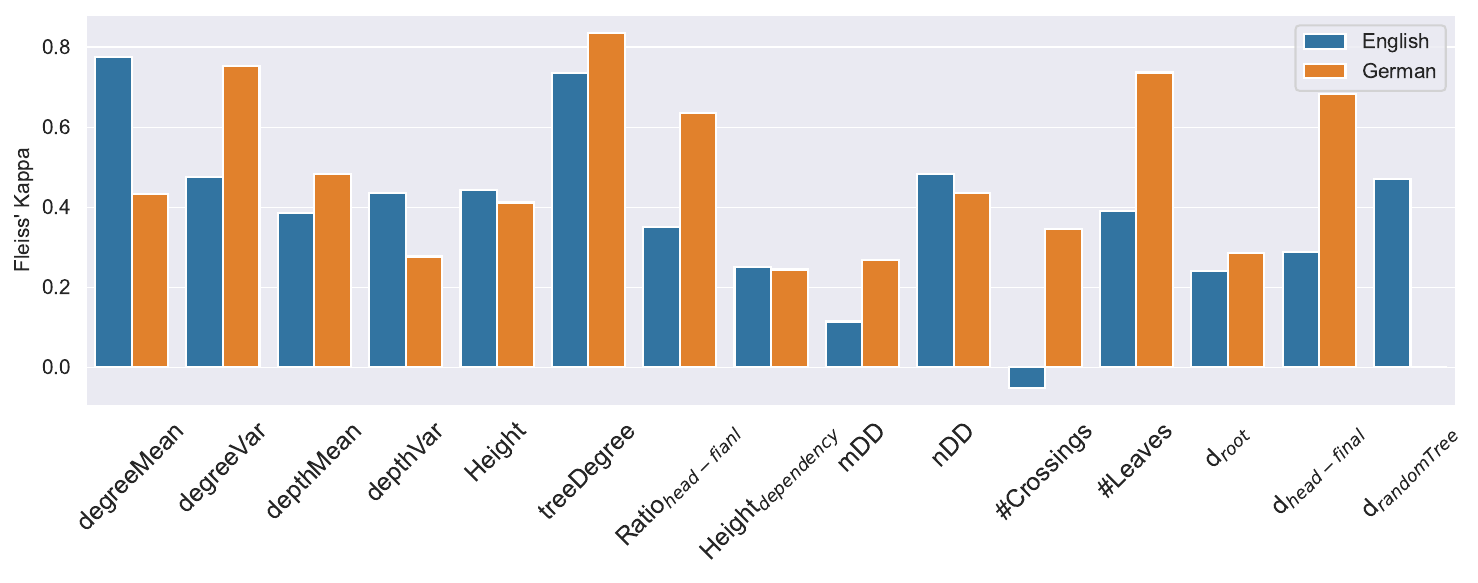}
    \caption{Fleiss' Kappa of the parsers on each metric.}
    \label{fig:fleiss}
\end{figure}

\ycj{Next, we examine \se{the} parsers' agreement on 
\se{a} 
trend for each metric. We regard the MK result based on each parser as an `annotation' for the corresponding trend; thus, for each trend, we have 5 different `annotations', allowing for calculating the Fleiss' Kappa \citep{fleiss1971measuring} 
among them. For each metric, we have 9 `annotation' instances, resulting from the 9 sentence lengths. As Figure \ref{fig:fleiss} shows, most values range from $\sim$0.2 to $\sim$0.5, implying a consistently low level of agreement among the parsers across the metrics. \degree{} is the only metric where the parsers obtain a high agreement across both languages ($\sim$0.8 Fleiss' Kappa). Among the metrics, parsers show the second least agreement on trends in \mdd{} for English ($\sim$0.1 Fleiss' Kappa). This \sen{further} questions the previous works 
in reporting diachronic trends 
\ycf{of dependency distance} based on a single parser, as results may change when using a different parser.
}

As a result, we employ a \textbf{majority vote} approach to identify the most likely stable trends in the metrics\se{:} 
if the increasing/decreasing trends can be detected based on
at least 3 out of 5 parsers, we consider them valid; an exception is made for \crossing{}, where we relax the threshold to 2 out of 5 due to CoreNLP's inability to predict crossing dependencies. The final trends for the 135 cases are listed in Table \ref{tab:trend} in the appendix, which we use for further analysis.

\subsection{Are syntactic changes more alike or distinct in English and German?}\label{sec:analysis_like}

\begin{figure}[!ht]
    \centering
    \centering
        \includegraphics[width=.7\textwidth]{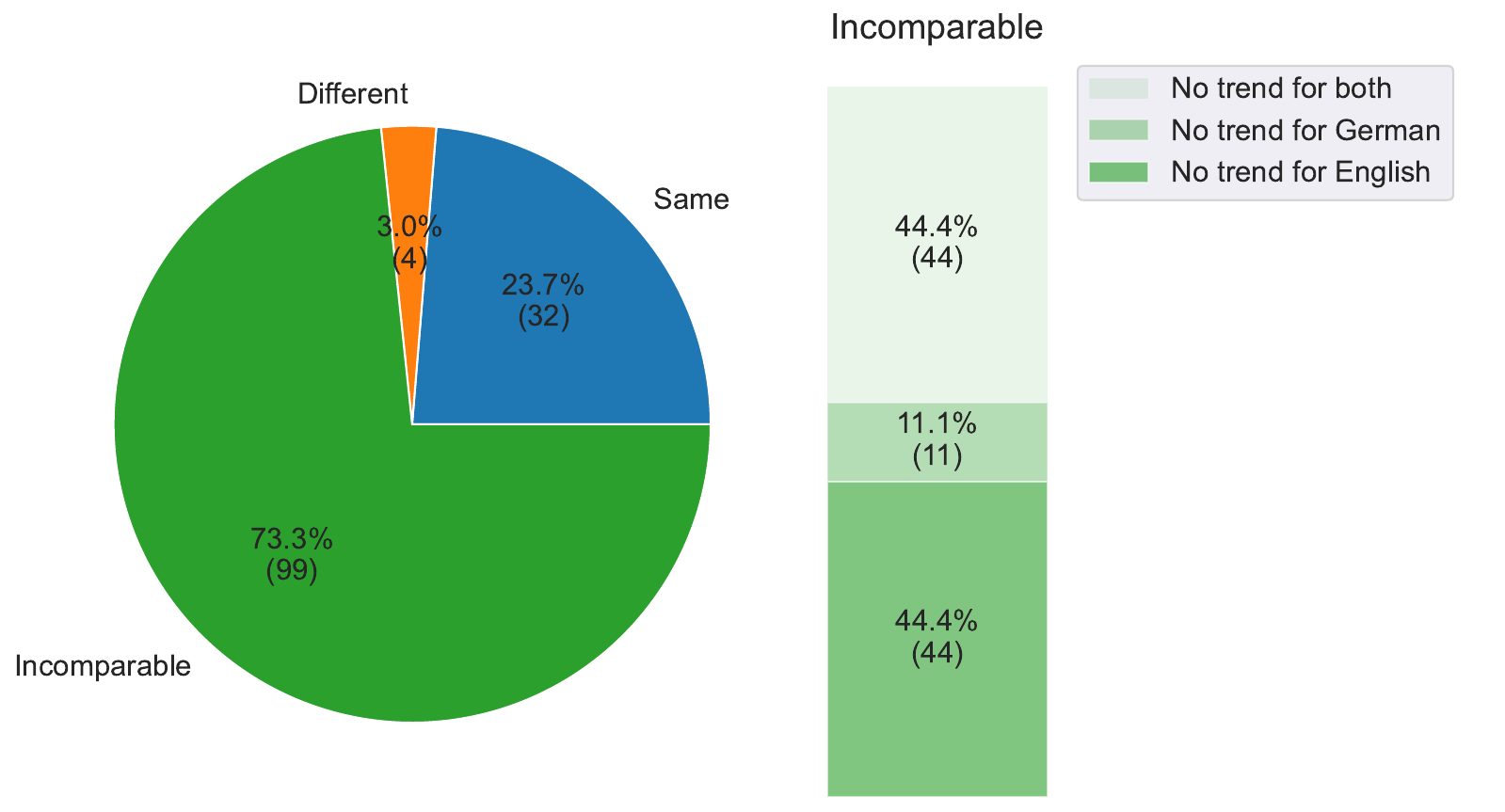}
    \caption{Left: percentages of the same, different and incomparable trends between English and German. Right: Percentages representing incomparable trends due to `no trend for English,' `no trend for German,' and `no trend for both' to all incomparable trends. \ycf{Values in brackets indicate the counts of trends.}}\label{fig:trend_perc}
\end{figure}


\vspace{-.3cm}
We compare the 135 trends 
between English and German
for the 15 metrics and 9 sentence lengths.
As shown in Figure \ref{fig:trend_perc} \ycj{(left)}, among all the trends, 99 cases (73.3\%) are incomparable
since there are no significant trends for German or English or both; 
32 cases (23.7\%) have the same trends, while only 4 cases (3\%) show different trends, which may suggest a more uniform \sen{(``convergence'')}, rather than diverse, syntactic change between English and German. 
\ycj{Among the incomparable trends, as shown in Figure \ref{fig:trend_perc} (right), 44 cases (44.4\%) are the result of no significant trends for both English and German, 11 cases (11.1\%) lack significant trends for German, and 44 cases (44.4\%) have no significant trends for English.} \se{From this perspective, the German language has undergone more significant syntactic changes compared to English.}

\vspace{-.2cm}
\subsection{Which sentence lengths exhibit the most distinct or similar syntactic changes between English and German?}\label{sec:analysis_len}


\begin{figure}[!ht]
    \centering
        \includegraphics[width=.7\textwidth]{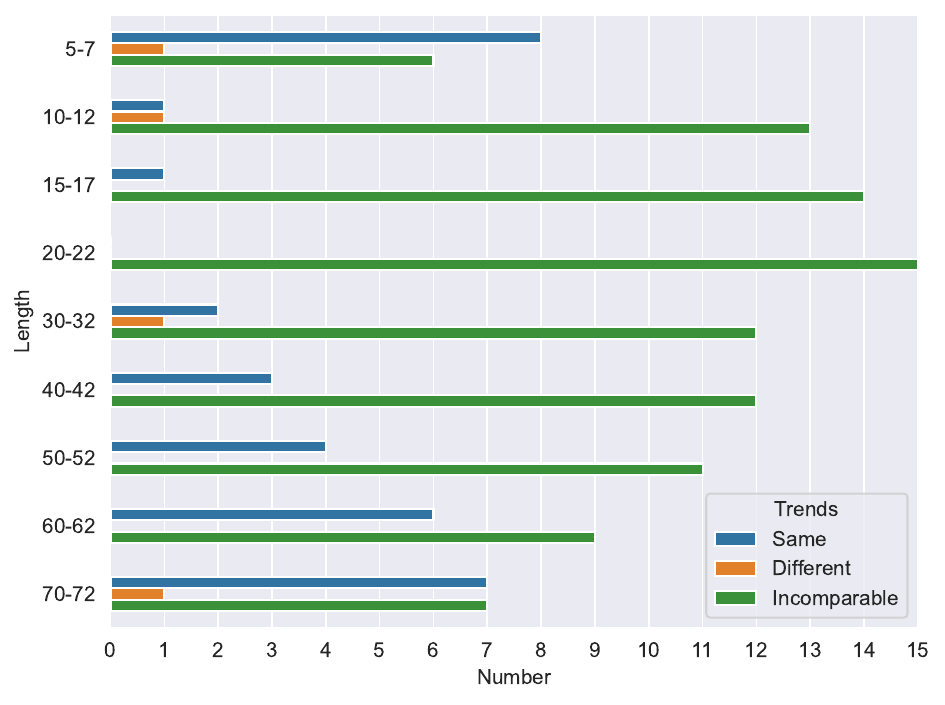}
    \caption{Number of same/different/incomparable trends for each \textbf{sentence length}. Larger numbers indicate more metrics exhibiting the same/different/incomparable trends between English and German for a specific sentence length group.}\label{fig:trend_count_len}
\end{figure}

We count the number of metrics showing different/same/incomparable trends between English and German for each sentence length, displayed in Figure \ref{fig:trend_count_len}. We observe that: (1) for sentences of lengths 5-7, 10-12, 30-32 and 70-72, 1 out of 15 metrics exhibits different trends, whereas there are no different trends for the other lengths; (2) \emph{sentences of length 5-7 behave most similarly}, shown with the highest number of metrics with the same trends (8 out of 15) among sentence lengths, followed by the longest sentences of length 70-72, which have 7 out of 15 metrics with the same trends; (3) the number of metrics with the same trends increases from 2 to 7 as sentence lengths go from 30-32 to 70-72.
\ycj{We also note that the distribution of the incomparable trends (green bars) seem to follow a Gaussian distribution, where there are more incomparable trends for the 
sentences of middle lengths (e.g., 15-22), while less incomparable trends exist for the shortest and longest sentences. Interestingly,
as we show in Figure \ref{fig:len_dis}, sentences of middle lengths dominate in the corpora; this may suggest that 
syntactic changes are more likely to occur in sentences of extreme lengths (i.e., longer and shorter sentences) rather than in sentences of common lengths.
}



\subsection{Which metrics exhibit the most distinct or similar trends between English and German?}\label{sec:analysis_measure}

\begin{figure}[!h]
    \centering
     \includegraphics[width=.8\textwidth]{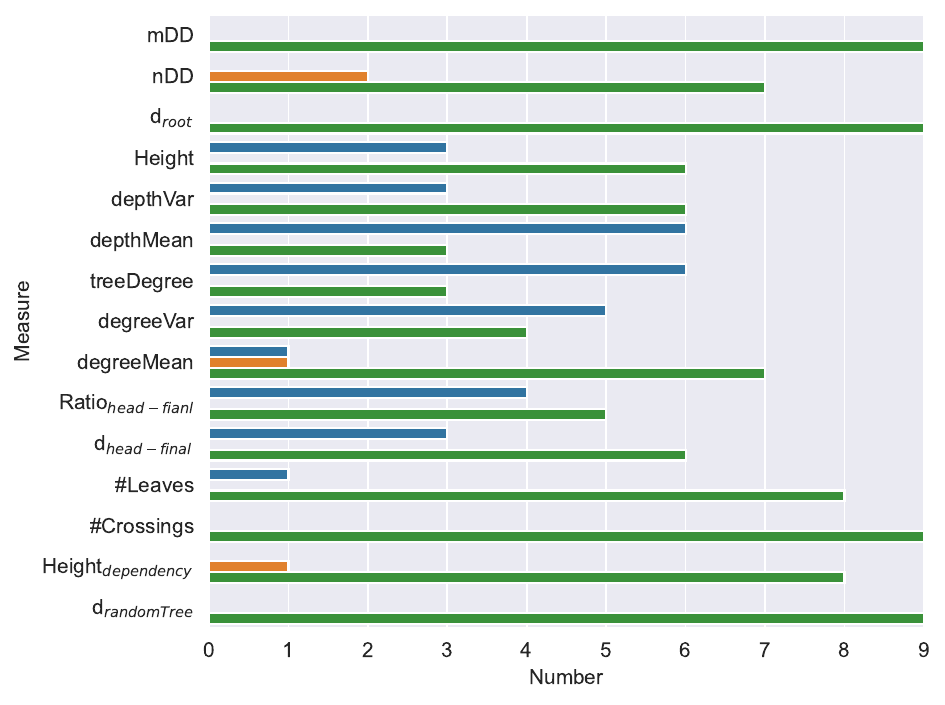}
        \caption{Number of same/different/incomparable trends for each \textbf{metric}. Larger numbers indicate more sentence lengths exhibiting the same/different/incomparable trends between English and German for a specific metric.}\label{fig:trend_count_measure}
\end{figure}

We count the number of sentence lengths showing different/same trends between English and German for each metric, visualized in Figure \ref{fig:trend_count_measure}. Only one metric shows both same and different trends for various sentence lengths, namely \degmean{}.
\emph{\ndd{} behaves most differently}---for 2 out of 9 lengths, it exhibits different trends---followed by \degmean{} and \lpath{}, whose trends are different for one length. \emph{\depmean{} and \degree{} are the most similar ones regarding changes in English and German}; for 7 out of 9 sentence lengths, they exhibit the same trends between English and German. \leaves{} and \degmean{} have the same trend only for one length. Among the others, \height{}, \depvar{}, \degvar{}, \headdis{} and \headratio{} behave similarly for 3-5 out of 9 sentence lengths.

\subsection{What are the differences in syntactic changes between English and German?}\label{sec:analysis_diff}
\begin{figure}[!ht]
    \centering
    \begin{subfigure}[t]{.4\linewidth}
    \centering
\includegraphics[width=.49\textwidth]{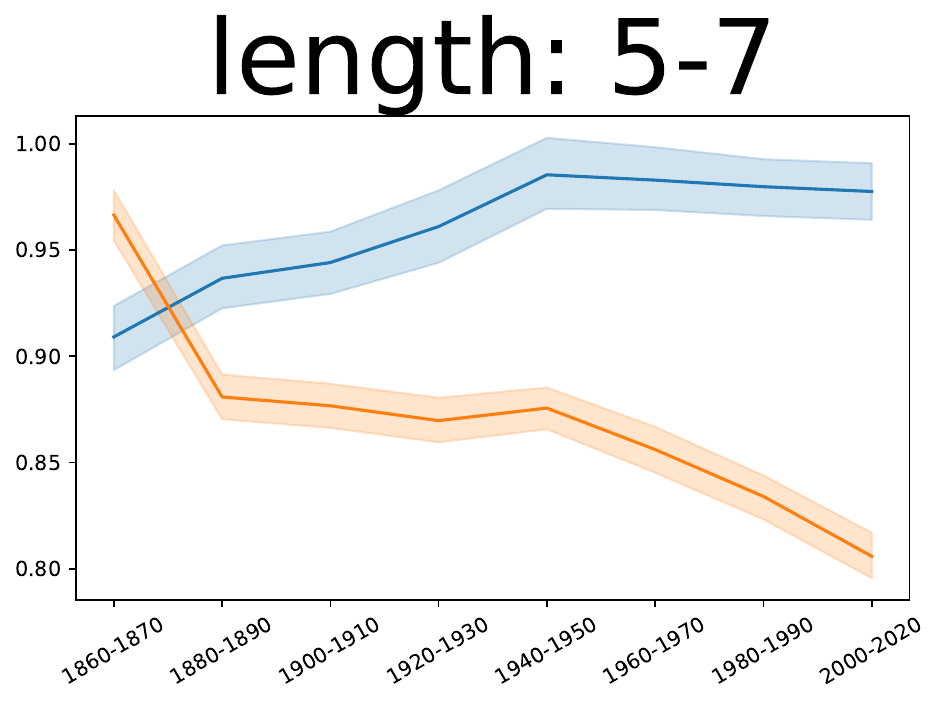}
\includegraphics[width=.49\textwidth]{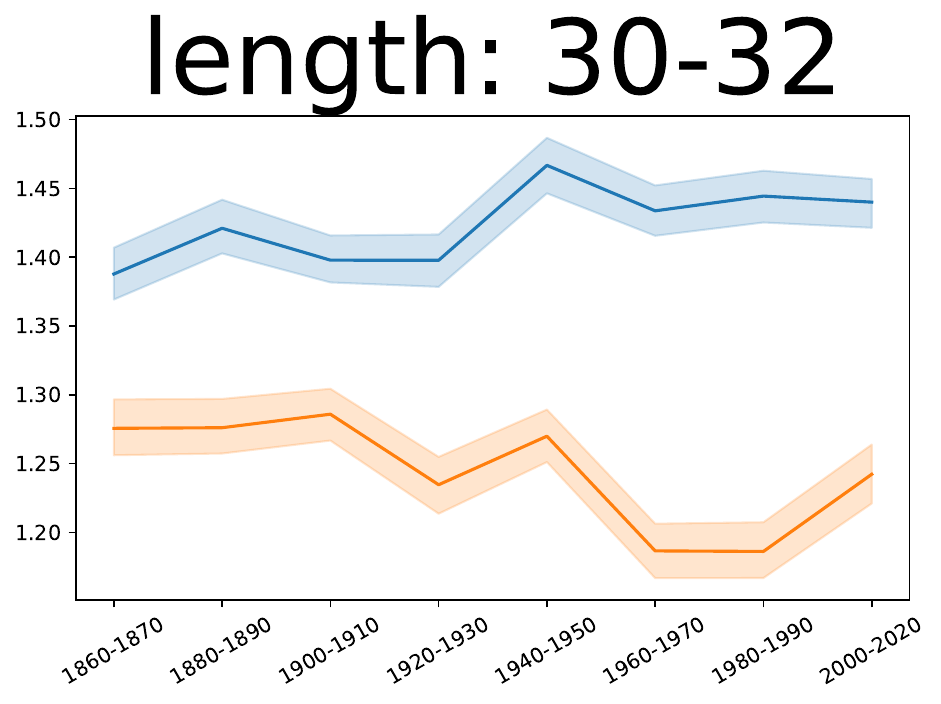}
    \caption{\protect\ndd{}}\label{fig:diff_ndd}
    \end{subfigure}
    \hspace{.05cm}
    \begin{subfigure}[t]{.2\linewidth}
    \centering
\includegraphics[width=\textwidth]{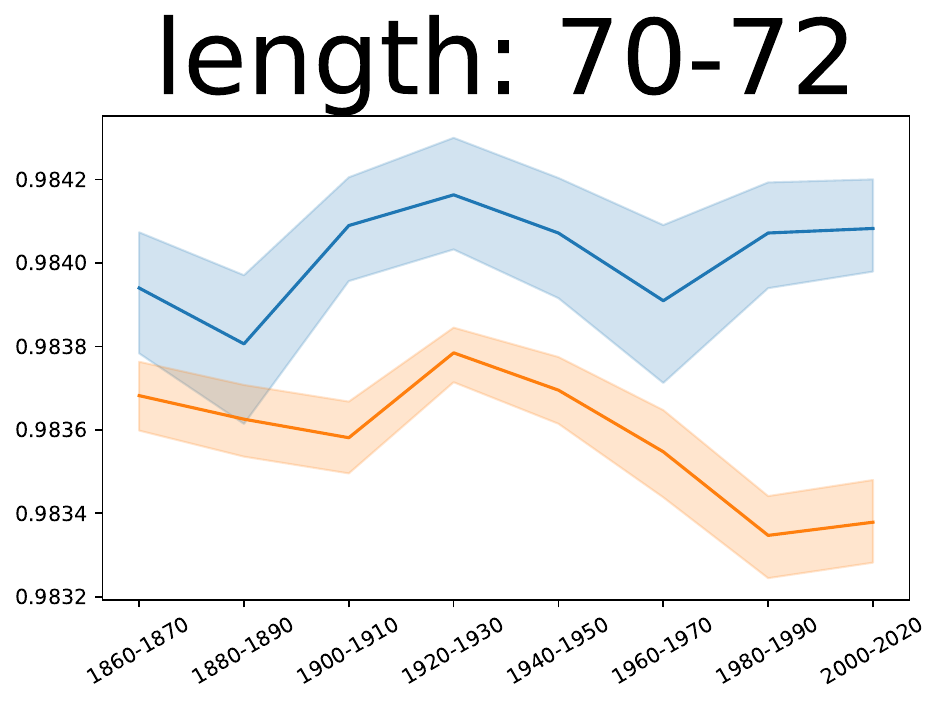}
    \caption{\protect\degmean{}}\label{fig:diff_degmean}
    \end{subfigure}
    \hspace{.02cm}
    \begin{subfigure}[t]{.2\linewidth}
    \centering
\includegraphics[width=\textwidth]{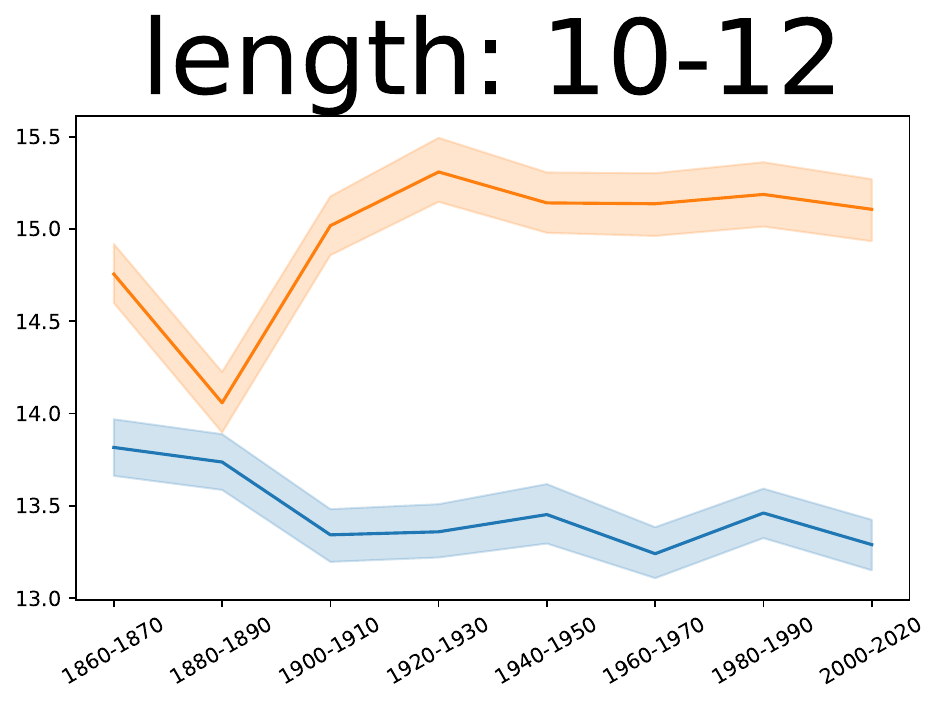}
\caption{\protect\lpath{}}\label{fig:diff_lpath}
    \end{subfigure}
    \begin{subfigure}[t]{.16\linewidth}
        \includegraphics[width=\textwidth]{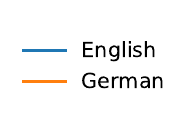}
    \end{subfigure}
    \caption{Metrics showing different diachronic trends based on MK between English (\textcolor{blue}{blue}) and German (\textcolor{orange}{orange}), averaged per decade group and over the 5 parsers. We show 95\% confidence interval\se{s}.}\label{fig:diff}
\end{figure}
We visualize the 4 distinct diachronic trends of metrics between English and German in Figure \ref{fig:diff}. It shows the metrics for English (blue) and German (orange) over time, averaged over the 5 parsers per decade group.
For \textbf{\ndd{}} (Figure \ref{fig:diff_ndd}), the trend goes up for English but down for German across both sentence lengths. Figure \ref{fig:diff_degmean} shows the trends of \textbf{\degmean{}} for sentences of length 70-72. Overall, while \degmean{} increases for English over time, it decreases for German, suggesting that, on average, words govern more dependents over time for long sentences in English but fewer in German. 
\textbf{\lpath{}} of sentences having 10-12 words generally increases for German  over time but decreases for English, as shown in Figure \ref{fig:diff_lpath}, indicating a tendency of the sum dependency distance from the root to a leaf word becoming larger in German but smaller in English for sentences of that length. 

\se{An interesting observation when closely examining the four different trend patterns in Figure \ref{fig:diff} is that for two (or even three) out of four cases, subtrends look quite similar across English and German. Witness \degmean{} as a point in case: there is a sequence of four subtrends: downward, upward, downward, upward in both English and German, where the German pattern seems to be lagged compared to the English one (i.e., German seems to mimic the English pattern). This is a limitation of our overall trend analysis which takes a global pattern into account, disregarding local subpatterns (cf.~Simpson's paradox \citep{simpson1951interpretation}).}

\begin{figure}[!t]
    \centering
    \begin{subfigure}[t]{0.495\linewidth}
    \centering
    \includegraphics[width=.325\textwidth]{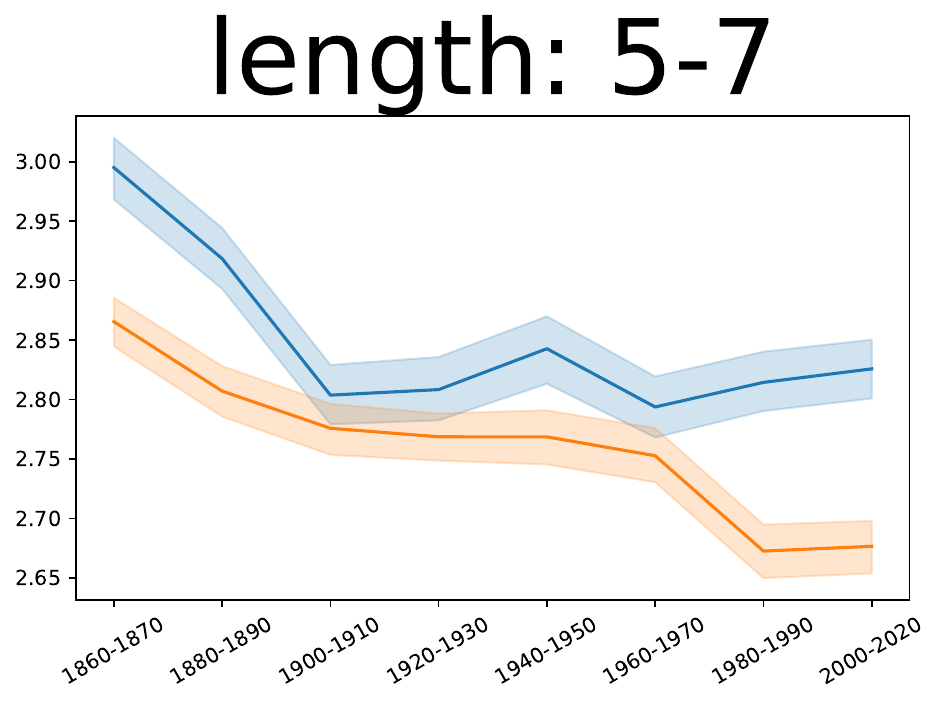}
\includegraphics[width=.325\textwidth]{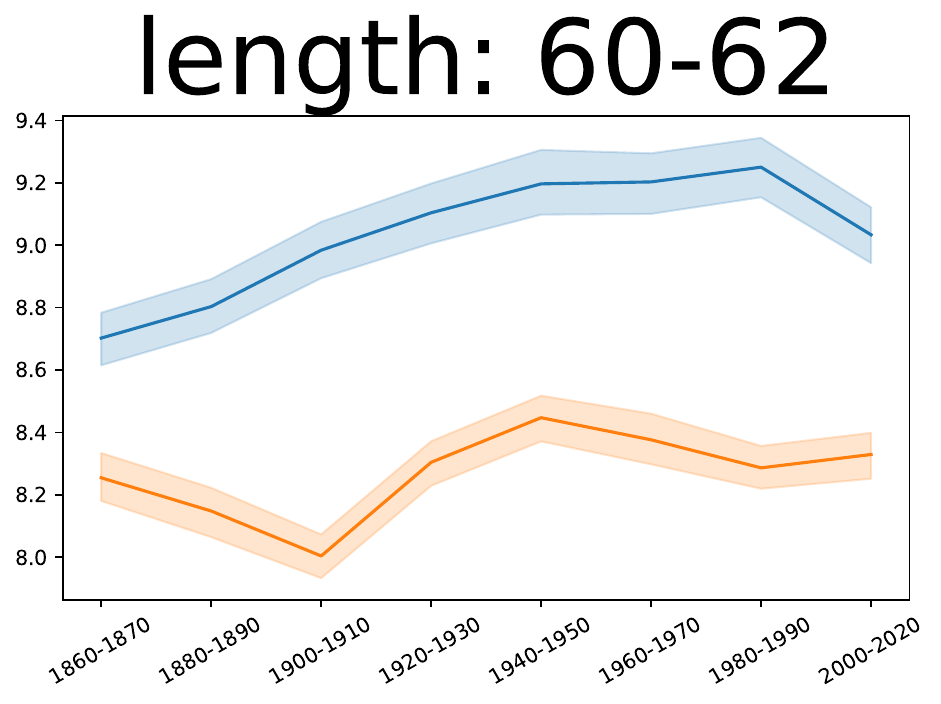}
\includegraphics[width=.325\textwidth]{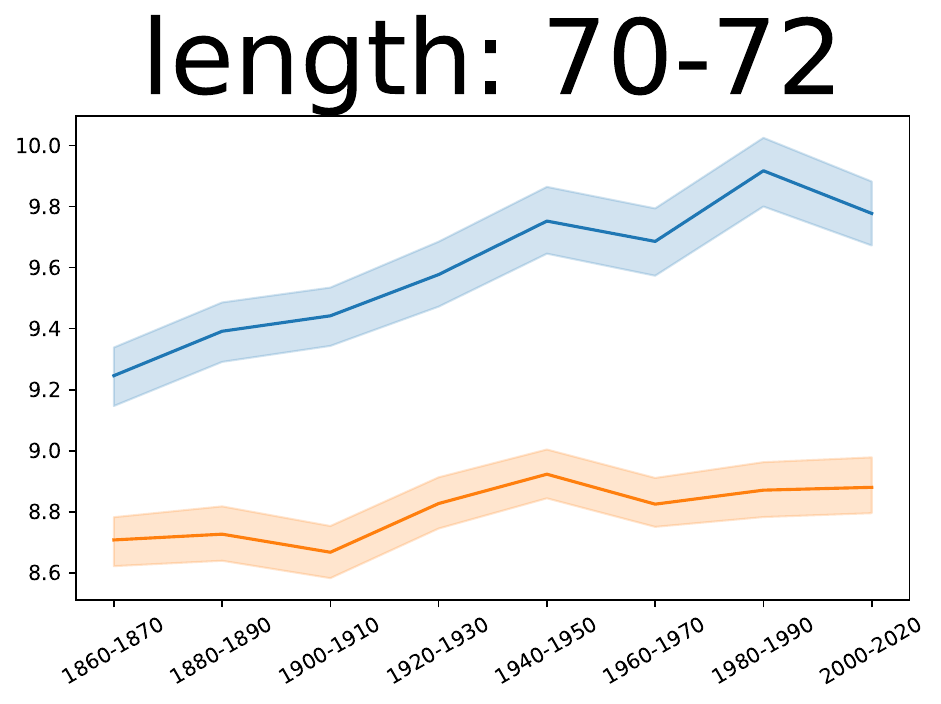}
\caption{\protect\height{}}\label{fig:same_height}
    \end{subfigure}
    \begin{subfigure}[t]{0.495\linewidth}
    \centering
\includegraphics[width=.325\textwidth]
{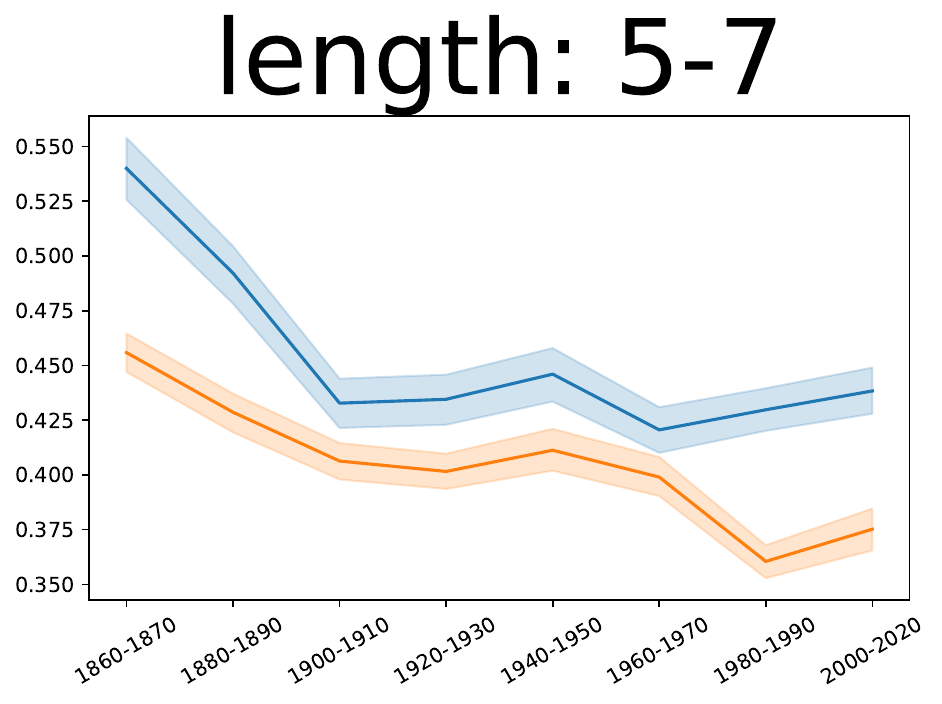}
\includegraphics[width=.325\textwidth]{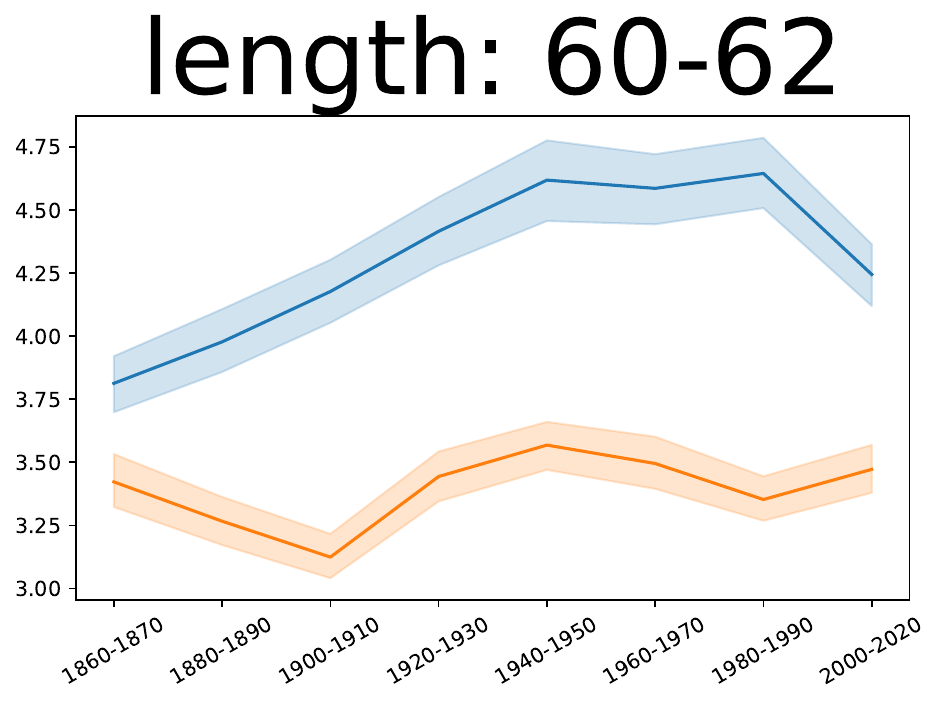}
\includegraphics[width=.325\textwidth]{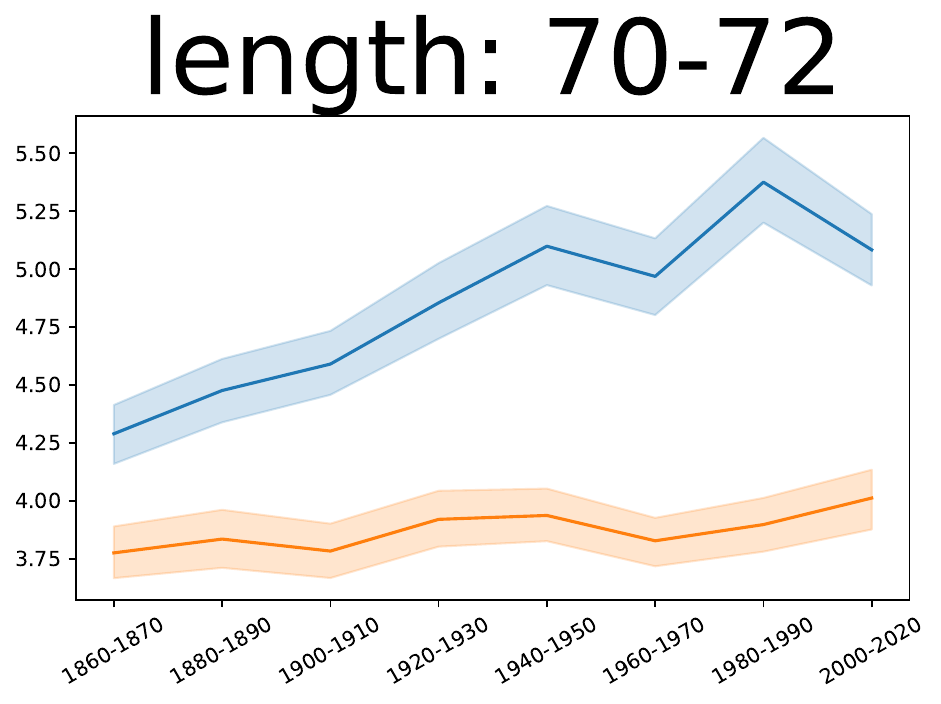}
\caption{\protect\depvar{}}\label{fig:same_depvar}
    \end{subfigure}
\begin{subfigure}[t]{\linewidth}
\centering
\includegraphics[width=.16\textwidth]{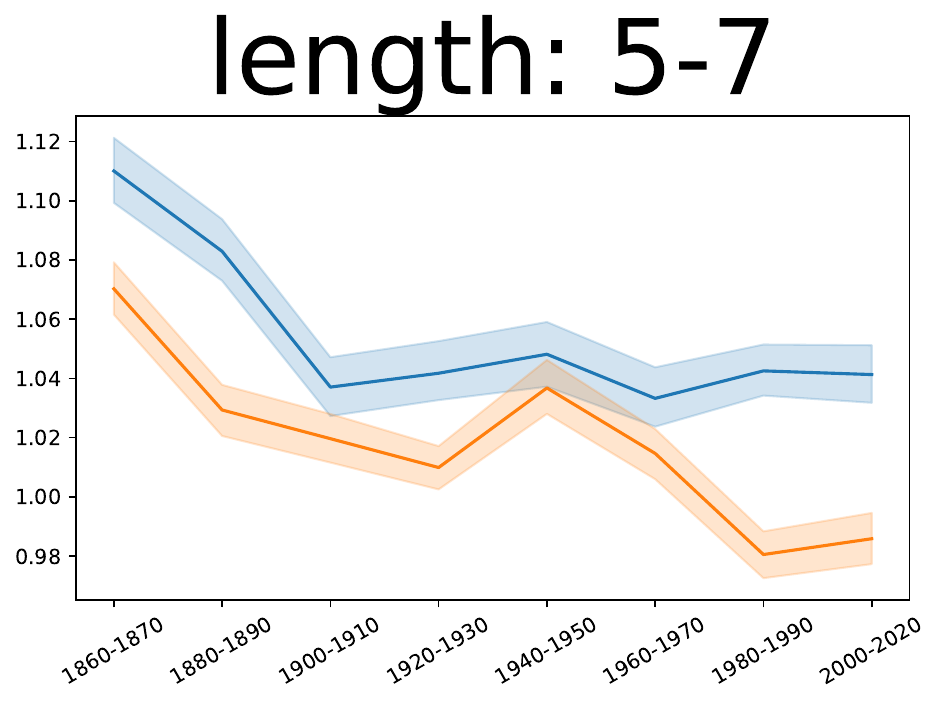}
\includegraphics[width=.16\textwidth]{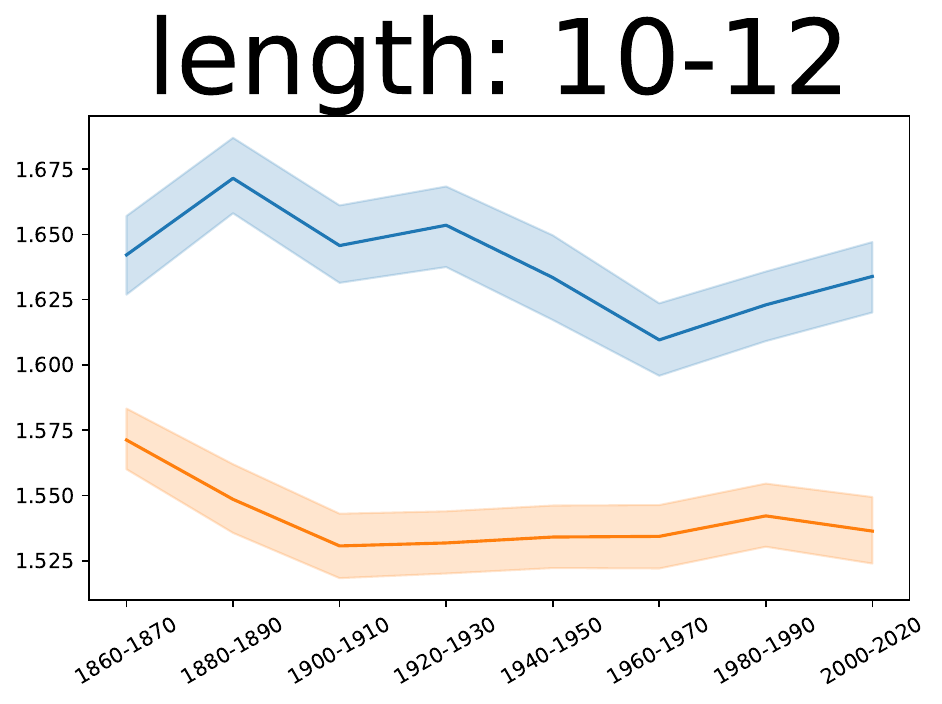}
\includegraphics[width=.16\textwidth]{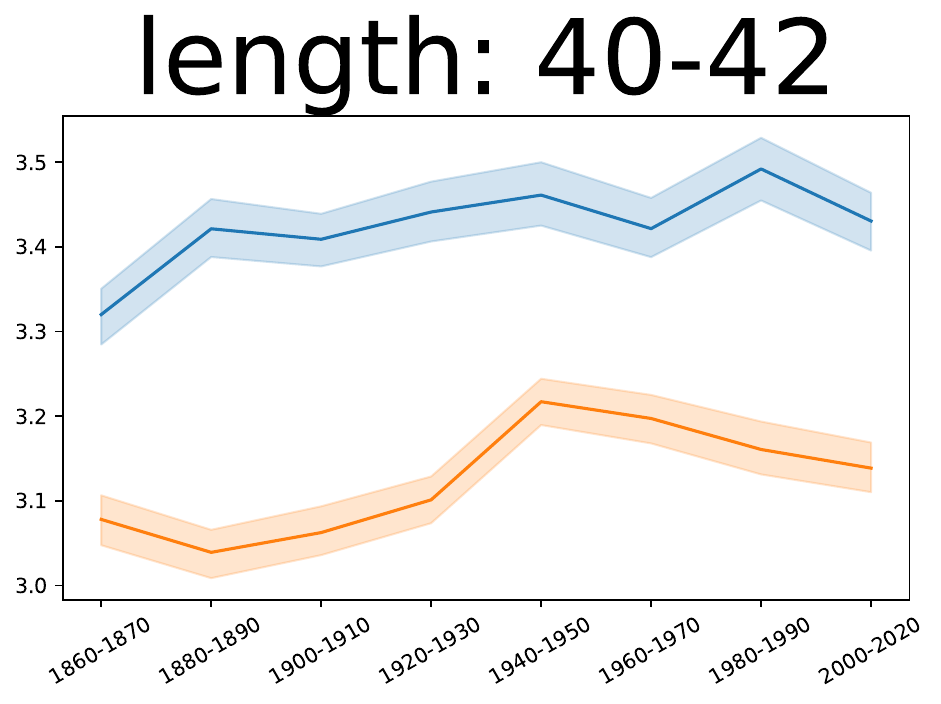}
\includegraphics[width=.16\textwidth]{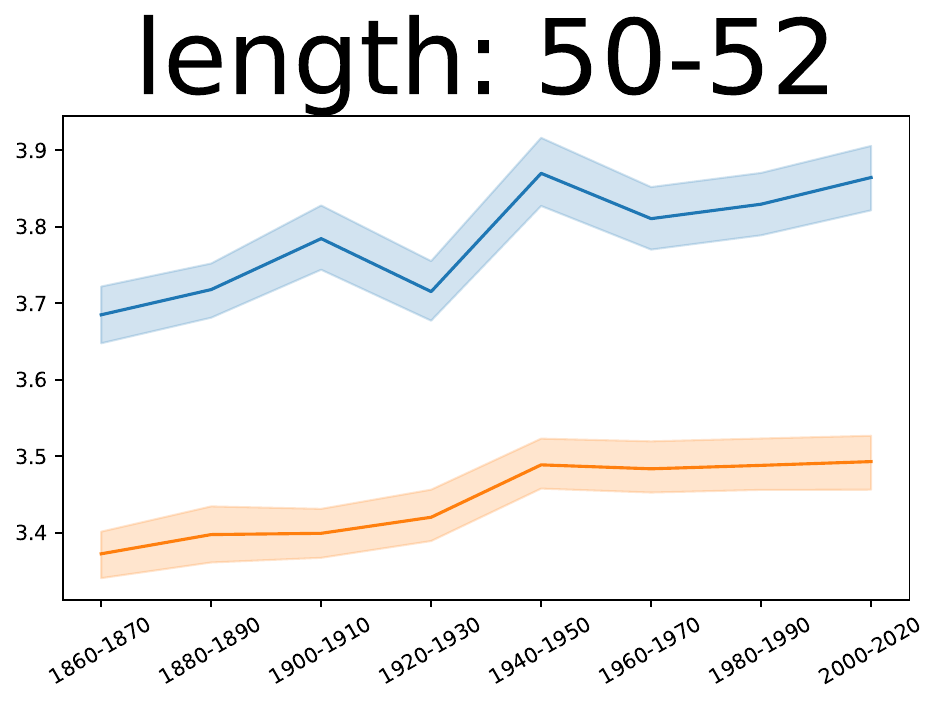}
\includegraphics[width=.16\textwidth]{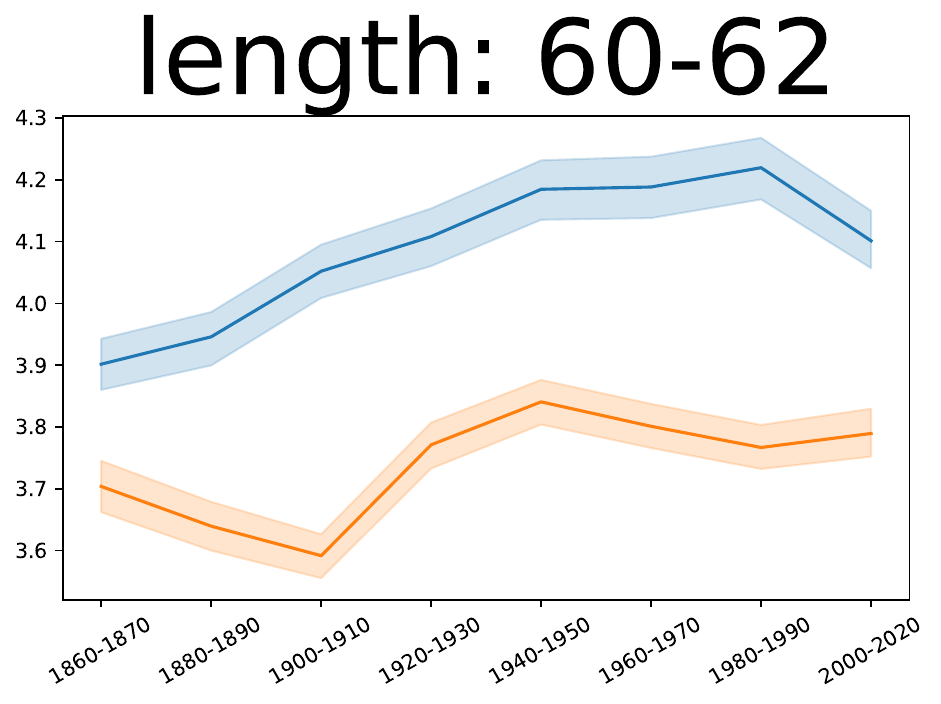}
\includegraphics[width=.16\textwidth]{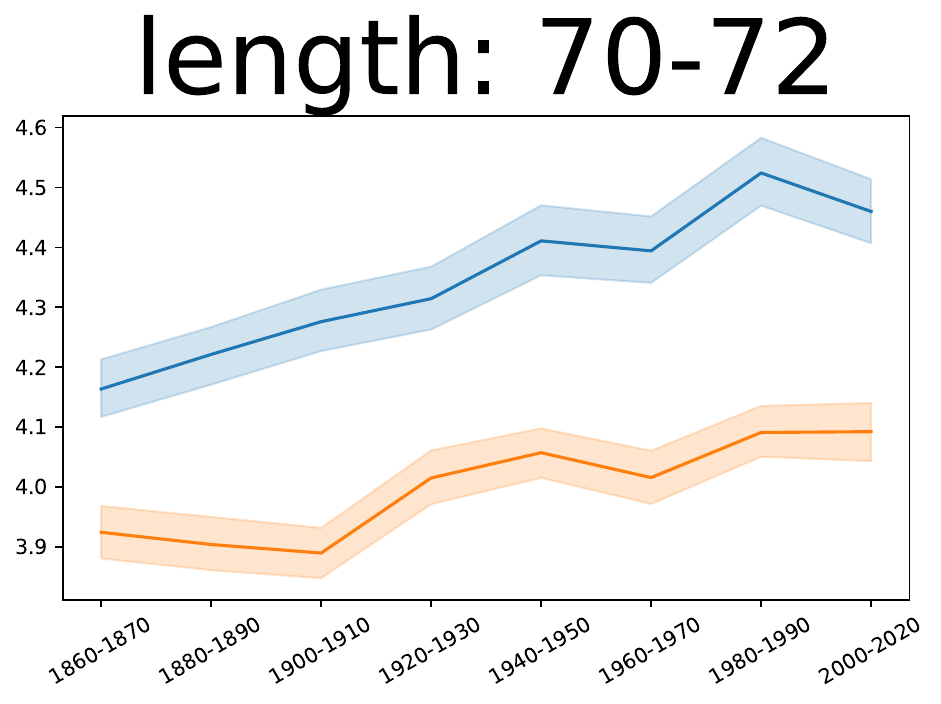}    \caption{\protect\depmean{}}\label{fig:same_depmean}
    \end{subfigure}
\begin{subfigure}[t]{\linewidth}
\centering
\includegraphics[width=.16\textwidth]{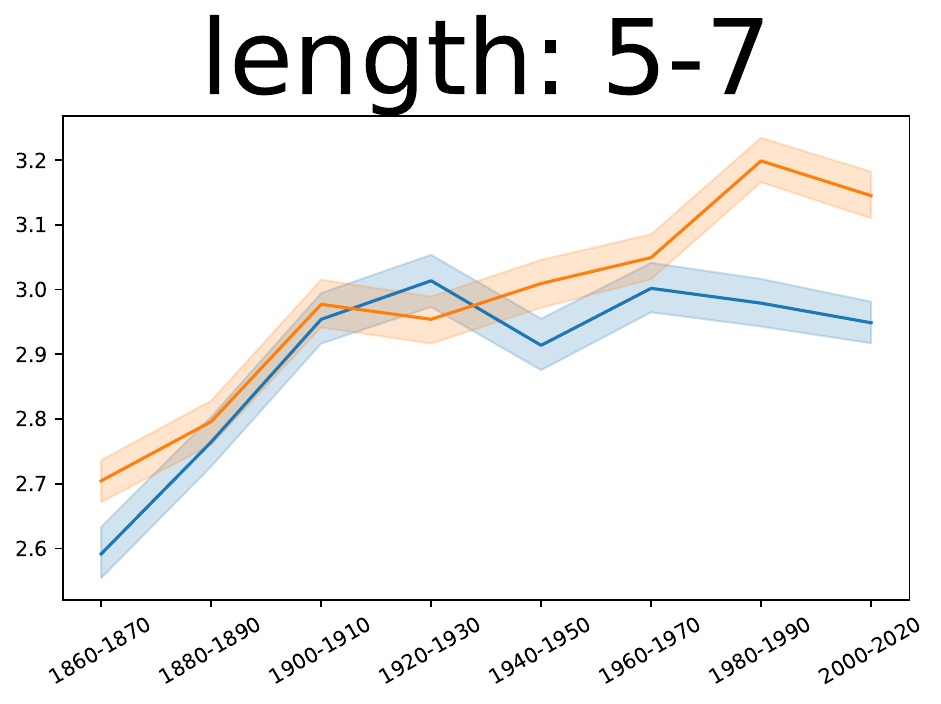}
\includegraphics[width=.16\textwidth]{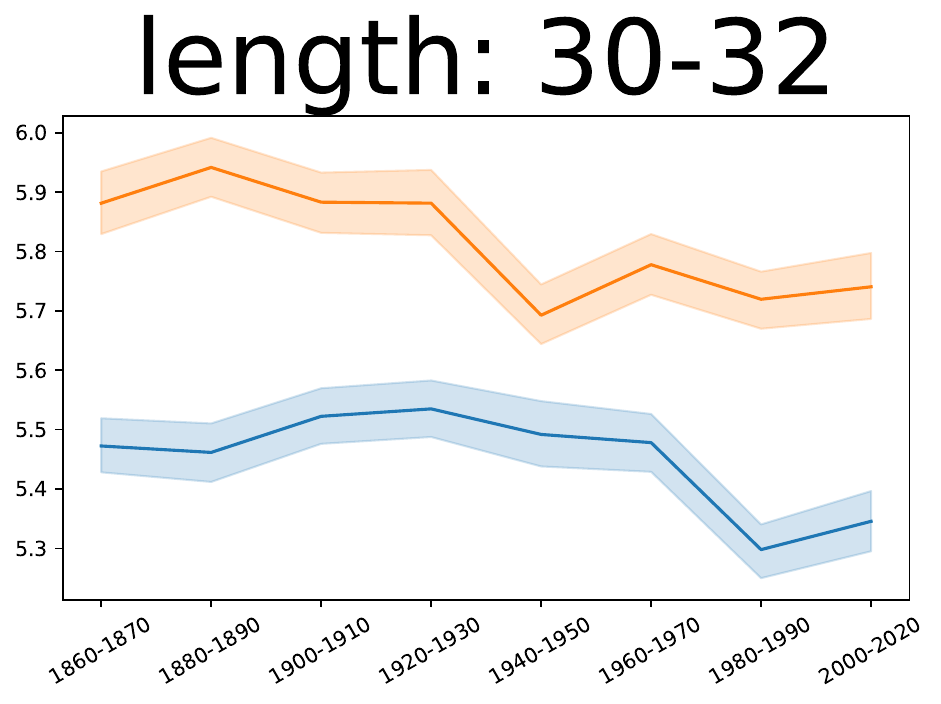}
\includegraphics[width=.16\textwidth]{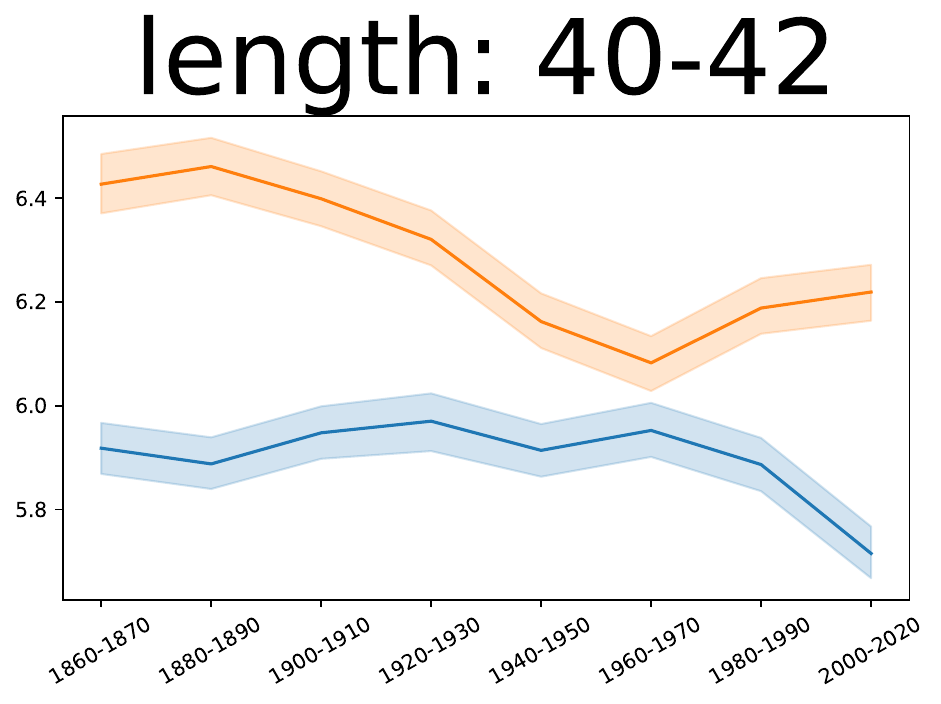}
\includegraphics[width=.16\textwidth]{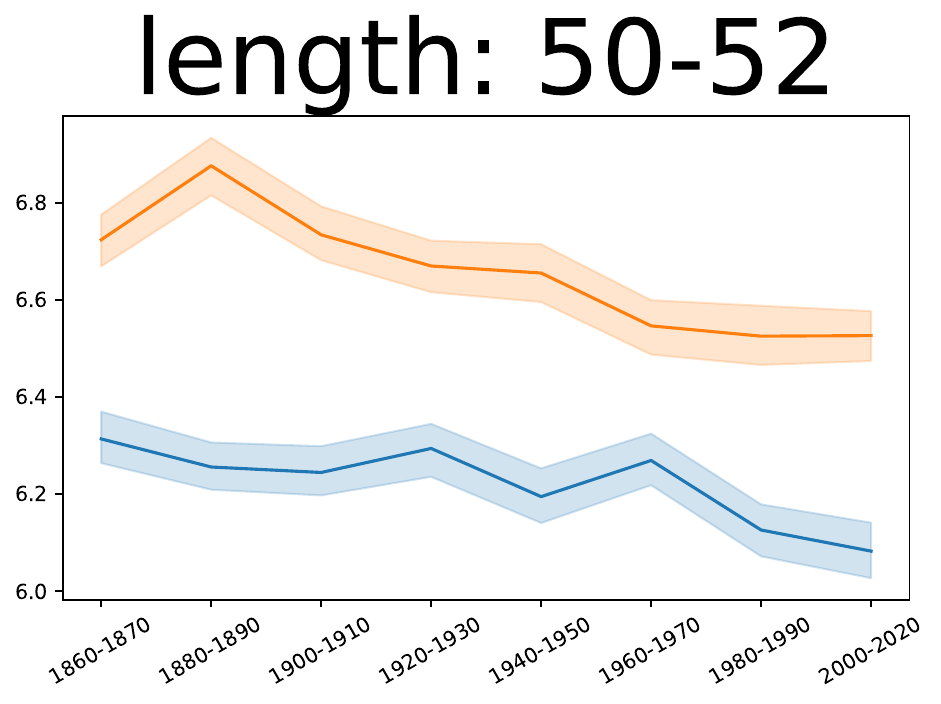}
\includegraphics[width=.16\textwidth]{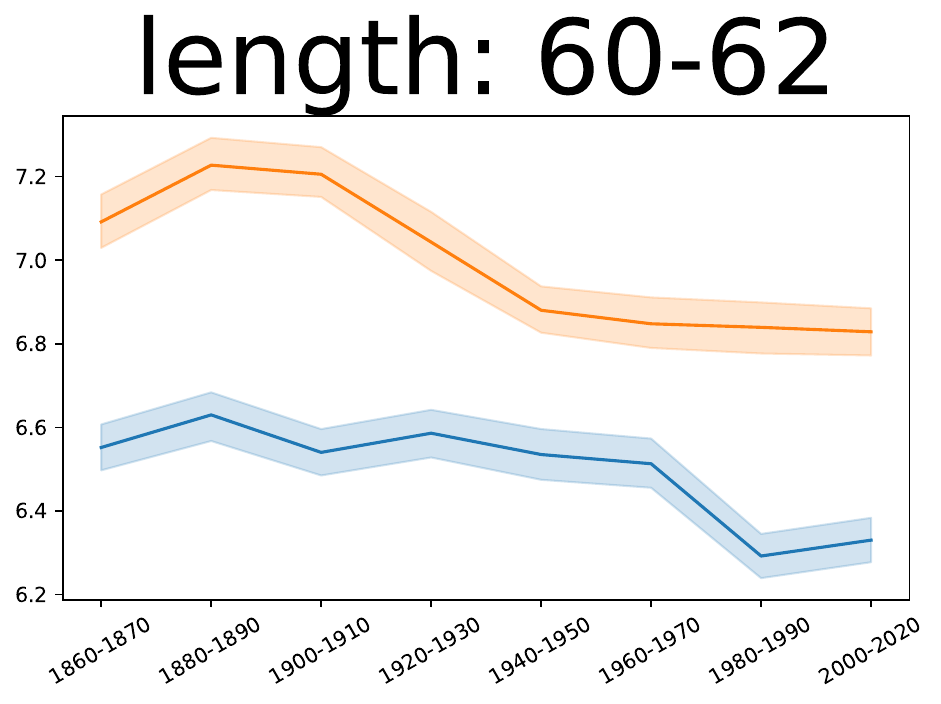}
\includegraphics[width=.16\textwidth]{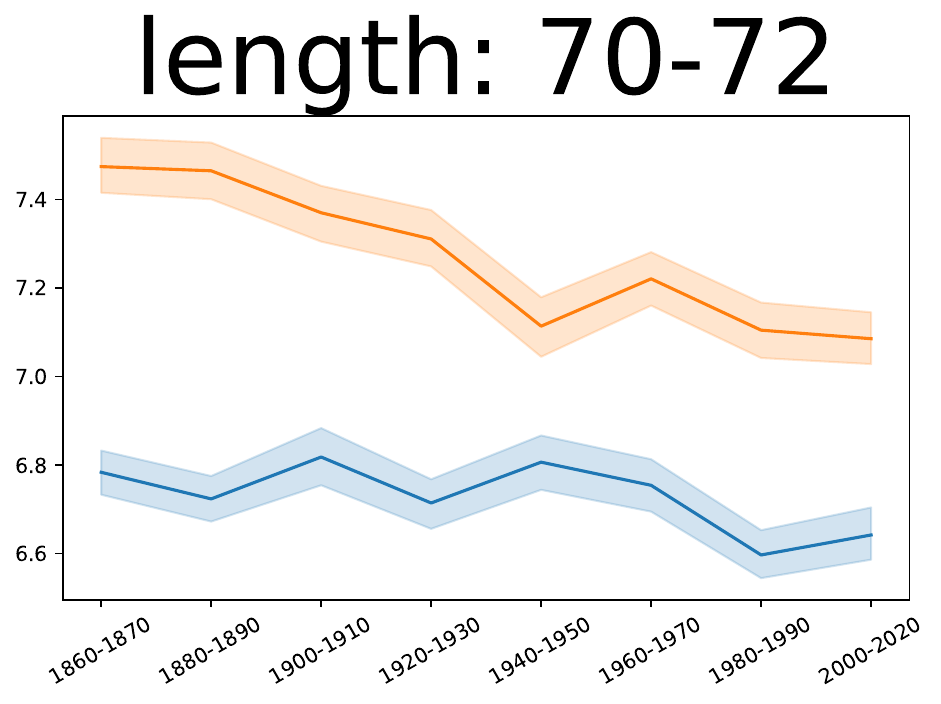}
\caption{\protect\degree{}}\label{fig:same_degree}
    \end{subfigure}
\begin{subfigure}[t]{.835\linewidth}
\centering
\includegraphics[width=.192\textwidth]{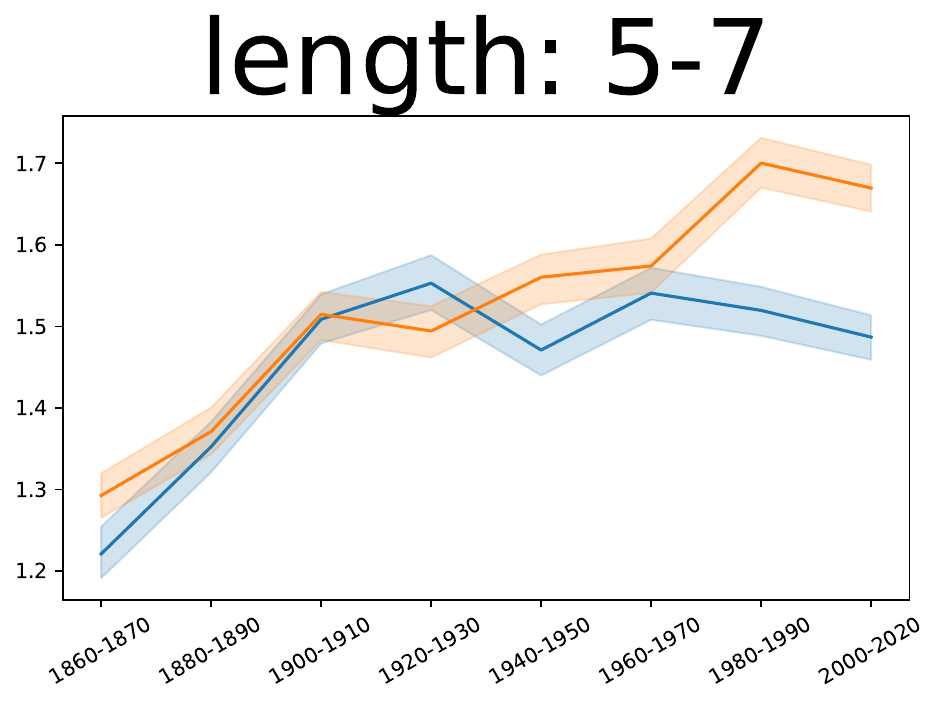}
\includegraphics[width=.192\textwidth]{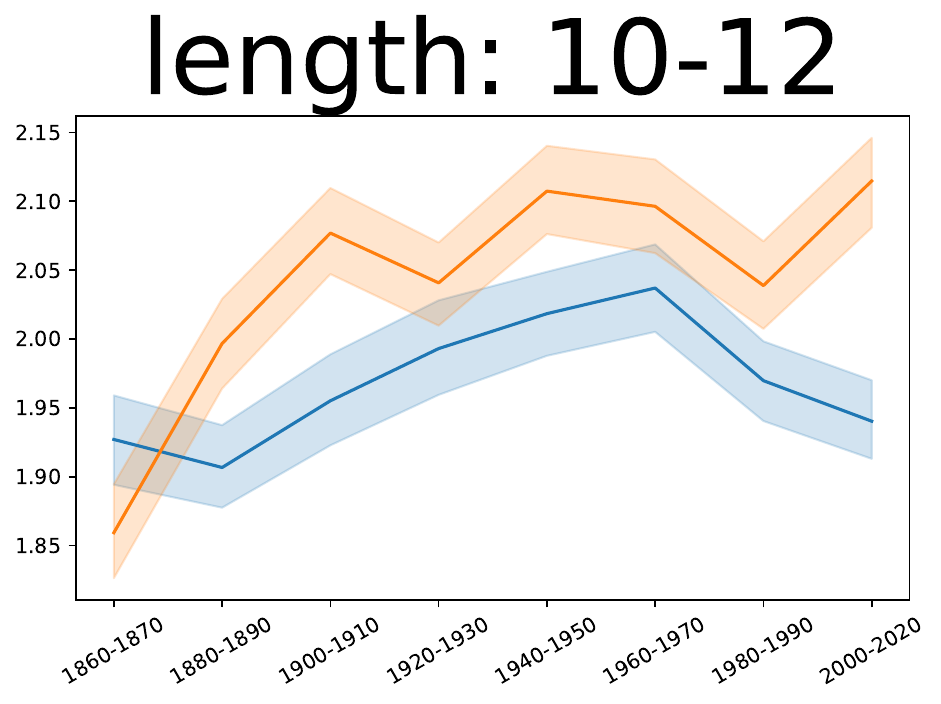}
\includegraphics[width=.192\textwidth]{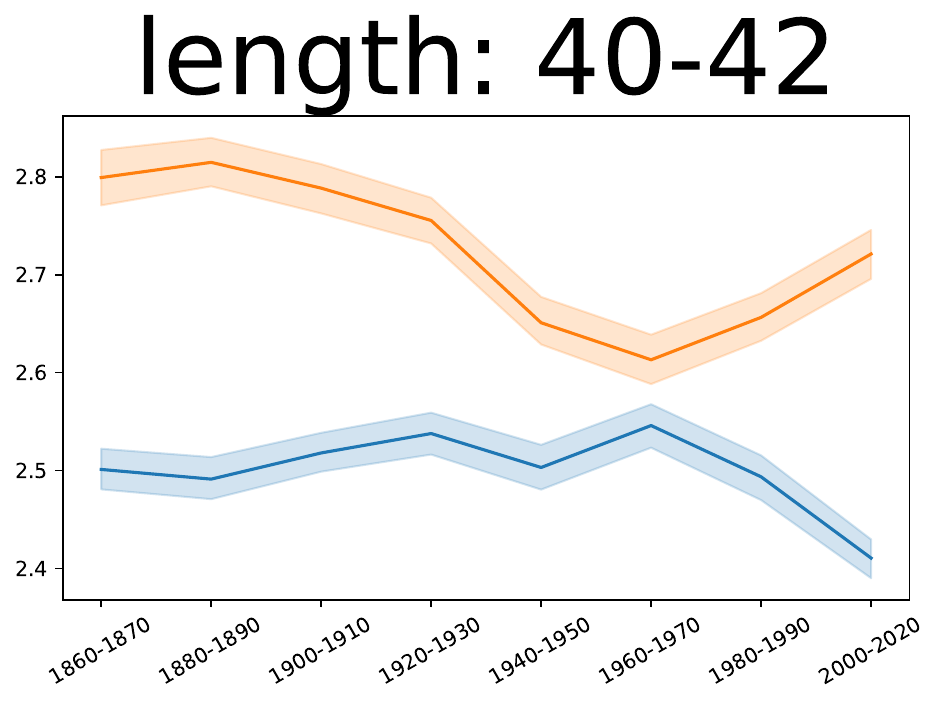}
\includegraphics[width=.192\textwidth]{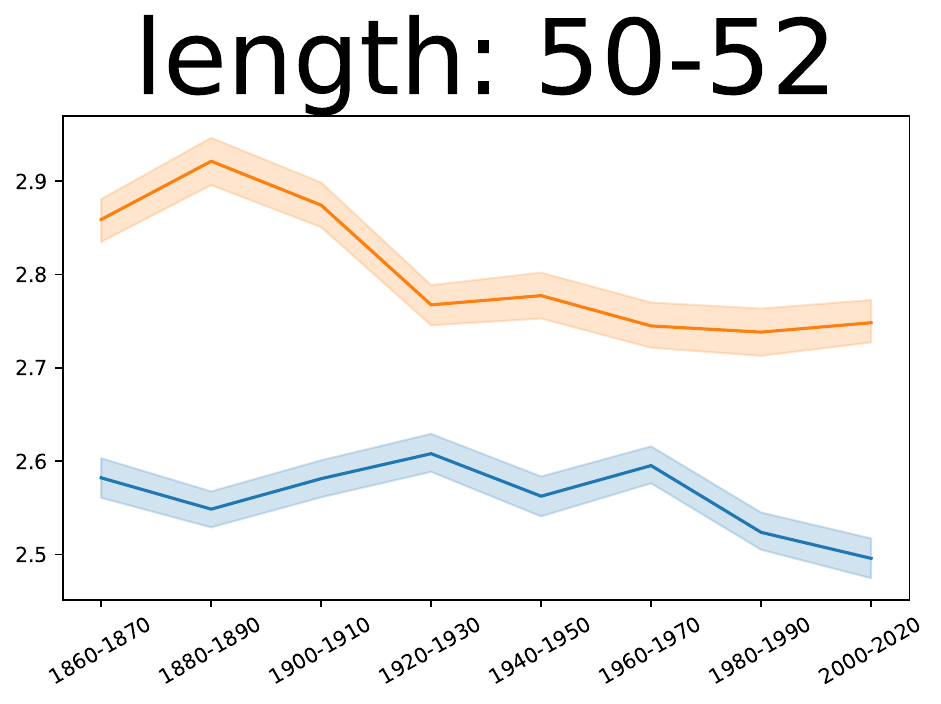}
\includegraphics[width=.192\textwidth]{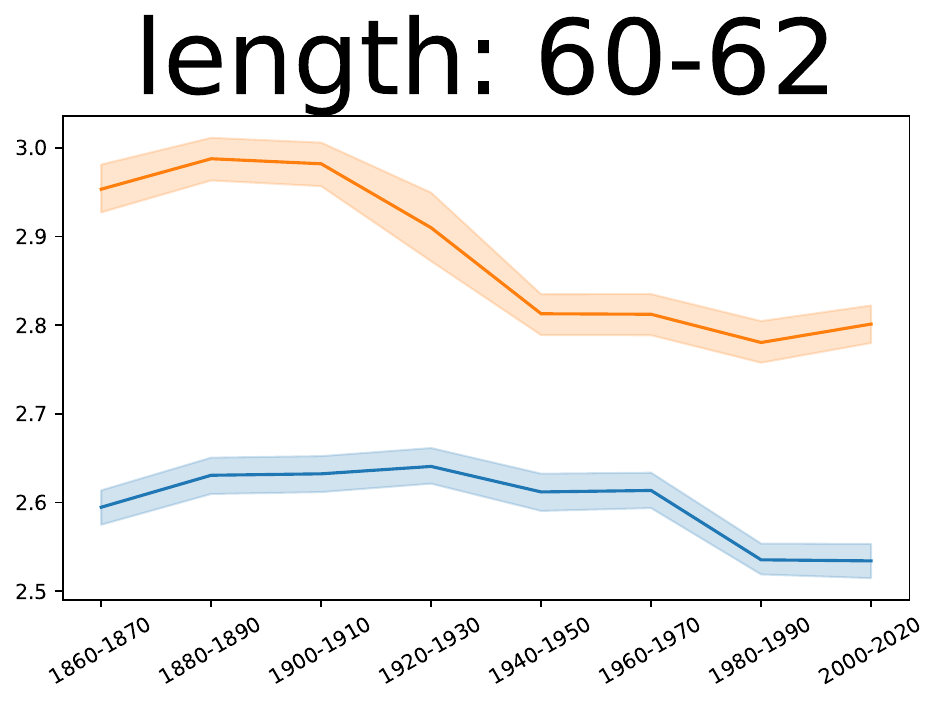}
\caption{\protect\degvar{}}\label{fig:same_degvar}
    \end{subfigure}
    
 \begin{subfigure}[t]{.16\linewidth}
\centering
\includegraphics[width=\textwidth]{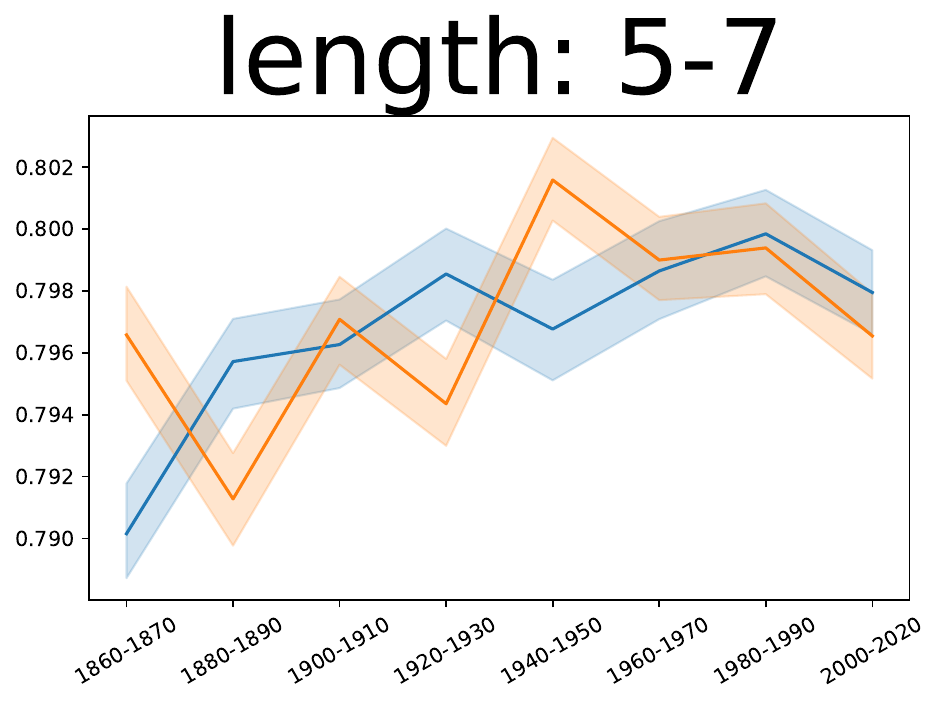}
\caption{\protect\degmean{}}\label{fig:same_degmean}
\end{subfigure}
\begin{subfigure}[t]{.5\linewidth}
\centering
\includegraphics[width=.325\textwidth]{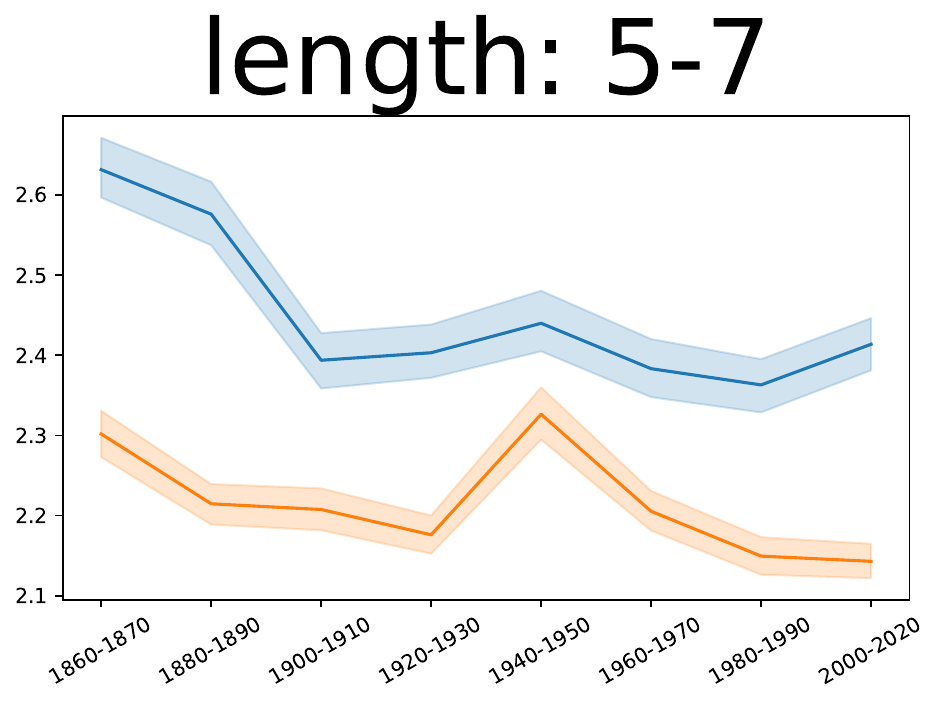}
\includegraphics[width=.325\textwidth]{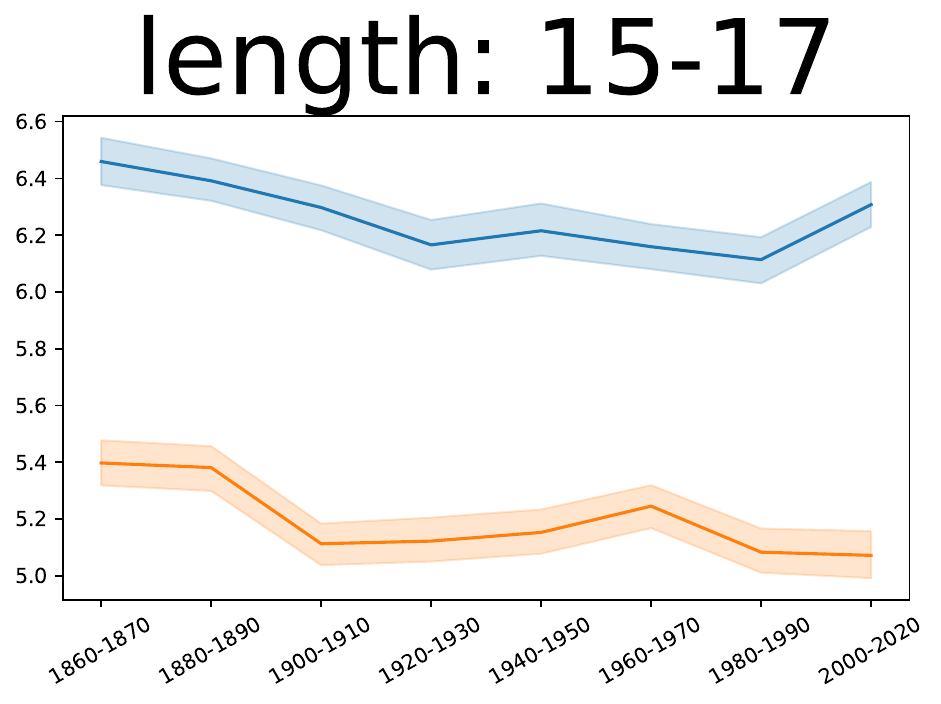}
\includegraphics[width=.325\textwidth]{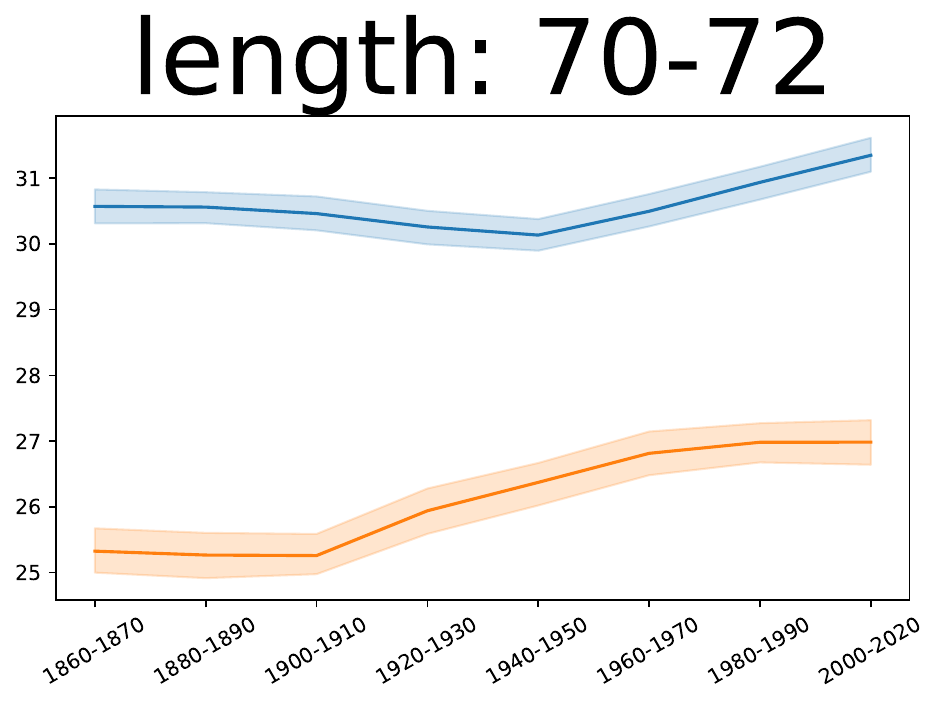}
\caption{\protect\headdis{}}\label{fig:same_headdis}
    \end{subfigure}
    \begin{subfigure}[t]{.16\linewidth}
\centering
\includegraphics[width=\textwidth]{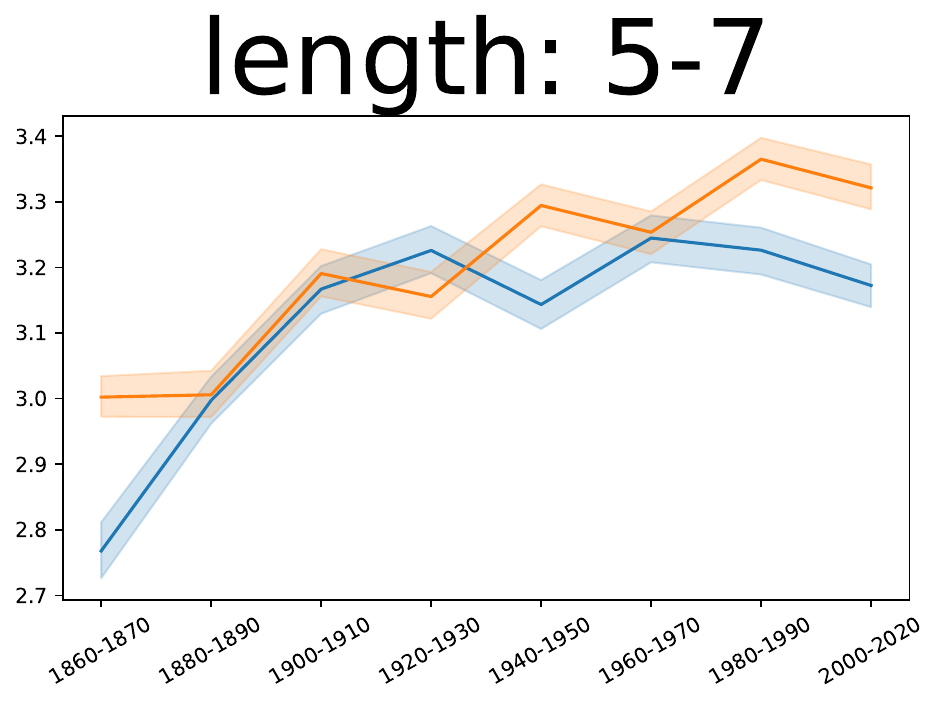}
\caption{\protect\leaves{}}\label{fig:same_leaves}
    \end{subfigure}
\begin{subfigure}[t]{.8\linewidth}
\centering
\includegraphics[width=.2\textwidth]{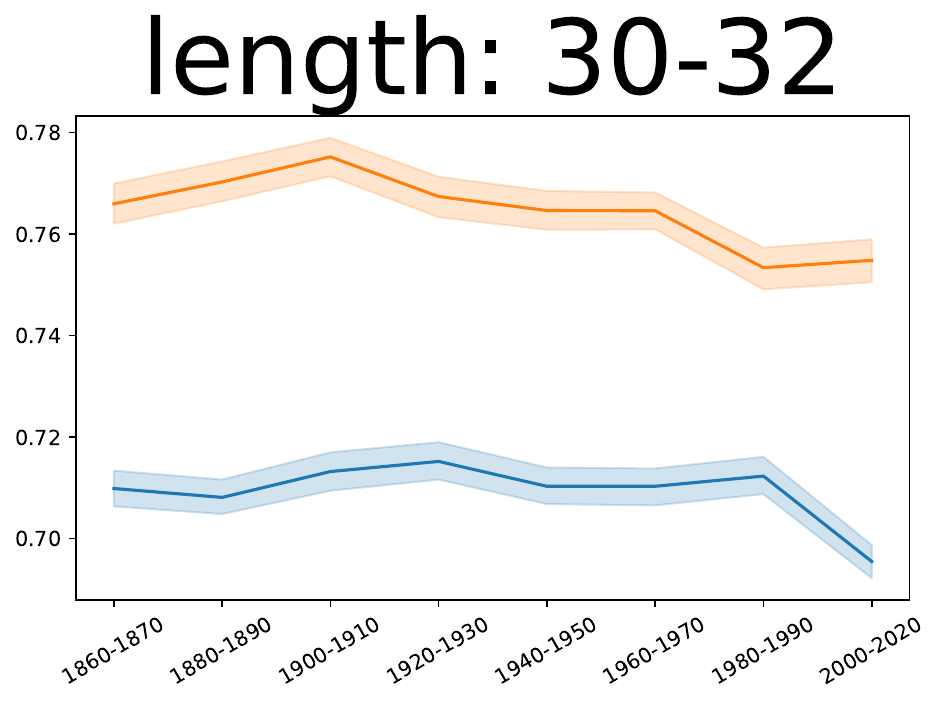}
\includegraphics[width=.2\textwidth]{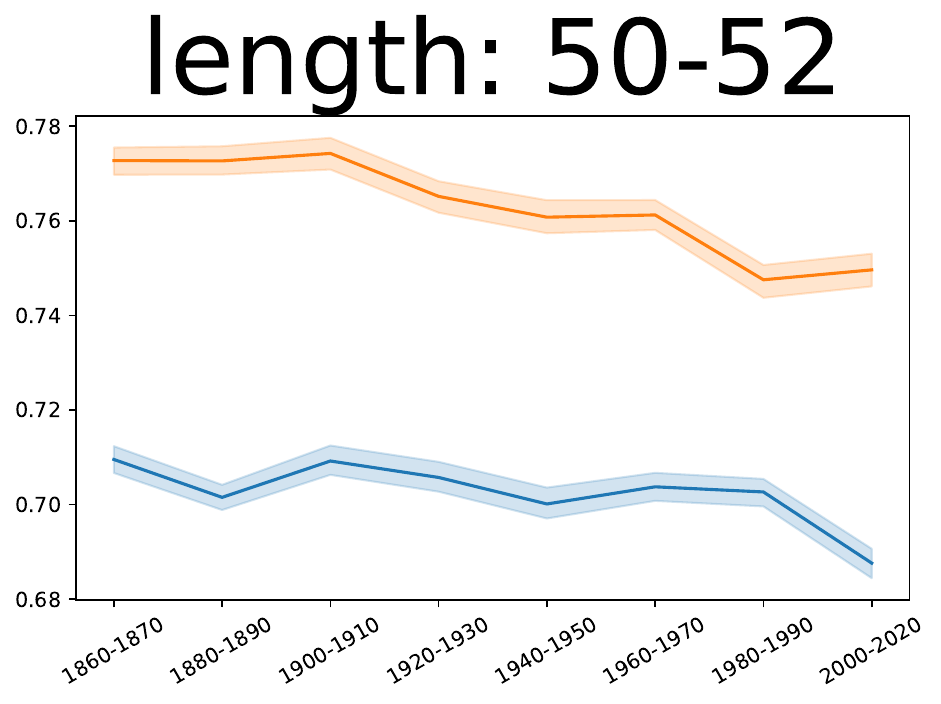}
\includegraphics[width=.2\textwidth]{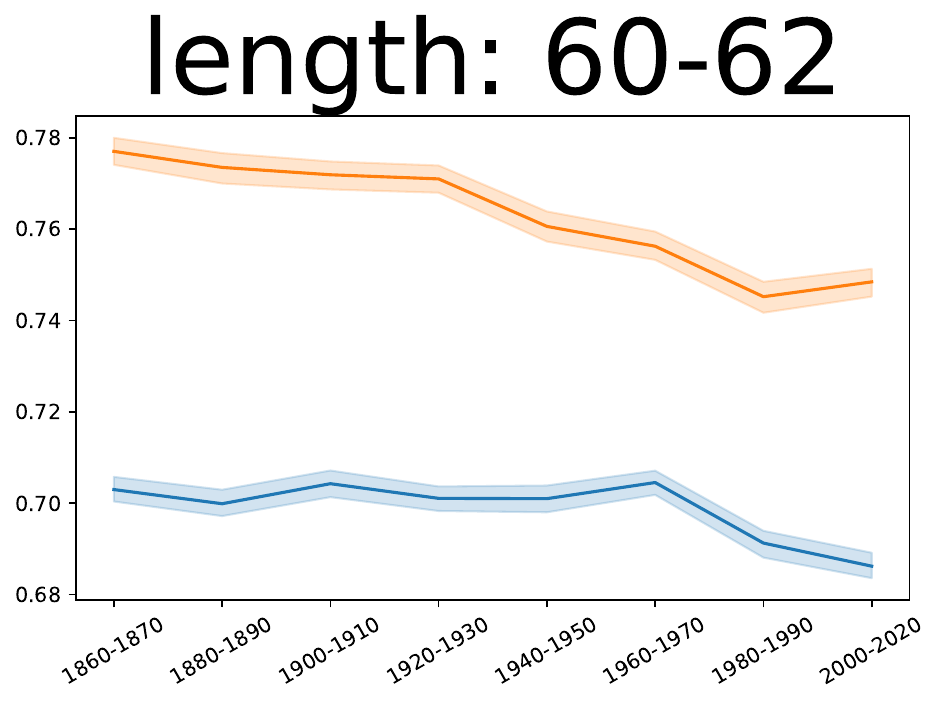}
\includegraphics[width=.2\textwidth]{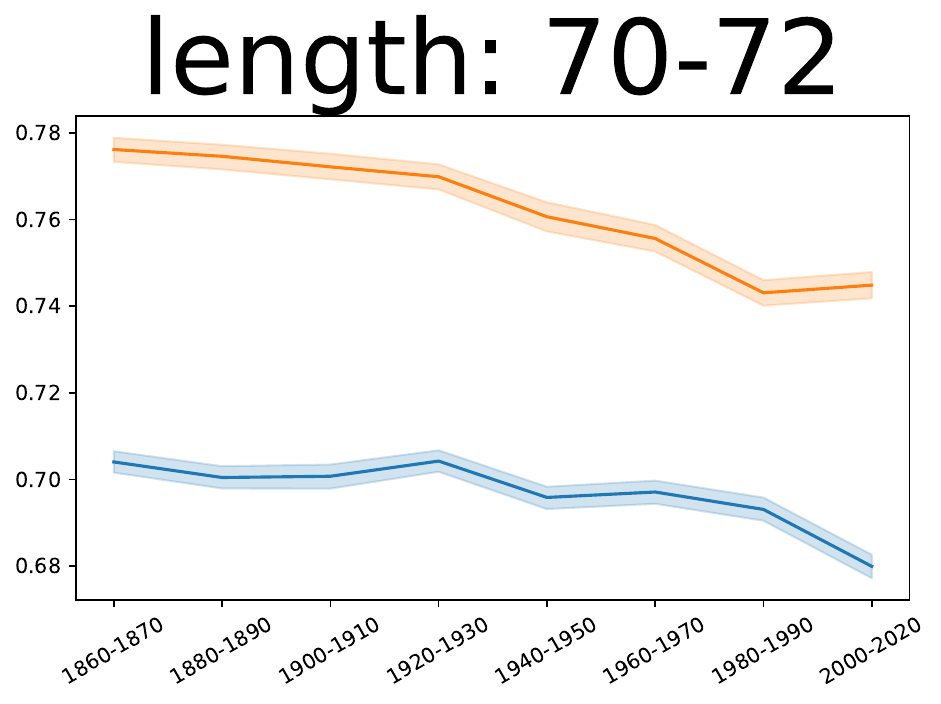}
\caption{\protect\headratio{}}\label{fig:same_headratio}
    \end{subfigure}
   \begin{subfigure}[t]{.16\linewidth}
        \includegraphics[width=\textwidth]{figs/same_period/lineplots/legend.pdf}
    \end{subfigure}
    \caption{Metrics showing the same diachronic trends based on MK between English (\textcolor{blue}{blue}) and German (\textcolor{orange}{orange}), averaged per decade group and over the 5 parsers. We show 95\% confidence intervals.}\label{fig:same}
\end{figure}

\subsection{What are the similarities in syntactic changes between English and German?}\label{sec:analysis_similarity}


Based on the MK trend test results, we identify 32 trends for 9 metrics that demonstrate simultaneous increase or decrease in both English and German. We show the average of those metrics per decade group over the 5 parsers in Figure \ref{fig:same}; blue lines represent the trends for English and orange lines for German. 

\begin{figure}[!ht]
    \centering
    \begin{subfigure}[t]{.25\linewidth}
    \centering
        \includegraphics[width=\textwidth]{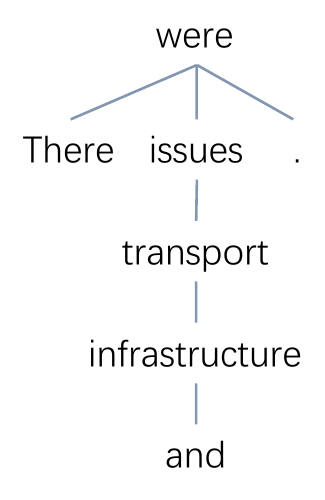}
        \caption{Taller and narrower: \protect\height=5, \protect\degree{}=2.}
    \end{subfigure}\hspace{1cm}
    \begin{subfigure}[t]{.5\linewidth}
    \centering
        \includegraphics[width=\textwidth]{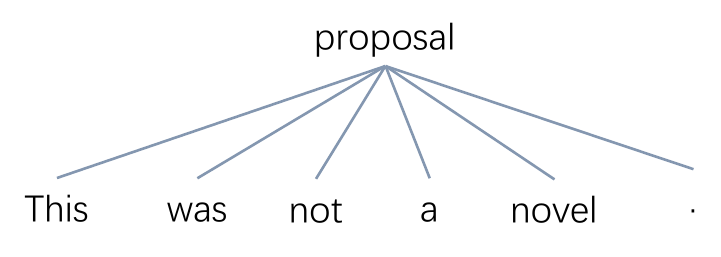}
        \caption{Shorter and wider: \protect\height=2, \protect\degree{}=5.}
    \end{subfigure}
    \caption{Dependency trees for sentences of length 7: 
    \ycf{taller and narrower vs.\ shorter and wider.}
    Predicted by Stanza. Both sentences taken from the Hansard corpus: left: `There were transport and infrastructure issues.'; right: `This was not a novel proposal.'}
    \label{fig:tree_example}
\end{figure}

For the metrics assessing dependency trees vertically, namely \textbf{\height{}} (Figure \ref{fig:same_height}), \textbf{\depvar{}} (Figure \ref{fig:same_depvar}) and \textbf{\depmean{}} (Figure \ref{fig:same_depmean}), the trends are similar: they decrease over time for short sentences having 5-7 and/or 10-12 words but increase for long sentences like those of length 60-62 and 70-72. In contrast, the degree-relevant metrics, which assess trees horizontally, increase for short sentences 
\se{but} 
decrease for long sentences: in Figure \ref{fig:same_degvar}, \ref{fig:same_degree} and \ref{fig:same_degmean}, we observe an increase in \textbf{\degvar{}}, \textbf{\degree{}}, and \textbf{\degmean{}} for sentences having 5-7 and/or 10-12 words, as well as a decrease in both \degree{} and \degvar{} for longer sentences of lengths from 40-42 to 60-62 words.
All of the above suggests that generally, for both English and German, the dependency trees of shorter sentences with $\leq$10-15 \se{words} become shorter and wider over time, while those of longer sentences having $\geq$30 words become taller and narrower over time. Short and wider tree structures have a higher chance that a word governs multiple dependents compared to the taller and narrower ones and vice versa. \ycj{An example for comparing between the shorter and wider vs.\ the taller and narrower dependency trees is given in Figure \ref{fig:tree_example}, where both sentences have 7 tokens.}

The trends of \textbf{\headdis{}} and \textbf{\headratio{}} are illustrated in Figure \ref{fig:same_headdis} and \ref{fig:same_headratio} respectively. A larger \headdis{} of a sentence indicates greater distance from the corresponding fully head-final sentence (where all dependency pairs of that sentence are head-final); thus, it suggests that the sentence is more head-initial. \headdis{} is expected to rise with declined \headratio{} to reflect language becoming more head-initial and vice versa.
We see that: (1) \headdis{} decreases for sentences with 5-7 and 15-17 words but increases for those having 70-72 words, implying that short sentences become more head-final, while long sentences become more head-initial; (2) \headratio{} decreases over time for sentences having $\geq$ 30 words, also suggesting a tendency for longer sentences being more head-initial over time.


Figure \ref{fig:same_leaves} displays the trends of \textbf{\leaves{}} for sentences having 5-7 words, where we see an increase over time for both languages. This suggests a tendency that more words never serve as a head and a head governs more dependents in short sentences, which is consistent with the changes in overall dependency tree structures, i.e., they become shorter and wider for short sentences.



\section{Discussion}

\begin{figure}[!ht]
    \centering
    \begin{subfigure}[t]{0.497\linewidth}
        \centering
        \includegraphics[width=\textwidth]{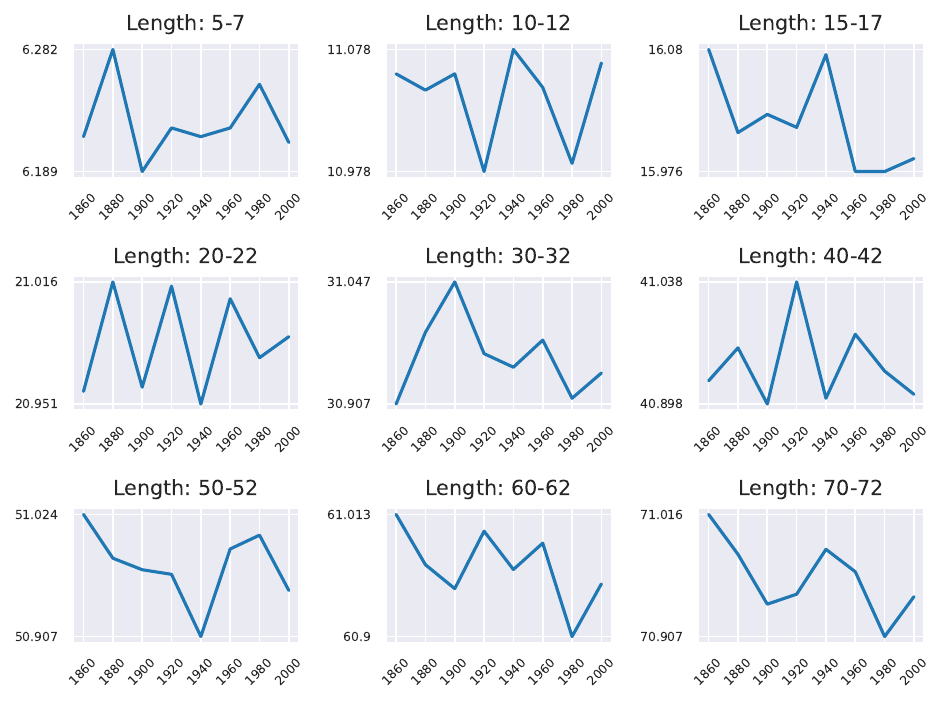}
    \caption{English}
    \label{fig:avg_len_en}
    \end{subfigure}
    \begin{subfigure}[t]{0.497\linewidth}
        \centering
        \includegraphics[width=\textwidth]{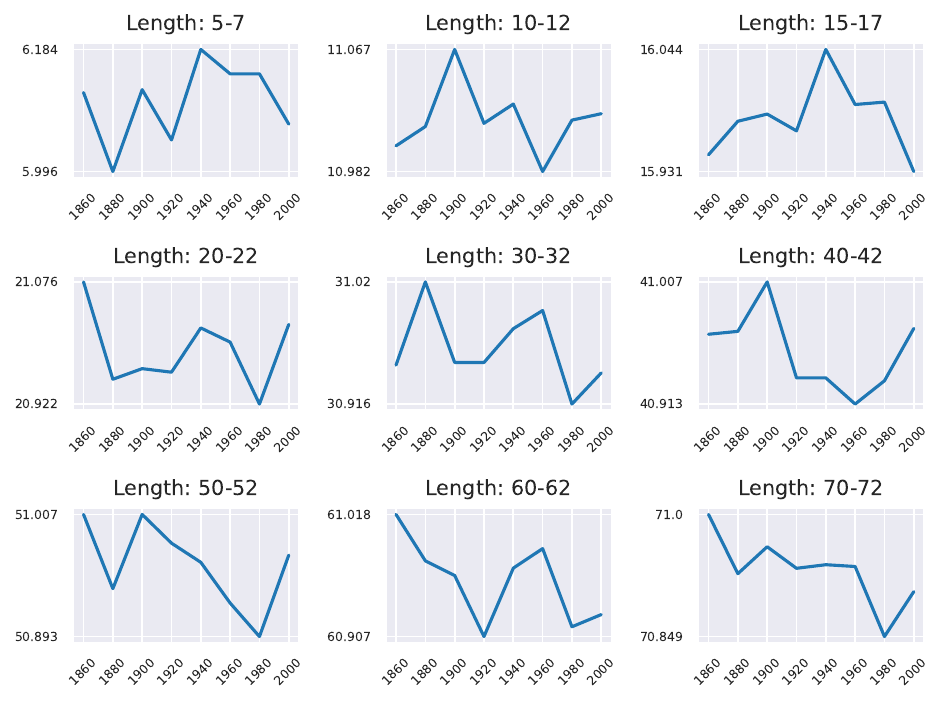}
    \caption{German}
    \label{fig:avg_len_de}
    \end{subfigure}
    \caption{Average sentence length in our observation data over time. We show the maximal and minimal average lengths on the y-axes.}
    \label{fig:avg_len}
\end{figure}

\subsubsection{Effect of sentence length}\label{sec:effect_len}
\ycj{
As mentioned in \S\ref{sec:datasets}, it is important to control the sentence length for diachronic 
\sen{syntactic}
observation\se{s} 
\se{as it may be a confounding variable}.\footnote{\sen{On the other hand, reducing sentence length over time may be one core aspect of syntactic simplification, e.g., splitting one long sentence into two shorter sentences.}} 
Thus, we compute the diachronic trend of the average sentence length per decade group for each length group in our data, visualized in Figure \ref{fig:avg_len}.
Even if the length difference within one group is up to 3 tokens in our data, we still often see a decrease in average sentence length over time, especially for the longer sentences with more than 50 tokens (last rows in Figure \ref{fig:avg_len}). However, the difference in the average sentence length within one length group is only up to $\sim$0.1 token, which we deem as an acceptable level. The reasons for this are the following: (1) Intuitively, such a small difference in sentence length is unlikely to influence 
\se{our} 
metrics substantially. For instance, it takes at least one additional token to increase the maximal possible \height{} and \degree{}.
(2) We 
observe trends of metrics that counter the development of sentence lengths, e.g., that longer sentences have a higher chance to produce taller dependency trees compared to shorter sentences; however, we see uptrends in \height{} in Figure \ref{fig:same_height} for 
sentences with 60-70 words, whose lengths are overall decreasing over time for both languages. 
This suggests that such a minor decrease in sentence length does not substantially impact the observations on syntactic changes. 
(3) The subtrends in metrics do not often follow the subtrends in sentence length. For instance, in 6 out of 8 metrics in Figure \ref{fig:same} for sentences of length 5-7 (\height{}, \depvar{}, \depmean{}, \degree{}, \degvar{}, \headdis{}), the corresponding trends behave consistently across both languages from the time window 1860-1870 to 1880-1890, even when German sentences become longer while English sentences become shorter 
during that period. 
}

\vspace{-.2cm}
\subsubsection{Scale of syntactic changes} \se{We note that 
\sen{several}
of our syntactic trends, even if they are statistically significant, are small in scale. For example, \degmean{} in Figure \ref{fig:diff_degmean} changes from slightly below 0.984 to slightly above 0.984 in English over the past 160 years (not all trends are on such low scales, however: \lpath{}  decreases from slightly below 14 to slightly above 13 in the same time period in English (Figure \ref{fig:diff_lpath}), for example). On the one hand, this makes sense as syntax is expected to change much slower, e.g., compared to lexical or semantic changes. On the other hand, this may also mean that some of our identified changes may hardly 
be noticeable (by linguist experts)  
in actual language or will more pronouncedly 
manifest themselves only in the future.} 
\vspace{-.2cm}
\subsubsection{About \protect\ddm{}}
\ycj{
Although some previous studies have shown that the \mdd{} is decreasing over time, 
in this work, we do not see direct evidence supporting this. In fact, \mdd{} for German is increasing over time across 7 out of 9 lengths in our experiment, while it is only decreasing for English sentences having 70-72 tokens (see Table \ref{tab:trend} in the appendix). Overall, our observations on \mdd{} are more in line with \citet{krielke2023optimizing} and \citet{zhu2022investigating}, who base observations on the sentences of identical lengths instead of grouping sentences by a range of lengths (e.g., \citet{lei2020dependency}). 
\se{The former}
mostly find no trend or uptrend in \mdd{} for `general' text domains (vs.\ scientific language) including news, magazines etc. 
Further, some syntactic structures have been argued to appear under the pressure of \ddm{}. E.g., 
\citet{liu2017dependency} demonstrate that dependency trees with more hierarchical depth and fewer parallel words are preferred since they are in a better position to produce shorter dependency distances; this is in accordance with our observations where the trees are becoming taller and narrower for longer sentences. On the other hand, head-initial dependencies are expected to produce shorter dependency distances than the head-final ones \citep{liu2017dependency}; we do observe that longer sentences are becoming more head-initial as well. Interestingly, the discrepancy in the trends between shorter and longer sentences also seems to `support' the anti-\ddm{} phenomenon in short sentences \mbox{\citep{ferrer2021anti}}. The above facts prompt an intriguing question on what is the causal relationship between these observed syntactic changes and \ddm{}---we defer this investigation to future research.}

\section{Conclusion}

This study delved into diachronic language changes in English and German.
We comprehensively examined 15 
\se{metrics}, 
including the mean dependency distance, across extensive corpora containing political debates. 
By analyzing the agreement among five 
dependency parsers, including the commonly used Stanford CoreNLP, we provided novel insights into \yc{the importance of used parsers in such language change research}. 
Notably, we discovered a distinctive uptrend in the mean dependency distance in German, which is consistent across various parsers and sentence lengths. 
\se{We also found that German and English seemed undergo similar syntactic changes in the two political corpora examined, with only few instances showing opposing trends. We also observed more syntactic diachronic variability in sentences of extreme lengths, e.g., below or above mean sentence lengths.} \sen{Our results may be relevant in downstream tasks, e.g., for text generation systems that map across historic time frames, e.g., diachronic summarizers or MT systems \citep{Zhang2023CrosslingualCS,Thai2022ExploringDL}.}


Possible limitations of this work are the following: (1) The text genre explored was \ab{restricted} to political debates, in order to make the subcorpora comparable, but this may be an important factor affecting the measures. While there are studies showing a trend towards colloquialization in the parliamentary debates of some languages \cite{hiltunen-etal-2020, hou-smith-2021-drivers}, this has in fact not yet been studied for German. It is possible that one factor in the increase of dependency distance in the present study was in fact one such colloquialization trend, namely that an earlier more nominal style (typical of German written registers, especially legalese, replacing finite clauses with nominal constructions) gave way to a more natural/colloquial `verbal' style with more finite clauses. 
This can be illustrated with the example in (\ref{nominalverbal}). (\ref{previous}) is an attested example from the corpus. 
\ab{Such complex nominal structures might over time have been replaced by finite clauses as in the constructed example in (\ref{now}).}

\pex \label{nominalverbal} 
\a die Beschlüsse, welche das Haus bei §55 \textbf{durch die Annahme der Kommissionsvorschläge} gefaßt hat. (1876-11-21)\label{previous}\\
    `the resolutions, which the house has taken regarding §55 through adoption of the proposals by the committee'
    \a die Beschlüsse, die das Haus bei § 55 gefasst hat, \textbf{als es die Kommissionsvorschläge angenommen hat}.\label{now} \\
    `the resolutions, which the house has taken regarding §55 when it adopted the proposals by the committee ''
\xe
\noindent
We will look at this possibility more closely in a future study. 
(2) The time depth of the used diachronic corpora only covers the last 200 years---which may be too short to draw conclusions about syntactic change, which tends to take several centuries to unfold (e.g. \citealt{Kroch2001}).
(3) We did not thoroughly investigate the origins of the changes in measures. For instance, \citet{juzek2020exploring} indicate  that the downtrend in dependency distance over time is because dependency relations with shorter distances appear more frequently in the later periods and vice versa, 
while \citet{liu2022dependency} show that longer sentences have more dependency relations with decreasing dependency distances compared to the short ones, which results in the discrepancy in the trends for short and long sentences. 



\ab{
\se{Our study has been mostly formal---investigating the 
\sen{role} 
of parsers and metrics on the identification of language change in English and German---}but remains preliminary regarding the detection of actual syntactic change. 
\se{Future work should complement our analysis in this respect.} 
}


\starttwocolumn
\bibliography{compling_style}

\clearpage
\section{Appendix}

\begin{figure*}[h]
	\centering
 \hspace*{-2cm}
 \vspace*{-2cm}
  \includegraphics[page=2,height=1.3\textwidth,angle=90]
 {figs/annotation_guidelines.pdf}
 \hspace*{-2cm}
	\includegraphics[page=1,height=1.3\textwidth,angle=90]{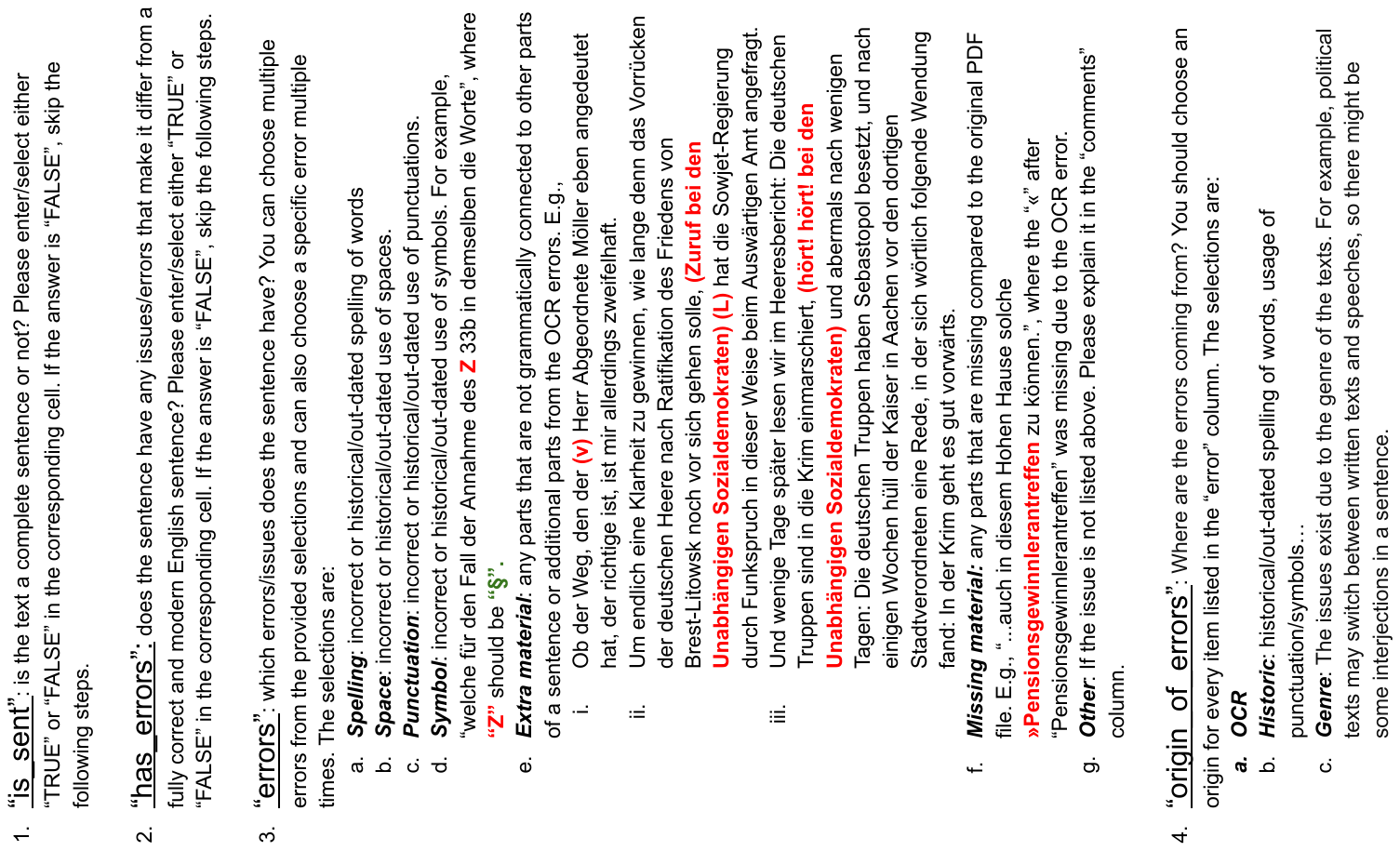}
	\caption{Annotation Guidelines for Hansard.}
	\label{fig:guideline}
\end{figure*}

\begin{figure*}[!h]
    \centering
    \begin{subfigure}[t]{0.5\linewidth}
        \includegraphics[width=\linewidth]{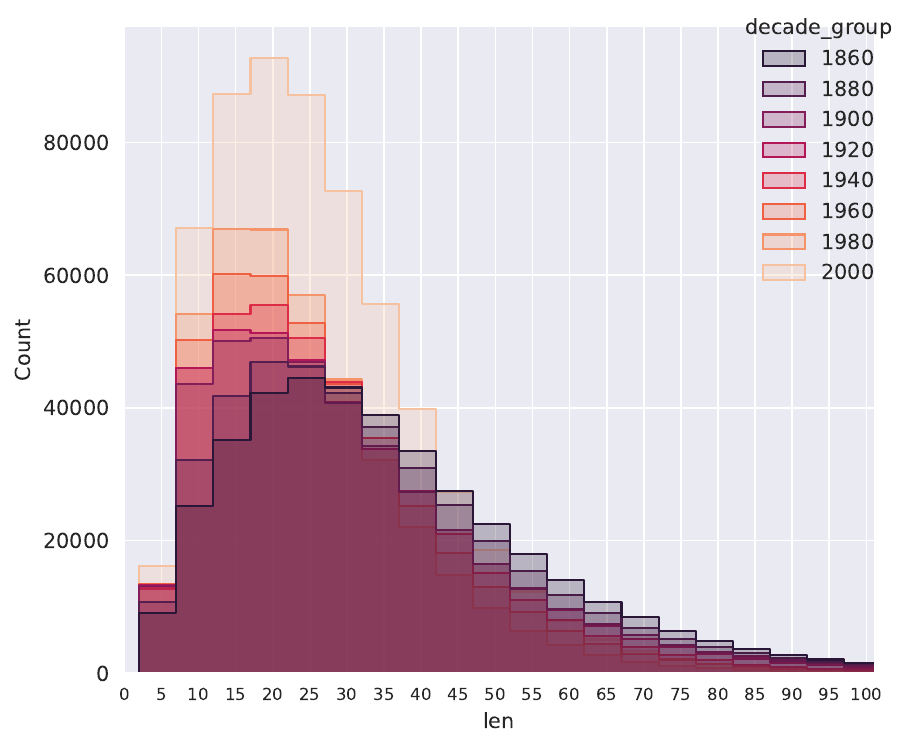}
        \caption{Hansard}
        \label{fig:len_dis_en}
    \end{subfigure}%
    \begin{subfigure}[t]{0.5\linewidth}
        \includegraphics[width=\linewidth]{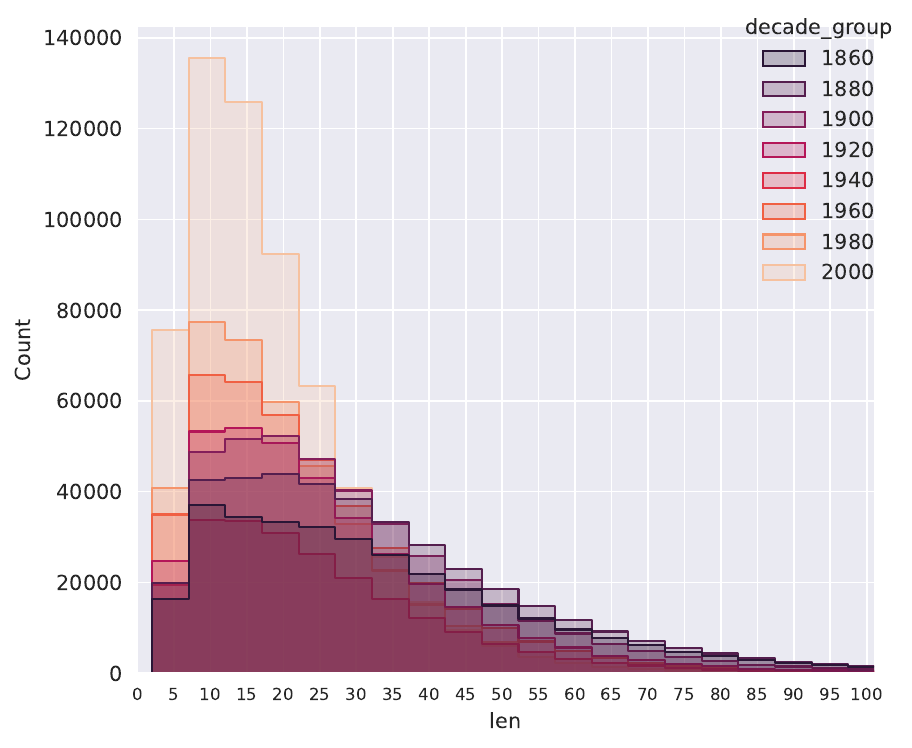}
        \caption{DeuParl}
        \label{fig:len_dis_de}
    \end{subfigure}%
    \caption{
    \yc{Histogram of sentence length distribution}
    with bin size of 5 for 
    the randomly sampled sentences having at most 100 tokens (which compose 99\% of the data). \ycj{Darker colors indicate older periods and vice versa.}
    }
    \label{fig:len_dis}
    \vspace{-.3cm}
\end{figure*}

\subsection{Parser Configurations}\label{app:parser_setting}
We train model checkpoints for StackPointer, Biaffine, and CRF2O on the concatenated train set of the treebanks in UD2.12, separately for each language.\footnote{We use the implementation of Biaffine and CRF2O from \url{https://github.com/yzhangcs/parser}, StackPointer from \url{https://github.com/XuezheMax/NeuroNLP2}.} 
The authors of TowerParse released the checkpoints trained on individual treebanks in UD2.5; 
as they did not provide the training script, we directly take the models trained on the largest treebank for each language, i.e., EWT for English and HDT for German.\footnote{\url{https://github.com/codogogo/towerparse}}
For CoreNLP and Stanza,
we use them with the out-of-the-box setting\se{s}.\footnote{We use Stanza of version 1.5.1 and CoreNLP via the client API bundled in Stanza.}
Note that the default models in Stanza 1.5.1 were also trained using UD2.12, 
however,
the default German model is only trained on the GSD treebank, and the default English model is trained on the combination of 5 out of 7 train sets in the treebanks. 

Except for Stackpointer, which does not have a built-in tokenizer, and CoreNLP, which has its own tokenizer, the other parsers implement the tokenizer from Stanza by default. Thus, for ease of use,
we implement the Stanza tokenizer for every parser consistently, disabling sentence segmentation, as we will apply them to the curated sentences from our target corpora, for which further sentence splitting is unexpected.

\subsection{Comparison to \citet{zeman-EtAl:2018:K18-2}}\label{app:comparison_parser}

\begin{figure*}[!h]
    \centering
    \begin{subfigure}[t]{0.48\linewidth}
        \includegraphics[width=\linewidth]{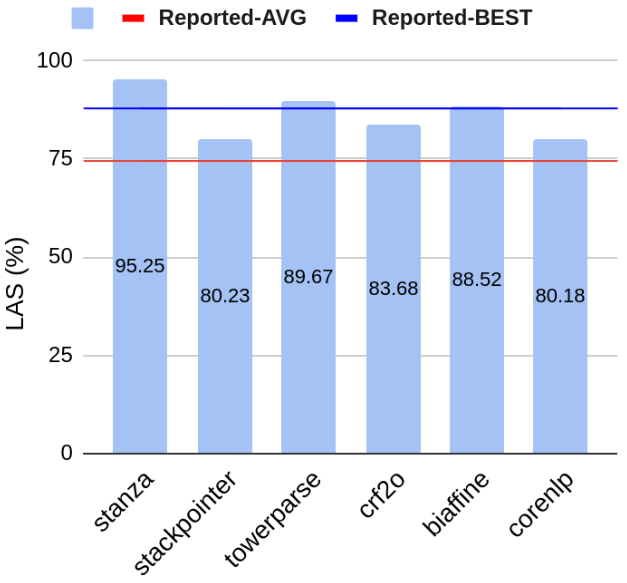}
        \caption{EN-PUD}
        \label{fig:enter-label}
    \end{subfigure}%
    \begin{subfigure}[t]{0.5\linewidth}
        \includegraphics[width=\linewidth]{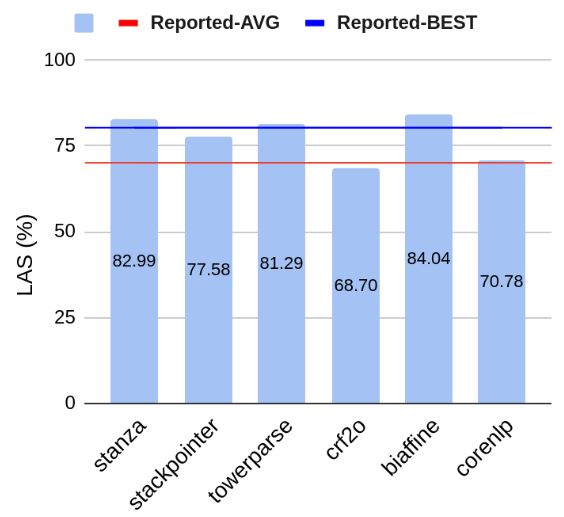}
        \caption{DE-GSD}
        \label{fig:enter-label}
    \end{subfigure}%
    \caption{Barplot of LAS of the parsers in our evaluation in comparison to the \textcolor{red}{average} and the \textcolor{blue}{best} LAS over various parsers on the English PUD and German GSD test sets reported in Table 15 of \citet{zeman-EtAl:2018:K18-2}. }
    \label{fig:conll}
    \vspace{-.3cm}
\end{figure*}
As Figure \ref{fig:conll} \ab{shows}, the great majority of the parsers obtain higher LAS than the average LAS over parsers in \citet{zeman-EtAl:2018:K18-2}, except for CRF2O on German, whose LAS is slightly lower than the average. Stanza, Towerparse, and Biaffine consistently outperform the best systems across different languages. Thus, we confirm the legitimacy of the used parsers.

\begin{figure*}[!h]
\vspace{-.2cm}
    \centering
    \begin{subfigure}[t]{0.5\linewidth}
        \includegraphics[width=\linewidth]{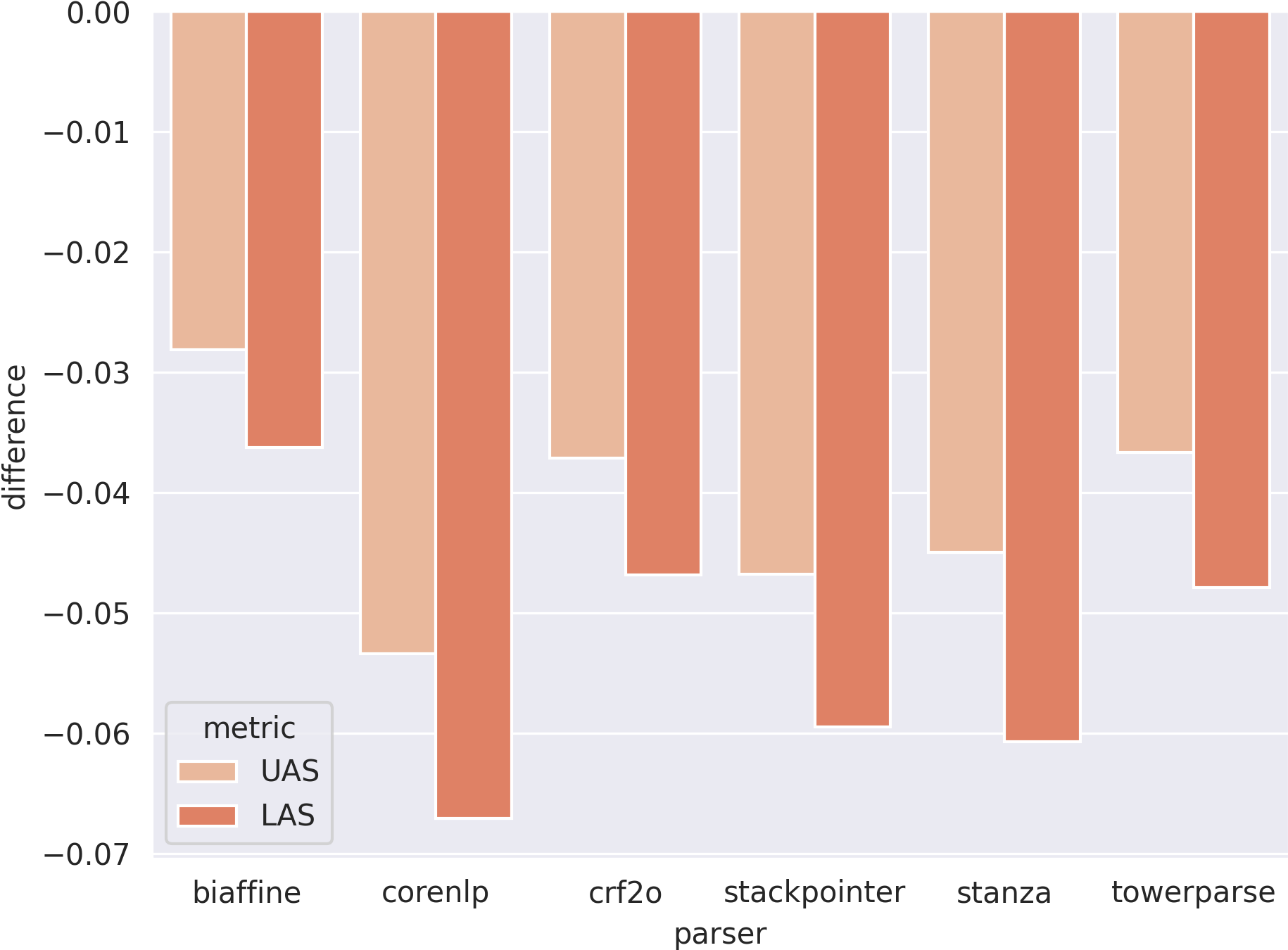}
        \caption{Historical Spelling Attack}
        \label{fig:attack_his}
    \end{subfigure}%
    \hspace{1cm}
    \begin{subfigure}[t]{0.4\linewidth}
        \includegraphics[width=\linewidth]{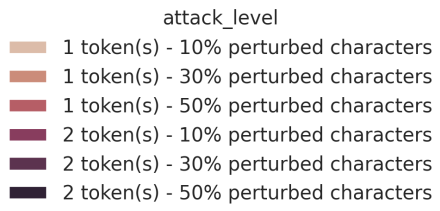}
    \end{subfigure}%
    
    \begin{subfigure}[t]{\linewidth}
        \includegraphics[width=\linewidth]{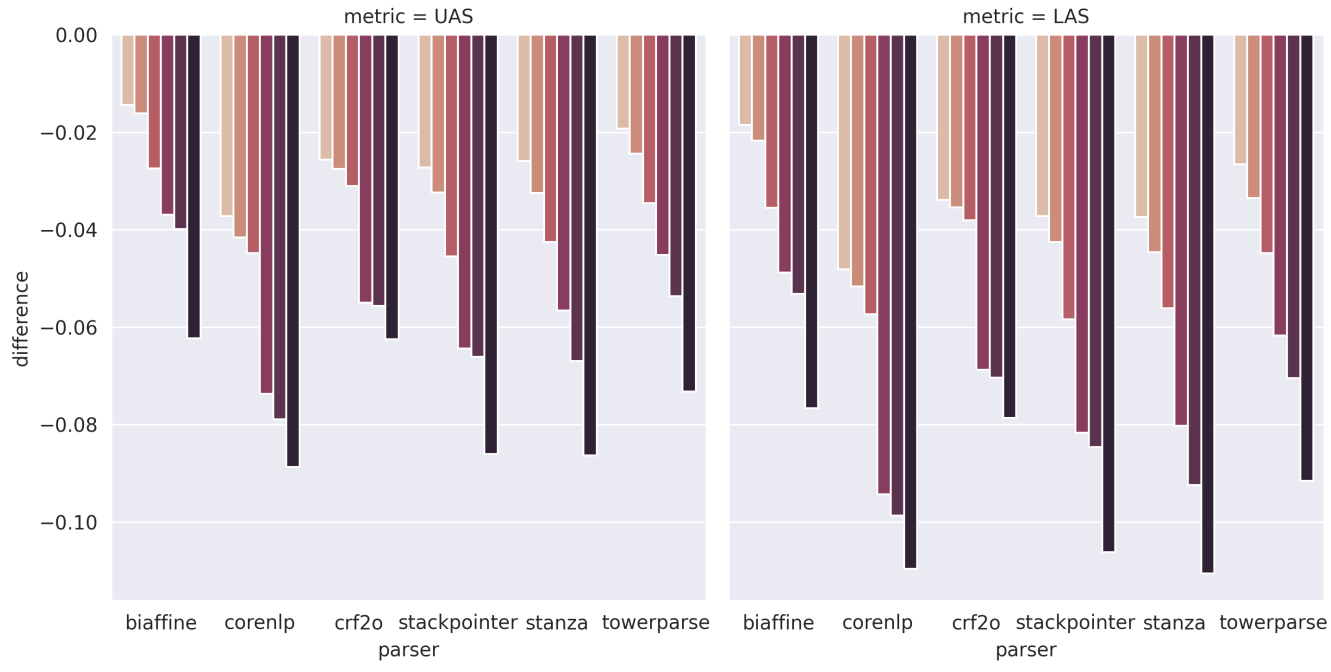}
        \caption{OCR Spelling Attack}
        \label{fig:attack_ocr}
    \end{subfigure}%
    \caption{Absolute difference in UAS and LAS between the sentences before and after the attack.}
    \label{fig:attack}
    \vspace{-.3cm}
\end{figure*}

\subsection{Sensitivity of metrics to data noise}\label{app:analysis_noise}
\yc{While our results in \S\ref{sec:parser} indicate that the parsers perform decently on the target corpora, it remains uncertain whether the metrics derived from their parsing results are affected by data noise and to what extent. Since we will inspect the 
trends of the metrics over time, it is important to ensure that the data noise does not disrupt the rankings of those metrics. Hence,}
we compute Spearman's $\rho$ correlations between 
\se{our 15} 
metrics 
on the \ycf{parses of the}
original and human corrected sentences (\se{without spelling and OCR errors, etc.}; collected in Section \ref{sec:data}); 
higher correlations indicate the rankings are less disrupted
and vice versa.
\begin{table*}[!ht]
\resizebox{.85\linewidth}{!}{%
\begin{tabular}{@{}ccccccc@{}}
\toprule
\multicolumn{1}{l|}{Metrics}                                & Stanza & CoreNLP & StackPointer & Biaffine & \multicolumn{1}{c|}{TowerParse}                 & AVG   \\ \midrule
\multicolumn{1}{l|}{\mdd{}}                  & 0.976  & 0.905   & 0.823        & 0.966    & \multicolumn{1}{c|}{0.888} & 0.912 \\
\multicolumn{1}{l|}{\ndd{}}                  & 0.995  & 0.892   & 0.862        & 0.985    & \multicolumn{1}{c|}{0.960} & 0.939 \\
\multicolumn{1}{l|}{\height{}}               & 0.983  & 0.915   & 0.862        & 0.986    & \multicolumn{1}{c|}{0.820} & 0.913 \\
\multicolumn{1}{l|}{\headratio{}}   & 0.975  & 0.954   & 0.959        & 0.972    & \multicolumn{1}{c|}{0.909} & \textbf{0.954} \\
\multicolumn{1}{l|}{\degree{}}               & 0.968  & 0.916   & 0.870        & 0.973    & \multicolumn{1}{c|}{0.815} & 0.908 \\
\multicolumn{1}{l|}{\leaves{}}          & 0.998  & 0.999   & 0.908        & 0.997    & \multicolumn{1}{c|}{0.832} & 0.947 \\
\multicolumn{1}{l|}{\degvar{}}          & 0.966  & 0.935   & 0.923        & 0.976    & \multicolumn{1}{c|}{0.834} & 0.927 \\
\multicolumn{1}{l|}{\degmean{}}         & 0.998  & 0.988   & 0.902        & 0.998    & \multicolumn{1}{c|}{0.832} & 0.944 \\
\multicolumn{1}{l|}{\depvar{}}           & 0.982  & 0.910   & 0.861        & 0.976    & \multicolumn{1}{c|}{0.827} & 0.911 \\
\multicolumn{1}{l|}{\depmean{}}          & 0.987  & 0.918   & 0.874        & 0.980    & \multicolumn{1}{c|}{0.818} & 0.915 \\
\multicolumn{1}{l|}{\headdis{}} & 0.989  & 0.967   & 0.891        & 0.990    & \multicolumn{1}{c|}{0.841} & 0.936 \\
\multicolumn{1}{l|}{\#Crossing}        & 0.942  & -       & 0.815        & 0.938    & \multicolumn{1}{c|}{0.782} & \underline{0.869} \\
\multicolumn{1}{l|}{\lpath{}}        & 0.995  & 0.935   & 0.874        & 0.994    & \multicolumn{1}{c|}{0.884} & 0.936 \\
\multicolumn{1}{l|}{\rootdis{}}       & 0.999  & 0.870   & 0.873        & 0.965    & \multicolumn{1}{c|}{0.976} & 0.937 \\
\multicolumn{1}{l|}{\random{}} & 0.991  & 0.991   & 0.894        & 0.986    & \multicolumn{1}{c|}{0.830} & 0.938 \\ \midrule
\multicolumn{1}{l|}{AVG}                  & \textbf{0.983}  & 0.935   & 0.879        & 0.979    & \multicolumn{1}{c|}{\underline{0.857}} & 0.926 \\ \bottomrule
\end{tabular}%
}%
\caption{Spearman's $\rho$ between the metrics on the original sentences and that on the human-corrected sentences (collected in Section \ref{sec:data}). The ``AVG'' values in the rows refer to the average correlation over different parsers for a specific metric; those in the columns denote the average correlation over different metrics for a specific parser. 
We bold the highest and underline the lowest average correlations. }
\label{tab:corr_measure}
\end{table*}

Overall, as
Figure \ref{tab:corr_measure} illustrates,
the metrics \ycf{for} the original and corrected sentences strongly correlate with each other, shown with an overall 
correlation of 0.926. This overall correlation is computed by first averaging over parsers for each metric and then averaging over all metrics. The lowest average correlation per metric is observed with \#crossing, 0.869, which, however, still implies a strong correlation. The least sensitive metric is the 
\yc{head-final ratio,}
shown with an average correlation of 0.954 over parsers. Among the parsers, Stanza is least sensitive (0.983), followed by Biaffine (0.979), whereas TowerParse and StackPointer are the most sensitive ones (0.857 and 0.879 respectively). For this reason, 
\ycj{we conclude that our metrics are not much 
sensitive to data noise in the target corpora across different parsers.} 

\begin{table*}[h]
\centering
\begin{subfigure}[]{\textwidth}
\resizebox{.95\linewidth}{!}{%
\begin{tabular}{lccccccccc}
\toprule
Metric & 5-7 & 10-12 & 15-17 & 20-22 & 30-32 & 40-42 & 50-52 & 60-62 & 70-72 \\ \midrule
\mdd{} & - & - & - & - & - & - & - & - & -3 \\
\ndd{} & +4 & - & - & - & +4 & - & - & - & - \\
\rootdis{} & +3 & - & - & - & - & - & - & - & - \\
\height{} & -5 & - & - & +3 & - & +4 & - & +5 & +5 \\
\depvar{} & -5 & - & - & +3 & - & +4 & +3 & +5 & +5 \\
\depmean{} & -5 & -3 & - & - & - & +3 & +3 & +5 & +5 \\
\degree{} & +5 & - & - & - & -5 & -3 & -5 & -5 & -4 \\
\degvar{} & +5 & - & +5 & - & - & -3 & -3 & -4 & -4 \\
\degmean{} & +5 & - & - & - & - & - & - & - & +5 \\
\headratio{} & +5 & - & - & -3 & -3 & - & -4 & -4 & -5 \\
\headdis{} & -5 & - & -3 & - & - & - & - & - & +3 \\
\leaves & +5 & - & - & - & - & - & - & - & - \\
\crossing{} & - & - & - & - & - & - & - & - & - \\
\lpath{} & - & -3 & - & - & - & - & - & - & - \\
\random{} & - & - & -3 & - & - & - & - & - & - \\ \bottomrule
\end{tabular}%
}%
\caption{English}\label{tab:trend_en}
\end{subfigure}%

\begin{subfigure}[]{\textwidth}
\resizebox{.95\linewidth}{!}{%
\begin{tabular}{lccccccccc}
\toprule
Measure & 5-7 & 10-12 & 15-17 & 20-22 & 30-32 & 40-42 & 50-52 & 60-62 & 70-72 \\ \midrule
\mdd{} & +5 & +5 & +5 & +4 & - & +5 & +5 & +4 & - \\
\ndd{} & -5 & -5 & - & -3 & -5 & - & - & - & - \\
\rootdis{} & - & - & - & - & -4 & - & - & - & - \\
\height{} & -5 & - & - & - & +4 & - & - & +3 & +3 \\
\depvar{} & -5 & - & - & - & +3 & - & - & +3 & +3 \\
\depmean{} & -5 & -3 & - & - & +4 & +5 & +4 & +3 & +3 \\
\degree{} & +5 & +5 & - & -4 & -4 & -5 & -5 & -5 & -5 \\
\degvar{} & +5 & +5 & - & -4 & -4 & -5 & -5 & -5 & -5 \\
\degmean{} & +4 & +5 & - & - & - & -3 & -3 & -4 & -4 \\
\headratio{} & - & - & - & - & -4 & -5 & -5 & -5 & -5 \\
\headdis{} & -5 & -5 & -3 & - & +4 & +4 & +5 & +5 & +4 \\
\leaves & +5 & +5 & - & - & -4 & -5 & -5 & -5 & -5 \\
\crossing{} & - & -3 & -3 & -4 & - & - & - & - & - \\
\lpath{} & +5 & +5 & +3 & - & +4 & +5 & +5 & +4 & +3 \\
\random{} & - & - & - & - & - & - & - & - & - \\ \bottomrule
\end{tabular}%
}%
\caption{German}\label{tab:trend_de}
\end{subfigure}
\caption{Metrics that have a stable trend supported by at least 3 parsers (except for \crossing{}, for which we relax the minimal requirement to 2 parsers, as CoreNLP is unable to predict crossing edges.). ``$+$'' and ``$-$'' denote the increasing and decreasing trend respectively; the values refer to the number of parsers that support this trend.}
\label{tab:trend}
\end{table*}

\end{document}